\documentclass[runningheads]{llncs}

\usepackage{eccv}

\usepackage{eccvabbrv}

\usepackage{graphicx}
\usepackage{booktabs}
\usepackage{url}            %
\usepackage{amsfonts}       %
\usepackage{nicefrac}       %
\usepackage{lmodern}        %
\usepackage{microtype}      %
\usepackage{xcolor}         %
\usepackage{subcaption}
\usepackage{enumitem}

\usepackage{amsmath}
\usepackage{mathtools}
\usepackage{makecell, tabularx}
\newcolumntype{Y}{>{\centering\arraybackslash}X}
\usepackage{rotating}
\usepackage[export]{adjustbox}
\usepackage{pgfplots}
\usepackage{array}
\usepackage{multirow}
\usepackage{stfloats}
\usepackage{placeins}

\usepackage[accsupp]{axessibility}  %

\usepackage{hyperref}

\usepackage{orcidlink}

\begin{document}

\title{P-CORE: Self-Supervised Surface Consistency for Point-Based Neural Editing} 

\titlerunning{P-CORE: Self-Supervised Surface Consistency for Editing}

\author{Yanshu Zhang\inst{1}\thanks{Corresponding author.}\orcidlink{0009-0002-6046-3966} \and Shichong Peng\inst{1}\orcidlink{0009-0005-8404-6392} \and Mehran Aghabozorgi\inst{1}\orcidlink{0009-0007-0447-1116} \and Alireza Moazeni\inst{1}\orcidlink{0009-0002-6532-0112} \and Ke Li\inst{1,2,3}\orcidlink{0000-0002-3229-271X}}

\authorrunning{Y.~Zhang et al.}

\institute{Simon Fraser University, Canada \and Alberta Machine Intelligence Institute (Amii), Canada \and Canadian Institute for Advanced Research (CIFAR), Canada \\ \email{\{yanshu\_zhang, shichong\_peng, maa143, sam62, keli\}@sfu.ca}}

\maketitle

\begin{abstract}
Advances in neural rendering have enabled high-fidelity multi-view reconstruction of 3D scenes. However, free-form non-rigid shape editing remains a significant challenge. 
Point-based neural representations are highly desirable for multi-view reconstruction because they lack fixed connectivity, which does not constrain the learned surface topology to that of the initialization.
Yet this same property causes point-based representations to struggle with holes and surface discontinuities under large deformations. To address this, we propose a novel self-supervised method to enable point-based representations to adapt to large deformations without requiring ground truth multi-view images of deformed geometry. The key idea is to generate random deformations and to ensure consistency in the predicted surface before and after deformation. In particular, the surface prediction from the deformed point cloud should be the same as the deformation applied to the surface prediction from the original point cloud. We incorporate our approach into attention-based point representations, which differ from splatting-based point representations in their use of a learned interpolation kernel between points as opposed to a Gaussian kernel around each point. This learned interpolation kernel can learn to adapt to large deformations, without requiring addition or removal of points. We show that our framework significantly enhances its robustness to large deformations. 
Experiments on synthetic geometry editing benchmarks (Neural Editor, Objaverse) demonstrate that our approach outperforms existing point-based methods in zero-shot editing and significantly reduces artifacts. Furthermore, qualitative results on the DTU and Mip-NeRF 360 datasets demonstrate our method’s effectiveness on real-world scenes. Project page: \href{https://zvict.github.io/p-core/}{https://zvict.github.io/p-core/}.
\end{abstract}

\section{Introduction}
\label{sec:intro}

Achieving free-form non-rigid editing of photo-realistic neural scenes is highly desirable for applications in animation, design, and gaming. Recent advances in neural rendering~\cite{mildenhall2021nerf,muller2022instant,Xu2022PointNeRFPN,kerbl20233dgauss,zhang2023papr} have enabled high-fidelity multi-view reconstructions, but performing \emph{zero-shot} deformation of these reconstructed scenes remains an open challenge. Point-based neural renderers have emerged as a highly desirable representation for reconstruction; by lacking fixed connectivity, they do not constrain the learned surface topology to that of the initialization, offering exceptional flexibility in capturing high-fidelity details of complex structures.

However, when transitioning from static reconstruction to free-form non-rigid editing, this lack of connectivity becomes a double-edged sword. When the point cloud is deformed, the points drift apart because there are no explicit edges or polygons holding them together. Without explicit topological connections, point-based methods severely struggle with holes, see-through artifacts, and surface discontinuities under large deformations. To tackle these issues, existing approaches often rely on explicit spatial regularizations~\cite{Huang2023SCGSSG, han2025arapgsdragdrivenasrigidaspossible3d,SorkineHornung2007AsrigidaspossibleSM} or reintroduce proxy geometry such as cages~\cite{xu2022cagedeforming,Peng2022CageNeRFCN,Jambon2023NeRFshop,Huang2024GSDeformerDC} or meshes~\cite{yuan2022nerfedit,Liu2023NeuralIE,Wang2023MeshGuidedNI,Zhou2023SERFFI,Yang2022NeuMeshLD,Guedon2024GaussianFE,Jiang2024VRGSAP,Waczynska2024GaMeSMA,Gao2024MeshbasedGS,Gao2024ManiGSGS}. While these proxies can enforce structural integrity, they introduce a host of new limitations. Constructing an artifact-free proxy that tightly encloses the geometry without self-intersections is complex and often requires manual intervention. Furthermore, proxy-based methods are restricted to edits that the rigid topology can support, making part-level manipulations extremely challenging. 

\begin{figure}
    \centering
    \includegraphics[width=\textwidth]{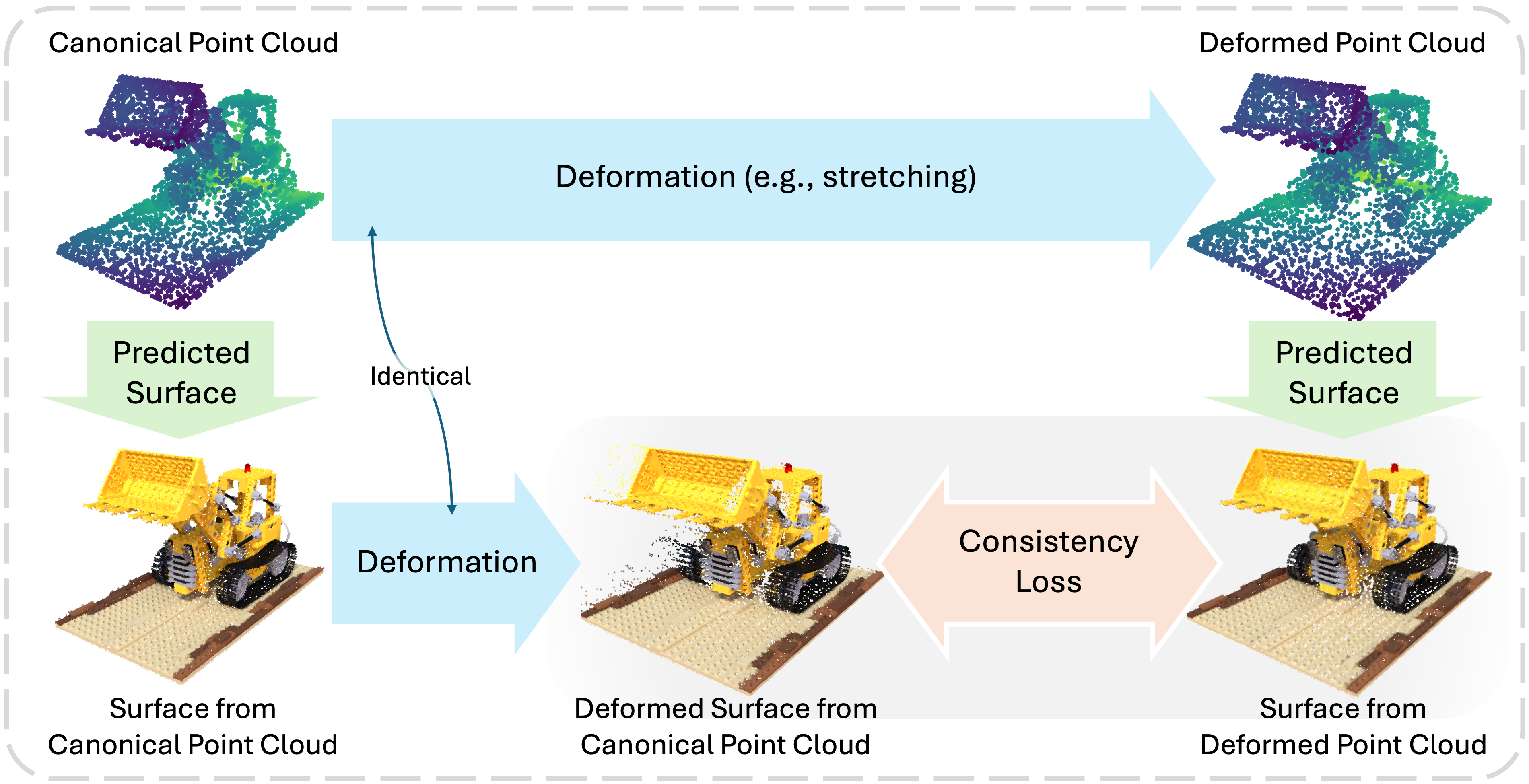}
    \caption{\textbf{Overview.} While point-based representations can learn a good surface prediction from multi-view images in its canonical space, they often struggle with holes and surface discontinuities during unseen deformations. To address this, we propose a self-supervised surface consistency framework. Given a point cloud representation learned in the canonical space, we predict the surface points from it. We then apply an identical deformation field to both the canonical point cloud and these extracted surface points. The resulting deformed surface points serve as geometric pseudo-ground truth to supervise the model's new surface predictions on the deformed point cloud. A stop-gradient operation is applied to the target points to prevent model collapse.}
    \label{fig:what}
\end{figure}

To overcome these limitations without sacrificing the flexibility of point-based representations, we introduce a novel \textbf{self-supervised surface consistency framework}, as illustrated in Figure~\ref{fig:what}. Instead of relying on ground truth deformed meshes or manually crafted spatial regularizations, our pipeline generates random deformations and enforces a surface consistency constraint: the surface predicted from the deformed point cloud should equal the deformation applied to the surface predicted from the original point cloud. By generating such geometric ``pseudo-ground truth'' for the deformed state and jointly enforcing photometric anchoring on the original state, our method significantly improves the generalizability of attention-based point renderers to unseen edits, ensuring hole-free, smooth surface reconstruction and high-fidelity rendering under non-rigid deformations.

We deliberately apply this framework to Proximity Attention Point Rendering (PAPR)~\cite{zhang2023papr}, an attention-based neural point renderer. Unlike 3D Gaussian Splatting (3DGS)~\cite{kerbl20233dgauss}, where discrete, anisotropic Gaussians must be coherently transformed (e.g., adjusting covariances) to avoid gaps and tears after deformation, PAPR treats each point as infinitesimal and fills gaps via a learned attention-based interpolation kernel between nearby points. This design is advantageous for two reasons: first, it eliminates the need to transform spatial extents under deformation, inherently supporting continuous surface prediction; second, the learned interpolation kernel can adapt to large deformations without requiring addition or removal of points, offering the capacity to generalize to unseen edits. We demonstrate that when coupled with our self-supervised framework, PAPR's robustness to large deformations is significantly enhanced, preventing topological breakage even under drastic non-rigid edits. 

\paragraph{Contributions} Below are our key contributions:
\begin{itemize}[itemsep=0pt, topsep=0pt, itemindent=0pt]
\item We propose a novel self-supervised fine-tuning framework that substantially reduces holes and surface discontinuities in attention-based point renderers without requiring ground-truth dynamic data or geometry proxies.
\item We demonstrate the successful coupling of this framework with an attention-based neural point renderer (PAPR), highlighting its inherent advantages over spatial-extent-based methods (e.g., 3DGS) for continuous surface prediction during large deformations.
\item We provide results on synthetic geometry editing benchmarks (Neural Editor, Objaverse) demonstrating that our method outperforms both proxy-based and proxy-free point-based methods. 
\end{itemize}

\section{Related Work}
\label{sec:related_work}

\subsection{3D Shape Deformation}
There is a long line of work in geometry processing on editing 3D shapes~\cite{Gain2008ASO, Yuan2021ARO}. Many surface-based methods use parametric patches or surface meshes as proxies to manipulate shapes~\cite{feng1996new, feng2006multiresolution, jin2000three, decaudin1996geometric, kobayashi2003t, singh1998wires, Ju2005MeanVC, angelidis2004swirling, von2006vector}, utilizing approaches like Laplacian~\cite{gao2019sparse, lipman2005laplacian, sorkine2007rigid, sorkine2004laplacian, sorkine2005laplacian} and cage-based~\cite{Ju2005MeanVC, yifan2020neural, zhang2020proxy} deformations. However, designing an appropriate proxy (e.g., mesh or cage) that closely fits the target model is challenging. Manual creation can be time-consuming and may require extensive expertise, while automatic generation methods might not always produce optimal proxies, especially for complex or highly detailed models.

To bypass proxy construction, traditional point-based deformation methods directly manipulate the shape using a set of freely positioned points or Moving Least Squares (MLS) handles~\cite{borrel1991deformation, borrel1994simple, bechmann1993animation, bechmann1995order, aubert1997volume, raffin2000curvilinear, bechmann2003arbitrary, hsu1992direct, hu2001direct, lee1995image, Moccozet1997DirichletFD, Ruprecht1993FreeFD, Botsch2005RealTimeSE, Zwicker2002Pointshop3A, Mller2005MeshlessDB, Alexa2000AsrigidaspossibleSI}. While offering greater flexibility, directly applying these traditional point editing algorithms to modern neural point renderers introduces a critical failure mode. Traditional algorithms only relocate the discrete 3D coordinates. Because point clouds lack connectivity, deformations that stretch or bend the shape cause the points to spread apart.

\subsection{Point-based Neural Rendering}
Point-based representations~\cite{Xu2022PointNeRFPN, kerbl20233dgauss, zhang2023papr} have gained significant attention in neural rendering. Some methods~\cite{Aliev2019NeuralPG,Rakhimov2022NPBGAN,Ost2021NeuralPL,Kopanas2021PointBasedNR} assume a given point cloud from Multi-View Stereo (MVS) or LiDAR, while others learn it from an initialization. Since point clouds natively lack connectivity, a fundamental challenge is how to fill the gaps between discrete points. One line of work focuses on predicting continuous surfaces from point clouds, employing classical techniques like Moving Least Squares (MLS)~\cite{Alexa2003ComputingAR, Levin2004MeshIndependentSI} and Poisson Surface Reconstruction~\cite{Kazhdan2006PoissonSR, Kazhdan2013ScreenedPS}, or more recent learning-based implicit surface prediction methods~\cite{Erler2020Points2SurfLI,Ma2020NeuralPullLS,Jiang2024VRGSAP,Lei2025OffsetOPTES}. Another line of work bypasses explicit meshing and directly renders the points. To fill the gaps during rendering, most methods associate a spatial extent with each point, either using 2D splats~\cite{Wiles2019SynSinEV,Lassner2021PulsarES,Rckert2021ADOPAD,Zhang2022DifferentiablePR,Zuo2022ViewSW} or 3D Gaussian splats~\cite{kerbl20233dgauss}. While these spatial extents enable high-quality rendering, they introduce parameters (e.g., covariances) that are difficult to correctly adjust during deformation. Instead, methods like PAPR~\cite{zhang2023papr} and Pointersect~\cite{chang2023pointersect} represent points without spatial extents, using an attention mechanism~\cite{Vaswani2017AttentionIA} to interpolate between nearby points and fill gaps.

\subsection{Editing of Neural Scene Representations}
To edit neural scene representations while preserving surface continuity, one family of approaches~\cite{yuan2022nerfedit,xu2022cagedeforming,Peng2022CageNeRFCN,Jambon2023NeRFshop,Wang2023MeshGuidedNI,Zhou2023SERFFI,Yang2022NeuMeshLD, Gao2024ManiGSGS} extracts and edits a proxy geometry (mesh or cage), then maps the edits back to the neural scene. However, this conversion introduces approximation errors and additional points of failure, because some geometric features easily represented implicitly cannot be represented in a proxy, and vice versa. Moreover, the edits are limited to what the rigid proxy connectivity supports—part-level editing (e.g., spinning wheels) becomes severely challenging.

Another line of work~\cite{Zheng2022EditableNeRFET,Huang2023SCGSSG,Wu20234DGS,Bian2025GSDiTAV,Feng2025St4RTrackS4} trains neural representations on dynamic scene observations to learn shape variations over time, associating each shape with keypoints so that the shape changes when the keypoints are edited. However, these methods fail to generalize to out-of-distribution shapes and edits not observed in the dynamic scenes.

\begin{figure}[!t]
  \centering
  \begin{subfigure}{0.528\linewidth}
    \centering
    \includegraphics[width=1\linewidth]{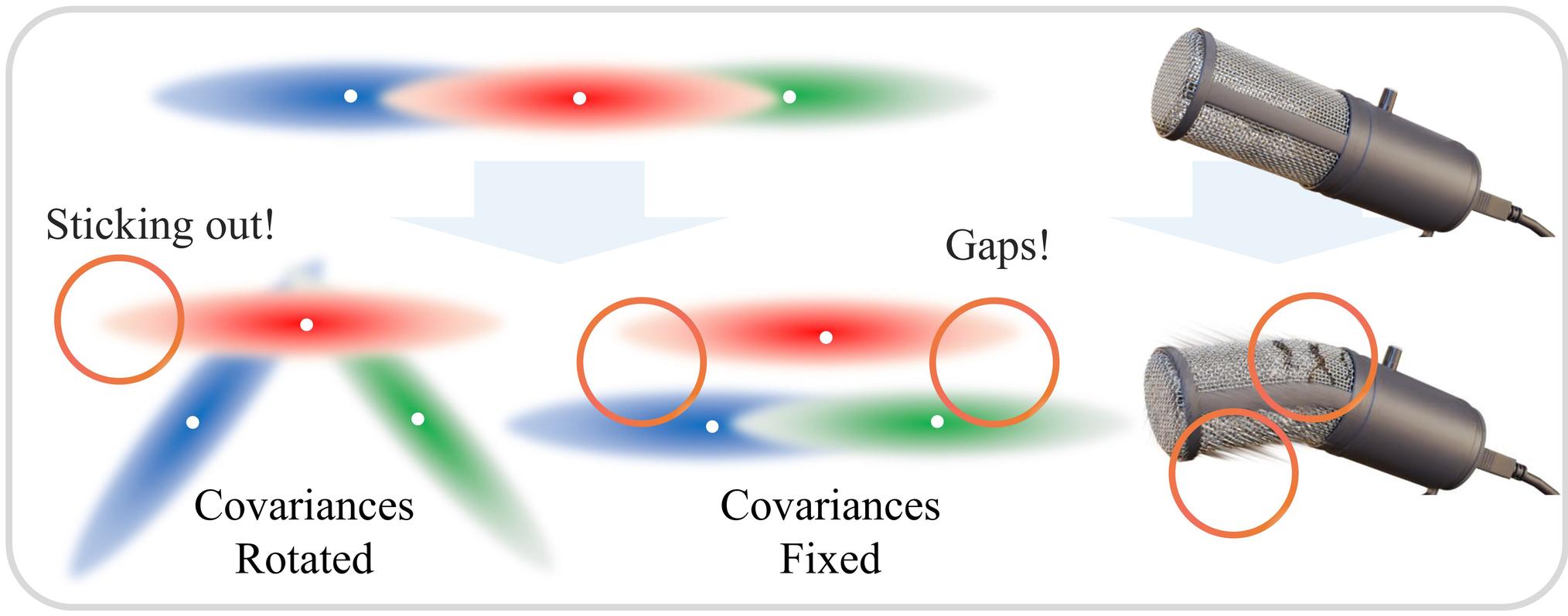}
    \caption{3D Gaussian Splatting}
  \end{subfigure}
  \begin{subfigure}{0.421\linewidth}
    \centering
    \includegraphics[width=1\linewidth]{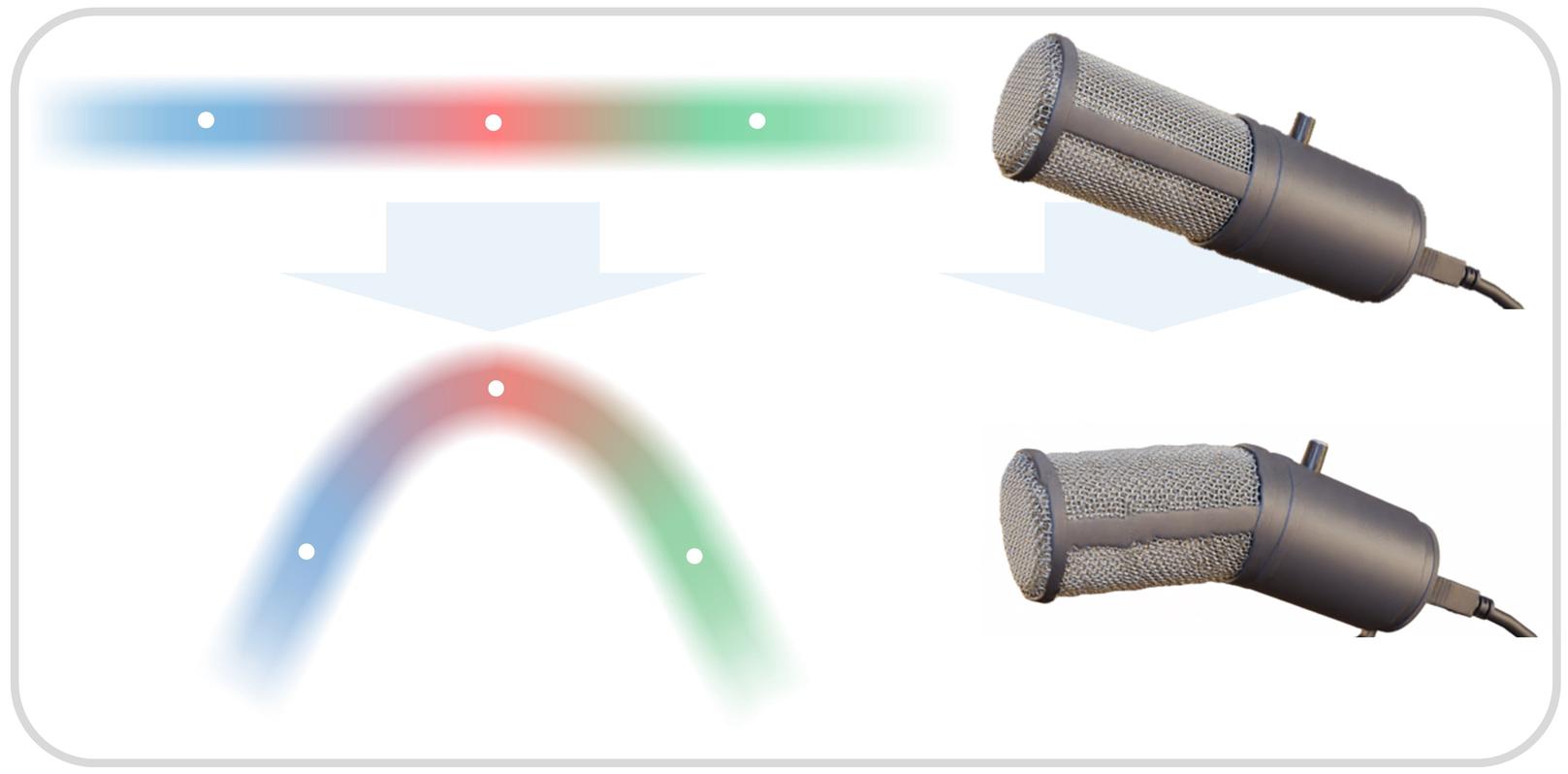}
    \caption{Attention-based Point Rendering}
  \end{subfigure}
  \caption{As shown on the left, in 3DGS editing the points requires adjusting each Gaussian splat's covariance to eliminate gaps. However, this adjustment may introduce distortions, as circled. Instead, attention-based methods model only the center of each point and learn a renderer that interpolates between points using an attention mechanism. This interpolation is based on the local point configurations, for example, the relative distances between points, with many variations observed from different rays during training. This variability enables attention-based methods to effectively generalize to unseen point configurations and find a promising interpolation of the points after editing, as shown on the right.} 
  \label{fig:why}
\end{figure}

More recently, a line of work~\cite{Wang2023RIPNeRFLR,Liang2022SPIDRSN,Wang2025RISEEditingRN,Wilczynski2026IRISIR,Huang2023SCGSSG,Gao2024ManiGSGS} edits scenes by directly manipulating neural point primitives. RIP-NeRF~\cite{Wang2023RIPNeRFLR} and RISE-Editing~\cite{Wang2025RISEEditingRN} learn rotation-invariant point features for fine-grained editing and cross-scene compositing. SPIDR~\cite{Liang2022SPIDRSN} couples a neural point field with an SDF to support geometry deformation and relighting. IRIS~\cite{Wilczynski2026IRISIR} attaches latent features to explicit Gaussian anchors for efficient part-level editing. The key distinction from our work lies in the rendering paradigm. First, all these methods use \emph{volume rendering}, whereas P-CORE builds on a \emph{surface rendering} point representation that produces a single attention-weighted ray–surface intersection (PAPR). For the editing task, volume rendering must populate the object interior with primitives, making it far less compact than a surface representation and requiring every edit to displace the interior primitives consistently with the surface. Second, 3DGS-based methods~\cite{Wilczynski2026IRISIR,Gao2024ManiGSGS,Huang2023SCGSSG} represent the scene with Gaussian primitives with spatial extent, so deformation must transform each primitive's covariance, which may cause distortion and surface tearing as shown in Fig.~\ref{fig:why}.

In contrast to these approaches, P-CORE operates directly on points, requiring neither a proxy geometry nor dynamic-scene observations. By rendering only the surface with extent-free points, P-CORE represents an object compactly and requires no interior or primitive extent transformation when editing.

\section{Preliminaries}
PAPR learns a point-based representation of a 3D scene from multi-view RGB images and corresponding camera parameters. The point-based representation $\mathcal{P} = \{(\mathbf{p}_i, \mathbf{\mu}_i)\},$ where $i \in \{1, 2, \dots, N\}$ represents a 3D scene with $N$ neural points. Each neural point has a learnable 3D position $\mathbf{p} \in \mathbb{R}^3$ and a learnable feature vector $\mathbf{\mu} \in \mathbb{R}^d$. 
Given a ray $\mathbf{r}_j$ represented by a camera center $\mathbf{o}_j \in \mathbb{R}^3$ and a ray direction $\mathbf{d}_j \in \mathbb{R}^3$ as the query, PAPR learns $K \ll N$ attention weights for each of the $K$ nearest points around the ray. It then uses the attention weights to aggregate the value embedding vectors for the $K$ points to get a feature vector $\mathbf{f}_j$ that captures the color of the ray. To map the query and key into the same feature space, PAPR uses MLPs to learn embedding vectors:
\begin{align}
    \mathbf{q}_{j} = f_{\theta_Q}\left(\gamma\left(\mathbf{d}_j\right)\right), \ \ 
    \mathbf{k}_{ij} = f_{\theta_K}\left( \left[\gamma\left(\mathbf{h}_{i,j}\right), \gamma\left(\mathbf{t}_{i,j}\right), \gamma\left(\mathbf{p}_i\right) \right] \right),
\end{align}
where $f_{\theta}$ are the embedding MLPs, $\gamma$ is the positional encoding function. $\mathbf{h}_{ij}$ and $\mathbf{t}_{ij}$ are two ray-dependent point feature vectors, computed as:
\begin{align}
    \mathbf{p}'_{ij} = \mathbf{o}_{j} + \langle \mathbf{p}_i - \mathbf{o}_{j}, \mathbf{d}_{j} \rangle \cdot  \mathbf{d}_{j}, \ \ 
    \mathbf{h}_{ij} = \mathbf{p}'_{ij} - \mathbf{o}_{j}, \; \mathbf{t}_{ij} = \mathbf{p}_i - \mathbf{p}'_{ij},
\end{align}
where $\mathbf{p}'_{ij}$ is the position of the point's projection on ray $\mathbf{r}_j$. These features capture the point configuration of the neighborhood by incorporating the features from all $K$ points around a ray. To model ray-dependent appearance, PAPR learns a ray-dependent feature vector $\mathbf{v}_{ij}$ for each point:
\begin{align}
    \mathbf{v}_{ij} = f_{\theta_V}\left( \left[\gamma\left(\mathbf{h}_{i,j}\right), \gamma\left(\mathbf{t}_{i,j}\right), \gamma\left(\mathbf{\mu}_i\right) \right] \right),
\end{align}
The ray's feature $\mathbf{f}_j$ can then be computed by $\mathbf{f}_j = \sum_{i=1}^K w_{ij} \mathbf{v}_{ij}$, where $w_{ij} = \text{softmax}(\langle\mathbf{q}_j, \mathbf{k}_{ij}\rangle / \sqrt{d_{\mathbf{k}}})$ are the attention weights, $d_{\mathbf{k}}$ is the dimension of $\mathbf{k}_{ij}$. By spatially aggregating the feature vectors of all the rays from the same camera view, PAPR produces a feature map of the view, which is then passed through a UNet-based renderer to predict RGB image $\hat{\mathbf{I}}$.

\section{Method}

\subsection{3D Gaussian Splatting vs. PAPR}

Compared to 3DGS, a key difference is that in PAPR, points have no spatial extent. Rather than using splats to fill gaps between points, PAPR's attention mechanism outputs a set of weights over the $K$ nearest points for every ray. As attention weights are non-negative and sum up to 1, it essentially learns a convex interpolation of those points. Such interpolations are based on the local point configurations around the rays, which vary significantly across different rays observed during training. This variability enables PAPR to generalize effectively to unseen point sets after deformation.

With 3DGS, it is crucial to adjust the covariance of each affected Gaussian when editing, otherwise it may cause surface discontinuities. As shown in Fig.~\ref{fig:why}, fixing covariances can result in gaps between splats after editing. Moreover, even with adjusted covariances, it's challenging to get them right, as adjustments vary between splats. In contrast, PAPR lacks spatial extent around points, so no spatial adjustment is needed---only the point locations must be changed. 

Furthermore, even if the covariances were adjusted optimally, as shown in the left of Fig.~\ref{fig:why}, the shape surface after editing can still be not as smooth as it was originally. Neither shrinking nor rotating the splats would work since the former causes gaps, while the latter makes splats stick out. 
This is caused by the fact that non-rigid deformations of Gaussians are not necessarily Gaussian anymore. As a result, more Gaussian splats would need to be added at locations where the non-rigidity is the greatest.
In contrast, since non-rigid deformations of a point set simply produce another point set, in PAPR, it suffices to move the points without adding new ones. The interpolator learned by PAPR should adapt to the changes in point positions, providing smooth and reliable interpolation for the updated point set, thereby preserving surface continuity.

\subsection{Canonical Initialization}

Our framework begins by capturing the canonical geometry and appearance of the target scene. We pre-train the PAPR model on a set of static multi-view images to extract a learned dense point cloud representation $\mathcal{P}$. This canonical initialization establishes a high-fidelity reference state, provides accurate implicit surface intersections and their associated photometric properties before any non-rigid edits are applied.

\subsection{Surface Point Extraction}

A key component of our framework is the extraction of differentiable surface intersection estimates from PAPR's attention mechanism. Given a camera ray $\mathbf{r}_j = (\mathbf{o}_j, \mathbf{d}_j)$ and its associated top-$K$ selected points $\{\mathbf{p}_{k}\}_{k=1}^{K}$ with attention weights $\{w_{kj}\}$, we define the surface intersection point for ray $\mathbf{r}_j$ as the attention-weighted average of the selected point positions:
\begin{align}
    \hat{\mathbf{s}}_j = \sum_{k=1}^{K} w_{kj} \cdot \mathbf{p}_k
\end{align}
As the model is only trained on the observations at the canonical state of the scene, the attention model often fails to generalize to unseen edits, resulting in holes in the surface prediction and artifacts in the renderings under deformation.

\subsection{Self-Supervised Fine-Tuning Objective}
\label{sec:ss_ft}

To ensure consistent surface prediction under deformation, we need to enhance the robustness of PAPR's attention model to unseen edits. A na\"ive solution is to fine-tune the learned PAPR model on the captures of the deformed scene, for example, in the case of dynamic scene reconstruction. However, such ground truth data are not often available. To bypass the need for ground-truth dynamic sequences, we introduce a self-supervised fine-tuning objective that utilizes analytical 3D deformation fields and creates ``pseudo'' supervision in the fine-tuning stage. 

At each fine-tuning step, we first perform a standard forward pass on the canonical point cloud $\mathcal{P}$ with a batch of training rays, extracting both rendered RGB values and \textit{canonical} surface points $\hat{\mathbf{s}}$ from the attention weights for each ray. We then apply an analytically defined 3D deformation field $\mathcal{D}: \mathbb{R}^3 \to \mathbb{R}^3$ to the point cloud, yielding the deformed point cloud $\mathcal{P}_{\text{def}} = \mathcal{D}(\mathcal{P})$.

We then re-query the model under the deformed point cloud with deformed rays (Sec.~\ref{sec:deformation_modes}) to extract predicted \textit{deformed} surface points $\hat{\mathbf{s}}_{\text{pred}}$, and compute a geometric loss between $\hat{\mathbf{s}}_{\text{pred}}$, and the deformed \textit{canonical} surface points $\mathcal{D}(\hat{\mathbf{s}}_j)$, using the identical deformation field $\mathcal{D}$:
\begin{align}
    \mathcal{L}_{\text{geo}} = \frac{1}{|\mathcal{F}|} \sum_{j \in \mathcal{F}} \| \hat{\mathbf{s}}_{\text{pred},j} - \mathcal{D}(\text{sg}[\hat{\mathbf{s}}_j]) \|^2 \label{eq:geo}
\end{align}
where $\mathcal{F}$ denotes the set of ray indices and $\text{sg}[\cdot]$ is the stop-gradient operator, which detaches the \textit{canonical} surface points $\hat{\mathbf{s}}$ from the computation graph before applying the deformation $\mathcal{D}$. This prevents gradients from flowing back through the canonical extraction and destabilizing the already-converged representation, ensuring only the deformed-state predictions $\hat{\mathbf{s}}_{\text{pred}}$ are optimized. Additionally, we enforce photometric consistency on the deformed rendering through a texture loss $\mathcal{L}_{\text{tex}} = \| \hat{I}_{\text{def}} - \hat{I}_{\text{can}} \|^2$, which anchors the appearance of the deformed state to the canonical rendering without requiring ground-truth deformed images. 

\subsection{Deformation Modes}
\label{sec:deformation_modes}

We introduce two complementary deformation modes that determine how camera rays interact with the deformed geometry, each addressing a distinct failure mode:

\paragraph{Points-only mode.} The point cloud and surface points are deformed, but the camera ray origins remain at their original positions. Foreground ray directions are recomputed to point toward the deformed ray termination points. This mode preserves the original camera viewpoint structure, ensuring robust rendering quality from distant views where the deformation is observed globally.

\paragraph{Rays-and-points mode.} In addition to deforming the point cloud and surface points, the ray origins are repositioned near the surface and then deformed by the same field. Ray directions are recomputed from the deformed origins toward the deformed termination points. This mode addresses the occlusion problem: when the point cloud deforms, the original ray directions may no longer correctly query the deformed surface due to self-occlusion from displaced geometry. By diversifying the set of ray directions the model encounters during training, this mode significantly enhances robustness to complex deformations.

In practice, we use both modes simultaneously.

\subsection{Joint Loss Formulation}

The total training objective combines geometric alignment on the deformed state with appearance anchoring on the canonical state:
\begin{align}
    \mathcal{L}_{\text{total}} = \mathcal{L}_{\text{rgb}} + \lambda_{\text{geo}} \cdot \mathcal{L}_{\text{geo}} + \lambda_{\text{tex}} \cdot \mathcal{L}_{\text{tex}}
\end{align}
where $\mathcal{L}_{\text{rgb}}$ is the canonical photometric rendering loss used in PAPR~\cite{zhang2023papr}, $\mathcal{L}_{\text{geo}}$ is the self-supervised geometric loss (Eq.~\ref{eq:geo}), and $\mathcal{L}_{\text{tex}}$ is the texture consistency loss, all defined in Sec.~\ref{sec:ss_ft}.

\section{Experiments}
\label{sec:exp}

\subsection{Datasets and Baselines}

To evaluate the structural resilience and rendering fidelity of our self-supervised framework, we test against both proxy-based and proxy-free point editing methods on the Neural Editor dataset~\cite{Chen2023NeuralEditorEN} and dynamic scene subsets from Objaverse~\cite{Deitke2022ObjaverseAU} introduced by PAPR-in-Motion~\cite{Peng2024PAPRIM}, a recent method that extends PAPR to synthesize smooth point-level 3D scene interpolations between different scene states without intermediate supervision. For these synthetic datasets, ground truth deformed states are available, enabling both qualitative and quantitative benchmark comparisons. We follow the scene selection and evaluation protocols from~\cite{Chen2023NeuralEditorEN} for Neural Editor and PAPR-in-Motion~\cite{Peng2024PAPRIM} for Objaverse, evaluating on scenes containing complex non-rigid transformations and part-level motions. Additionally, we qualitatively evaluate our method on real-world scenes from the DTU and Mip-NeRF 360 datasets as we lack ground truth deformation fields and renderings of the deformed scenes for these real-world datasets. We utilize a variety of analytical 3D deformation fields during our self-supervised training across all datasets, such as global twisting, scaling, dilating, and bending along random directions and varying degrees.

For proxy-based baselines, we compare with Deforming-NeRF~\cite{xu2022cagedeforming}, Neural Editor~\cite{Chen2023NeuralEditorEN} (cage-based), and Mani-GS~\cite{Gao2024ManiGSGS} (mesh-based). For proxy-free baselines, we compare with SC-GS~\cite{Huang2023SCGSSG} and the vanilla PAPR~\cite{zhang2023papr} model to explicitly demonstrate the robustness provided by our framework. SC-GS and Mani-GS represent two common ways to deform the covariance of the Gaussians (which must be deformed, otherwise Gaussians will easily extrude out of the surface); SC-GS uses the ARAP geometry regularizer, and Mani-GS uses a mesh as proxy geometry. In contrast, our method inherently treats each point as infinitesimal, so we can directly update the point positions without worrying about the shape of the points. For a fair comparison we apply identical deformation fields across all methods and evaluate the resulting visual and structural fidelity. In our method, the fine-tuning requires only $\sim$500 steps ($<$ 1min)---roughly 1/500 of the pre-training duration---using the same learning rates and batch size, making it a lightweight post-processing step.

\begin{figure}[htb]
\tiny
\centering
\begin{tabularx}{1\linewidth}{l@{\hskip 1em}YYYYYYYY}
    &   Chair
    &   Drums
    &   Ficus
    &   Hotdog
    &   Lego
    &   Materials
    &   Mic
    &   Ship
    \\
\rotatebox[origin=c]{90}{SC-GS}          
    &   \includegraphics[width=\hsize,valign=m]{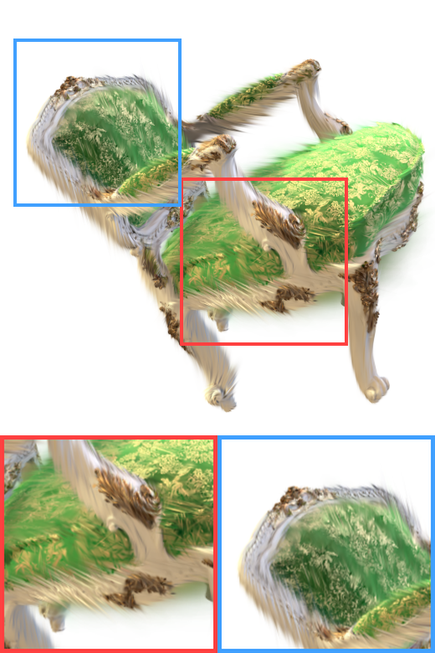}
    &   \includegraphics[width=\hsize,valign=m]{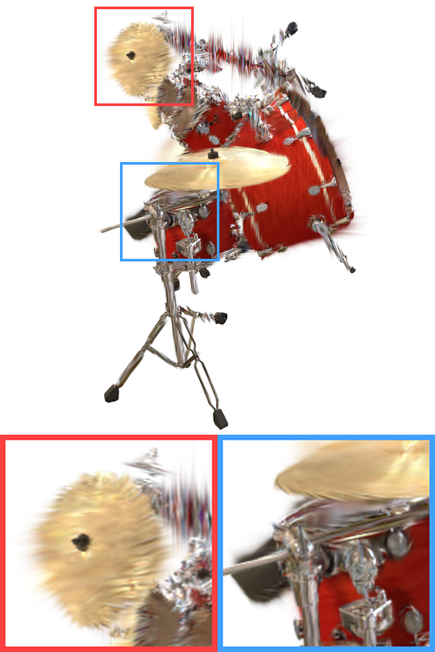}
    &   \includegraphics[width=\hsize,valign=m]{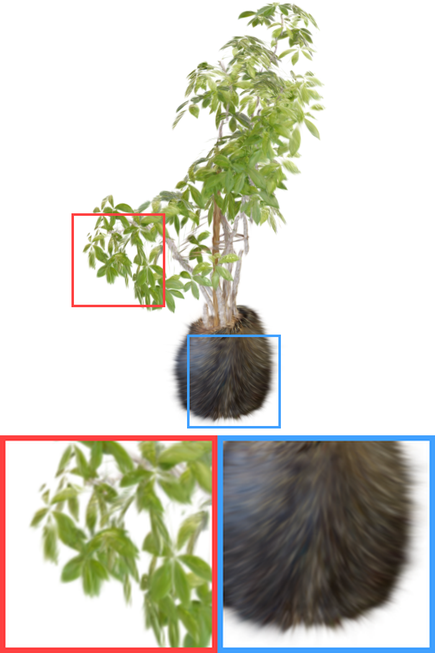}
    &   \includegraphics[width=\hsize,valign=m]{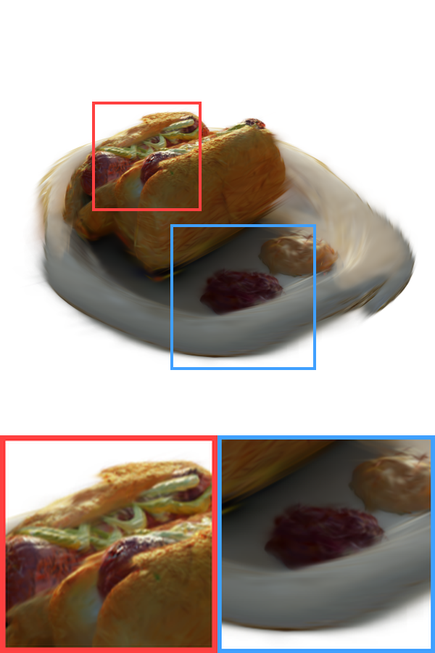}
    &   \includegraphics[width=\hsize,valign=m]{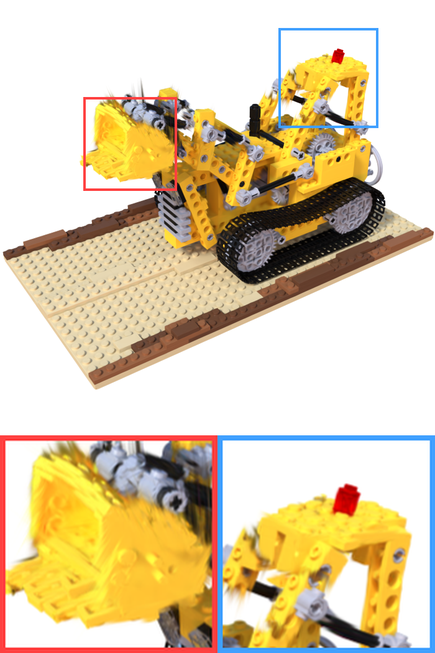}
    &   \includegraphics[width=\hsize,valign=m]{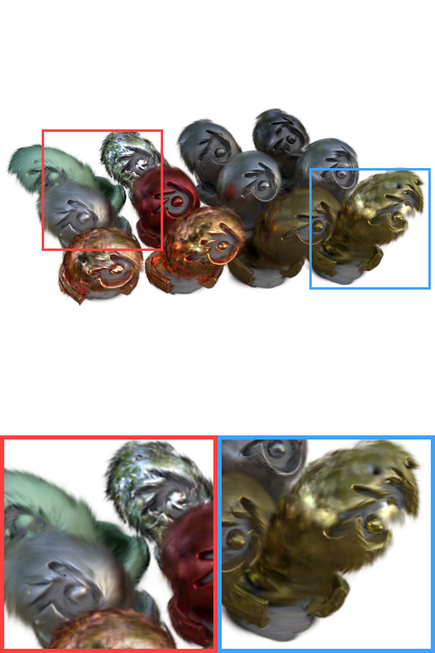}
    &   \includegraphics[width=\hsize,valign=m]{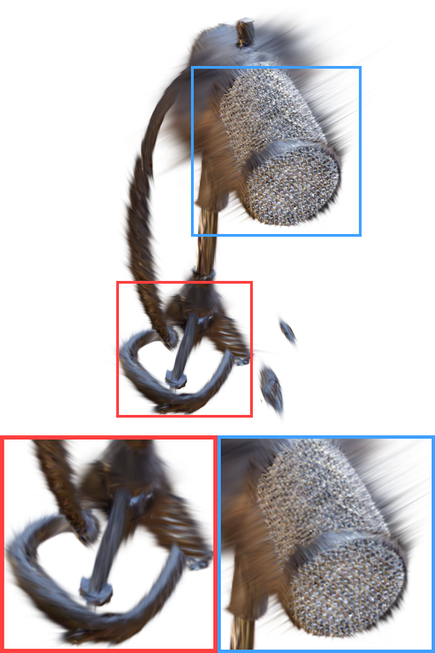}
    &   \includegraphics[width=\hsize,valign=m]{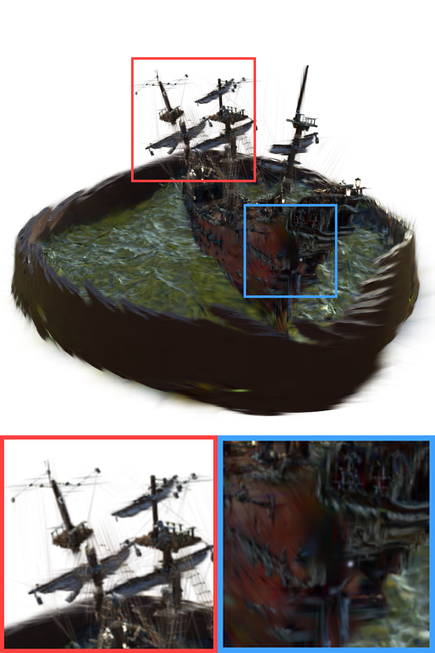}
    \\
\rotatebox[origin=c]{90}{Mani-GS}          
    &   \includegraphics[width=\hsize,valign=m]{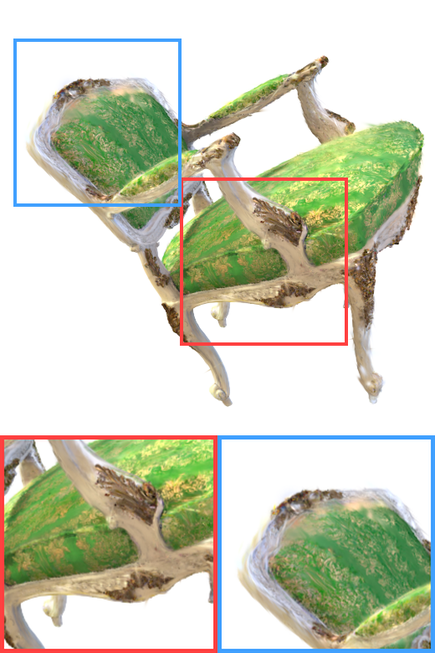}
    &   \includegraphics[width=\hsize,valign=m]{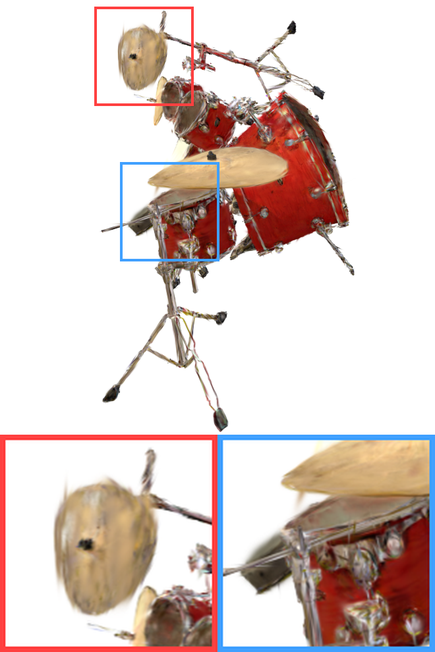}
    &   \includegraphics[width=\hsize,valign=m]{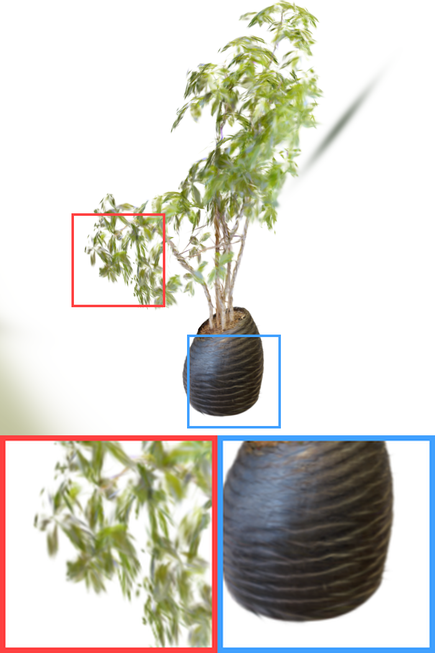}
    &   \includegraphics[width=\hsize,valign=m]{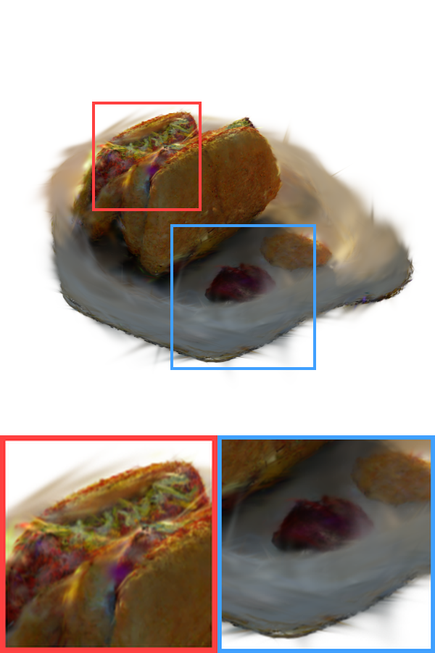}
    &   \includegraphics[width=\hsize,valign=m]{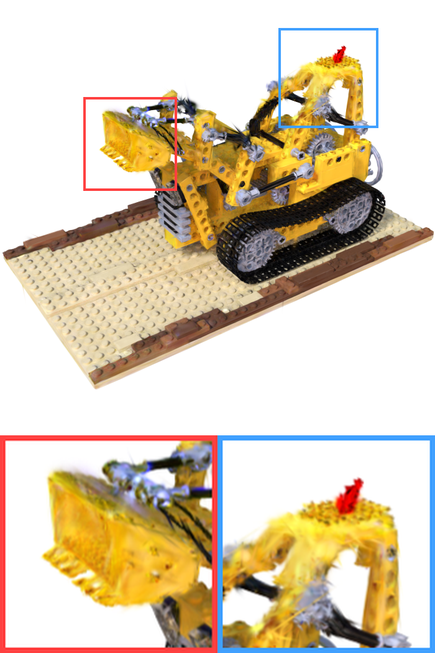}
    &   \includegraphics[width=\hsize,valign=m]{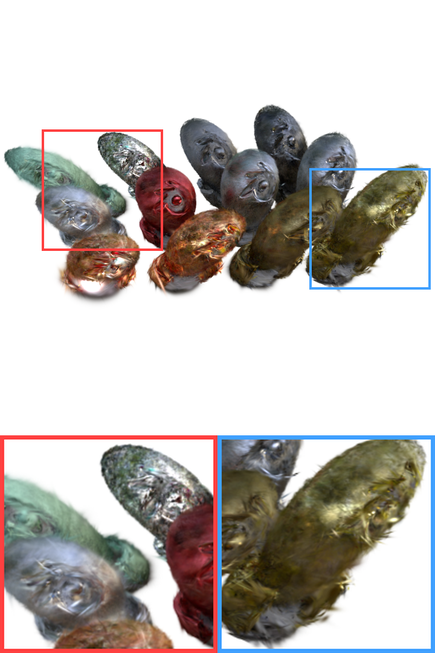}
    &   \includegraphics[width=\hsize,valign=m]{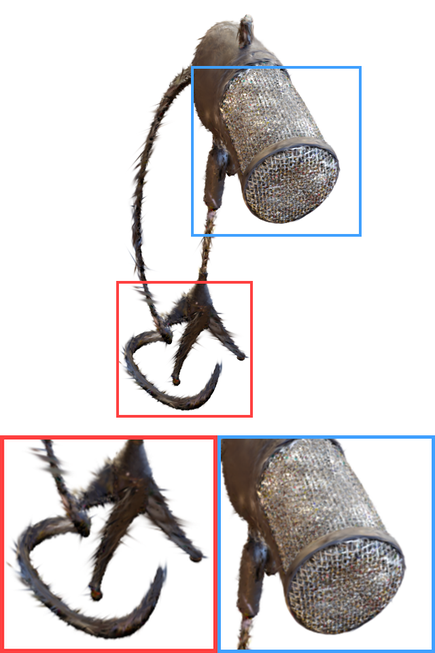}
    &   \includegraphics[width=\hsize,valign=m]{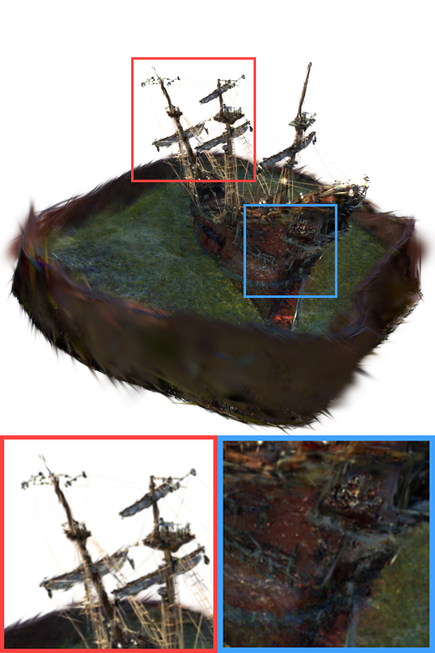}
    \\
\rotatebox[origin=c]{90}{PAPR}          
    &   \includegraphics[width=\hsize,valign=m]{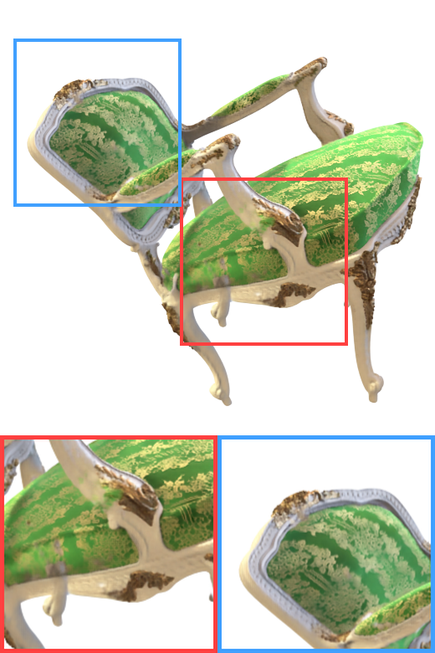}
    &   \includegraphics[width=\hsize,valign=m]{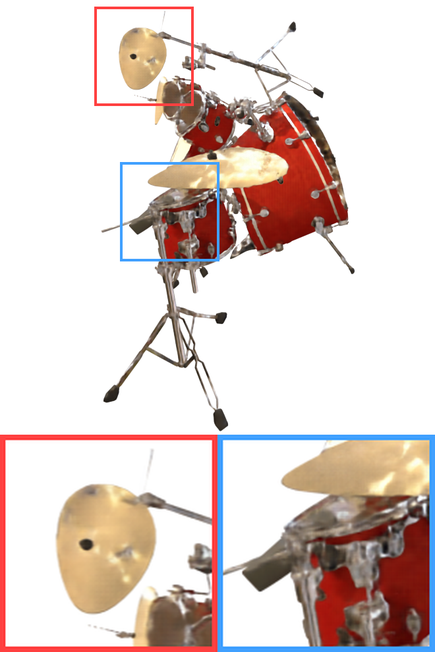}
    &   \includegraphics[width=\hsize,valign=m]{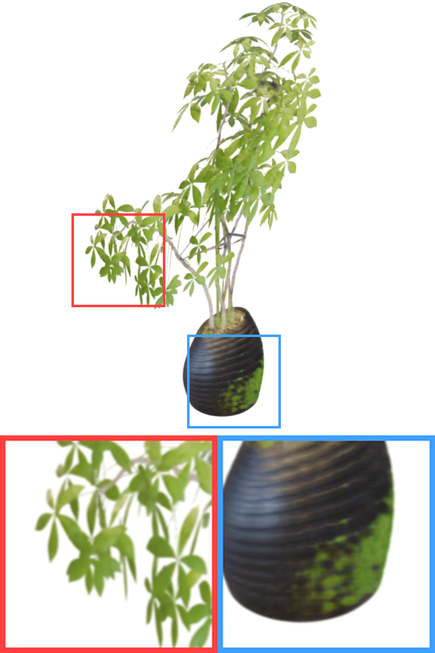}
    &   \includegraphics[width=\hsize,valign=m]{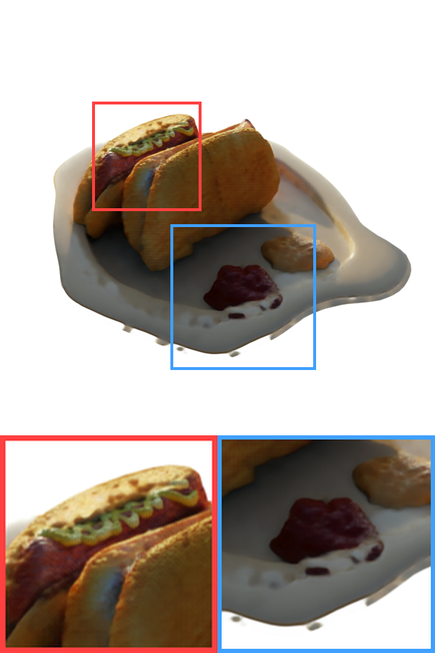}
    &   \includegraphics[width=\hsize,valign=m]{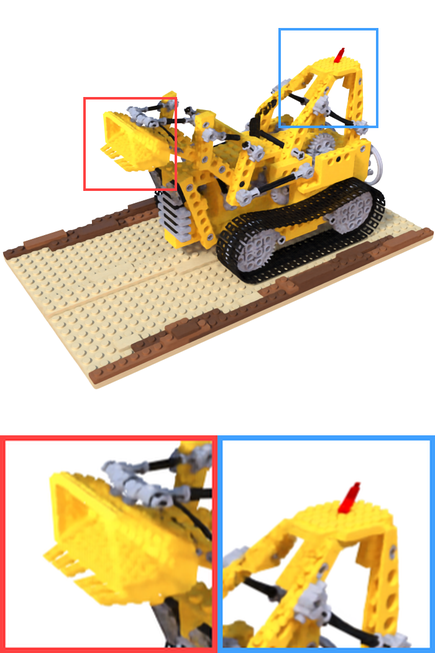}
    &   \includegraphics[width=\hsize,valign=m]{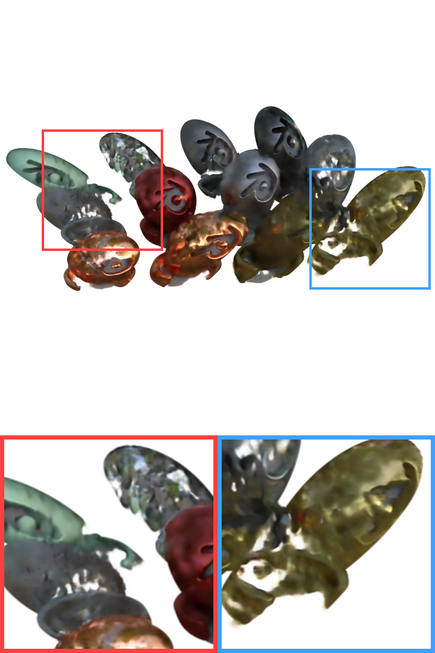}
    &   \includegraphics[width=\hsize,valign=m]{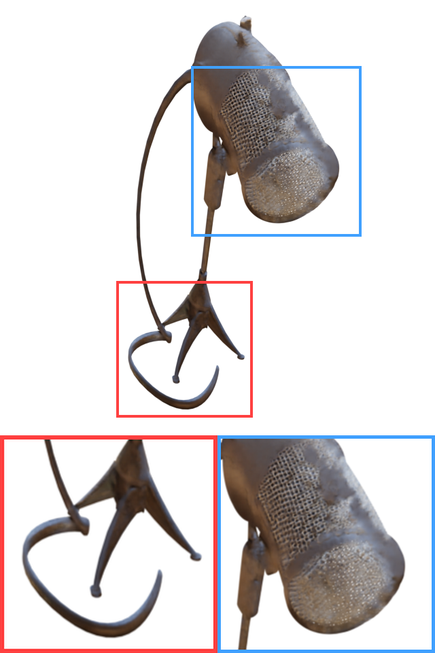}
    &   \includegraphics[width=\hsize,valign=m]{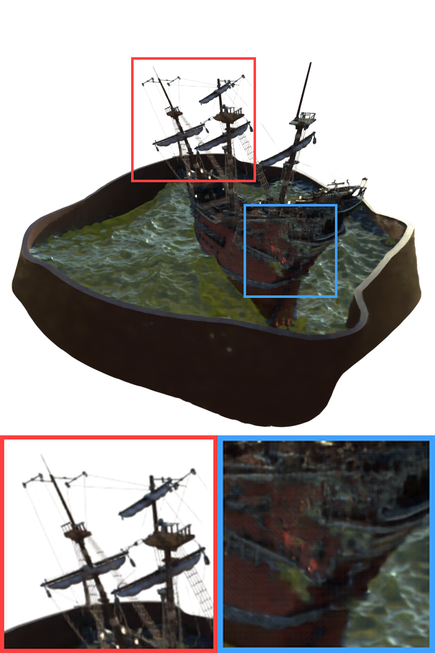}
    \\
\rotatebox[origin=c]{90}{Ours}          
    &   \includegraphics[width=\hsize,valign=m]{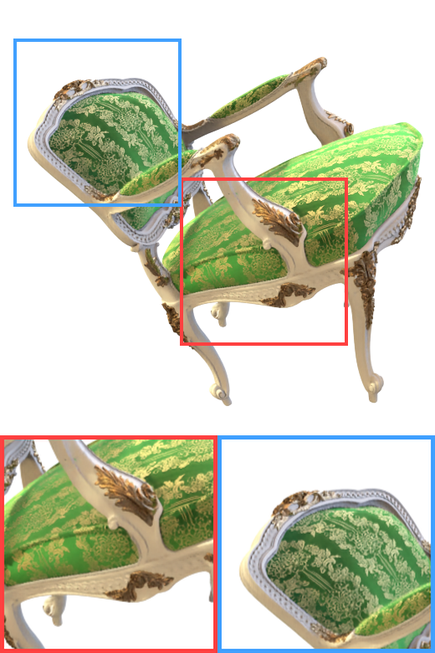}
    &   \includegraphics[width=\hsize,valign=m]{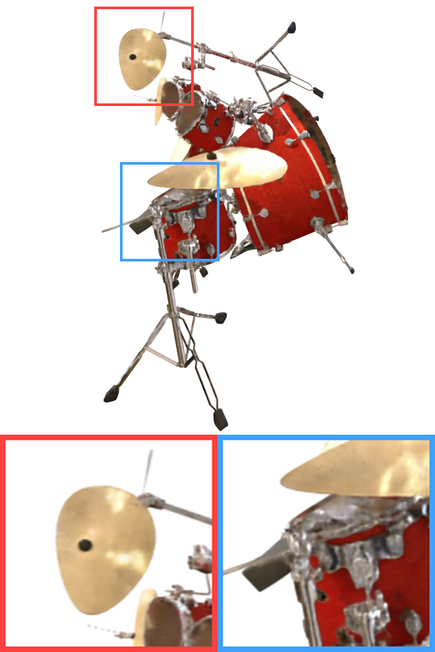}
    &   \includegraphics[width=\hsize,valign=m]{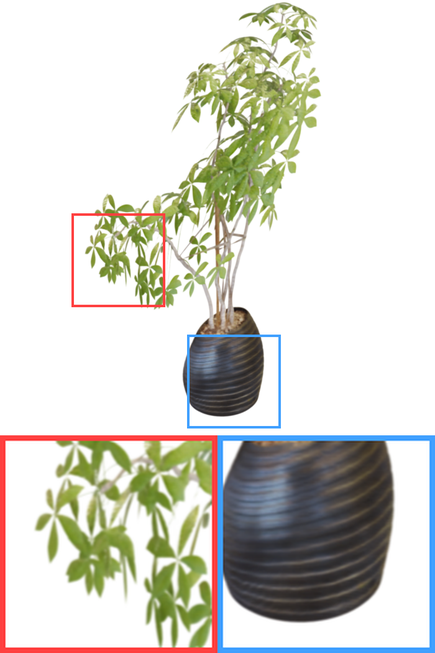}
    &   \includegraphics[width=\hsize,valign=m]{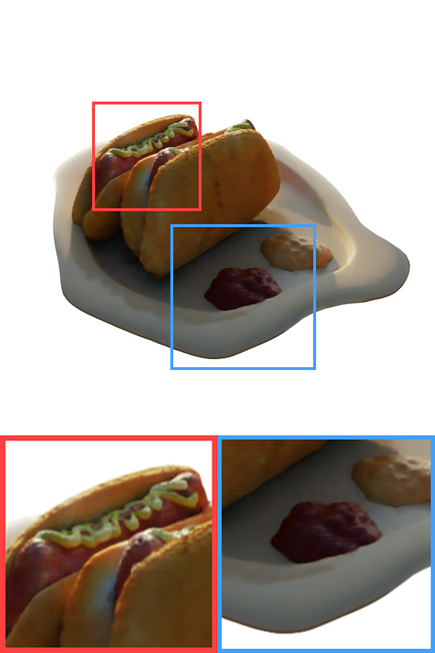}
    &   \includegraphics[width=\hsize,valign=m]{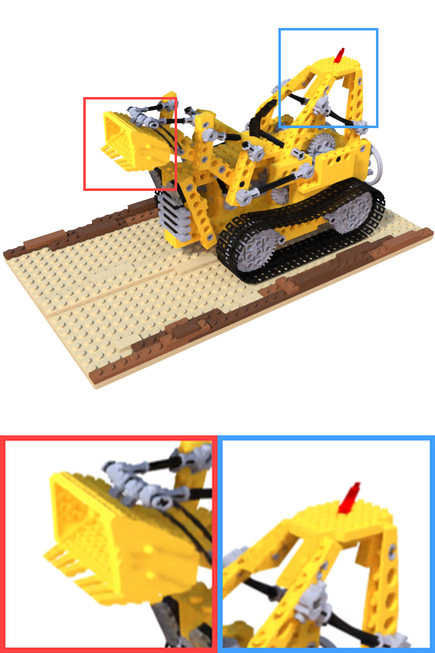}
    &   \includegraphics[width=\hsize,valign=m]{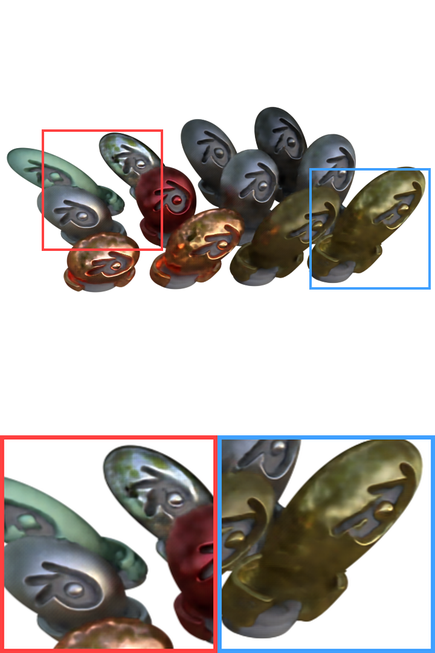}
    &   \includegraphics[width=\hsize,valign=m]{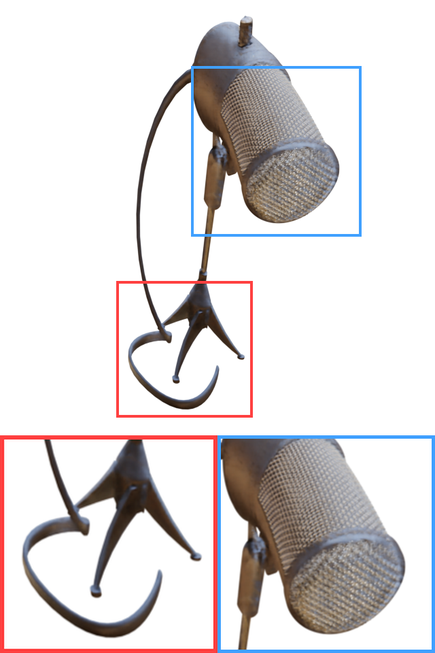}
    &   \includegraphics[width=\hsize,valign=m]{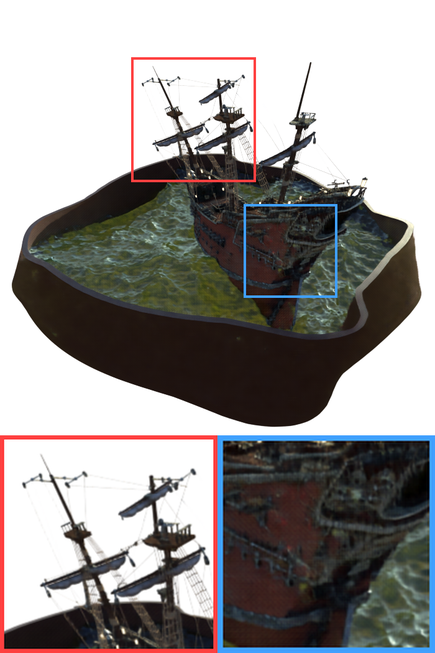}
    \\
\rotatebox[origin=c]{90}{GT}          
    &   \includegraphics[width=\hsize,valign=m]{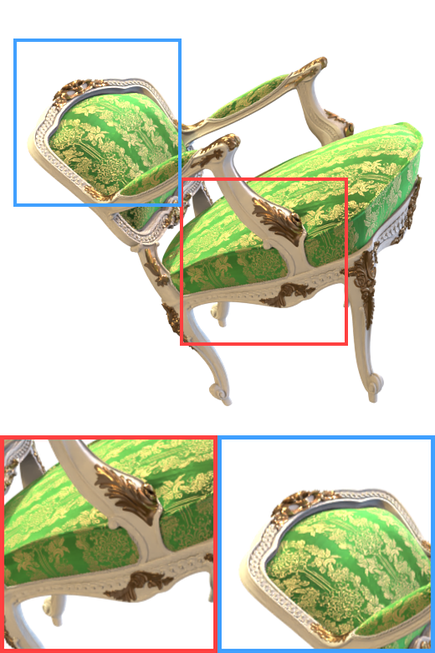}
    &   \includegraphics[width=\hsize,valign=m]{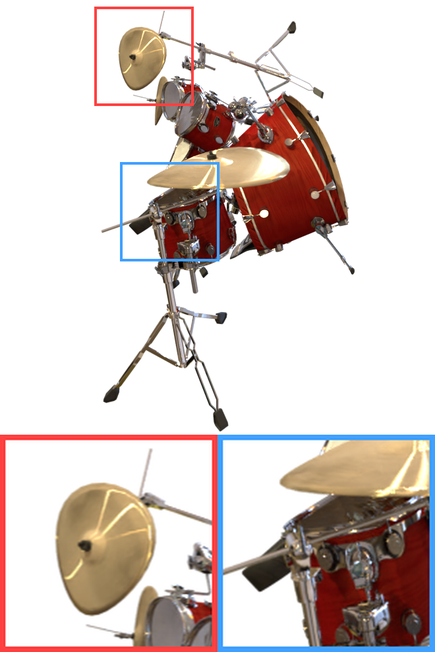}
    &   \includegraphics[width=\hsize,valign=m]{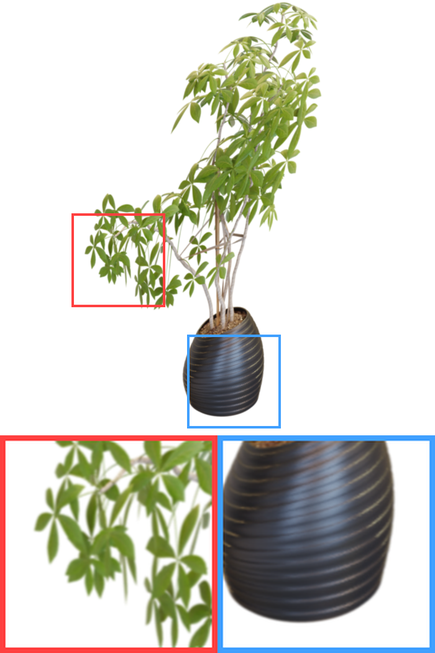}
    &   \includegraphics[width=\hsize,valign=m]{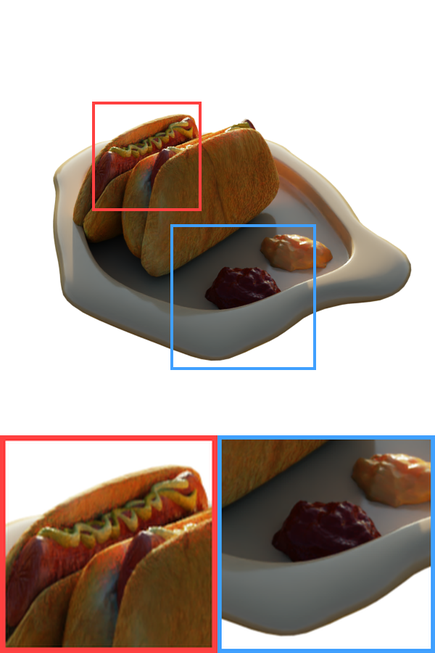}
    &   \includegraphics[width=\hsize,valign=m]{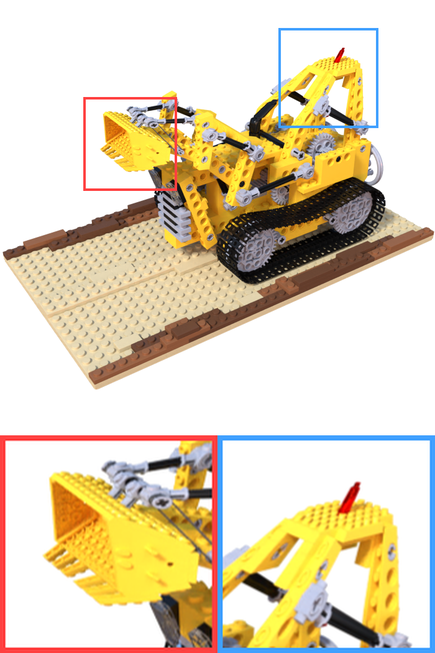}
    &   \includegraphics[width=\hsize,valign=m]{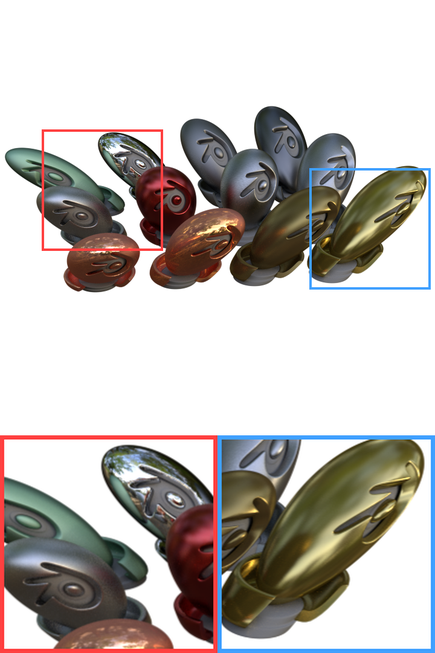}
    &   \includegraphics[width=\hsize,valign=m]{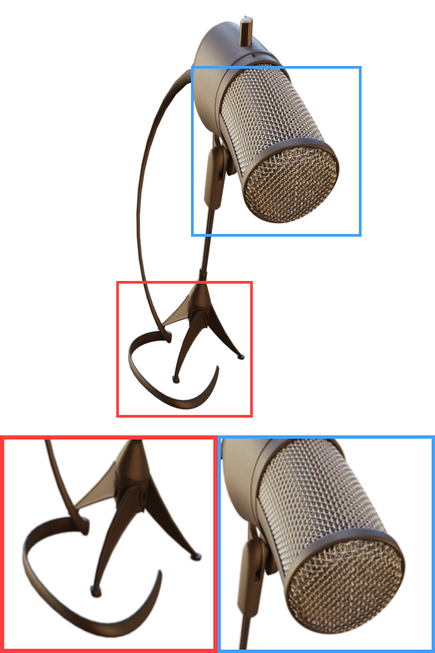}
    &   \includegraphics[width=\hsize,valign=m]{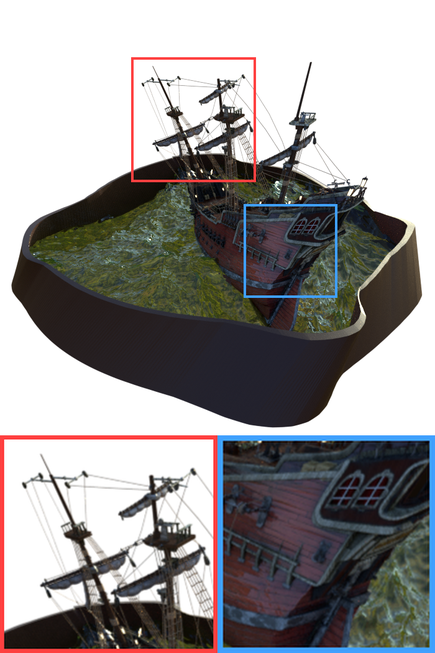}
\end{tabularx}
\caption{
    Qualitative comparison on the Neural Editor dataset.
}
\label{fig:quali-all}
\end{figure}

\subsection{Qualitative Results}

Fig.~\ref{fig:quali-all} and Fig.~\ref{fig:quali-pim} show the qualitative comparison between P-CORE and the baselines on the Neural Editor and Objaverse datasets. To align the deformations across different methods for fairness, we first manually deform the cage from Deforming-NeRF, and then use the cage to deform the mesh or the control points of other methods through cage-based deformation. 

As shown in the figures, proxy-based approaches like Mani-GS often struggle with part-level manipulations and complex topological changes. Because they are bound by a rigid mesh, they fail to separate distinct parts or introduce blocky artifacts when the underlying topology is severely stretched. In contrast, proxy-free methods like SC-GS do not rely on geometry proxies, making them more adaptable to various types of deformations. However, SC-GS severely fails to preserve surface continuity after editing, leading to pervasive holes and ``extruded Gaussians'' around the boundaries of the edited shapes. Furthermore, while the base PAPR model handles minor interpolations well, it still exhibits noticeable tearing and structural collapse under large non-rigid edits. 

In contrast, our P-CORE framework enhances structural robustness. It significantly reduces holes, tearing, and other artifacts, consistently preserving perfectly continuous surfaces across all scenes even under severe deformation.

\begin{figure}[htb]
\centering
\begin{subfigure}[t]{0.55\linewidth}
    \centering
    \setlength\tabcolsep{1.5pt}
    \scriptsize
    \begin{tabular}[t]{cccc}
        Before Edit & Edit 1 & Edit 2 & Edit 3 \\
        \includegraphics[width=0.24\linewidth]{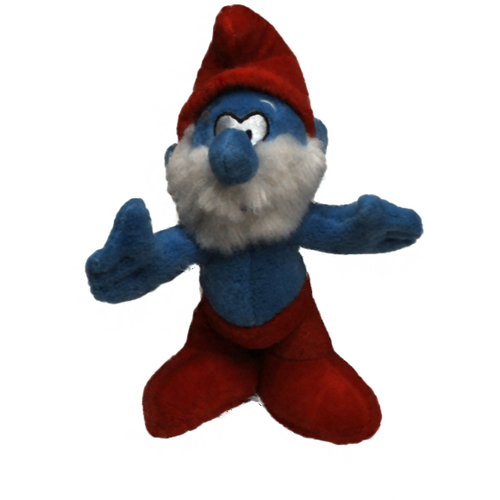} &
        \includegraphics[width=0.24\linewidth]{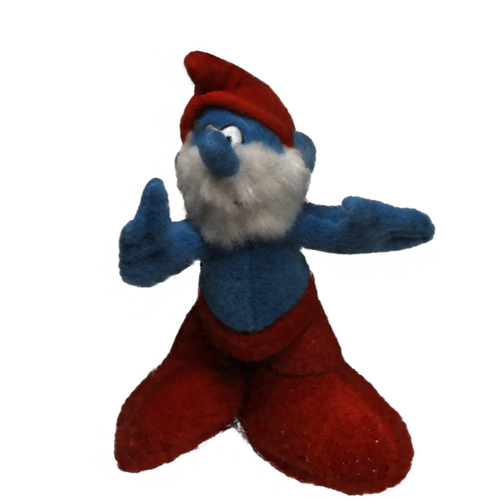} &
        \includegraphics[width=0.24\linewidth]{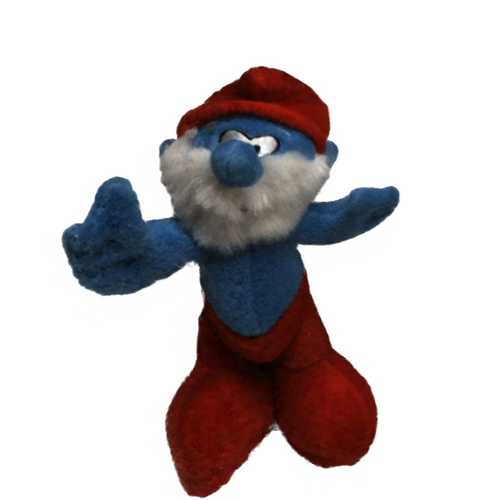} &
        \includegraphics[width=0.24\linewidth]{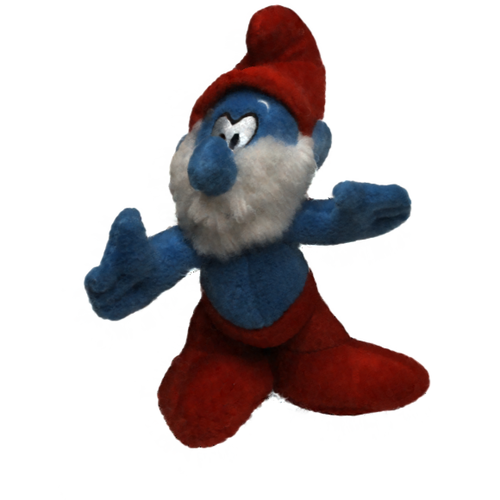} \\
        \includegraphics[width=0.24\linewidth]{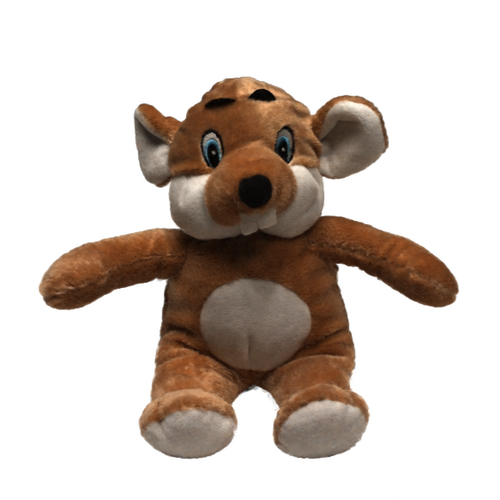} &
        \includegraphics[width=0.24\linewidth]{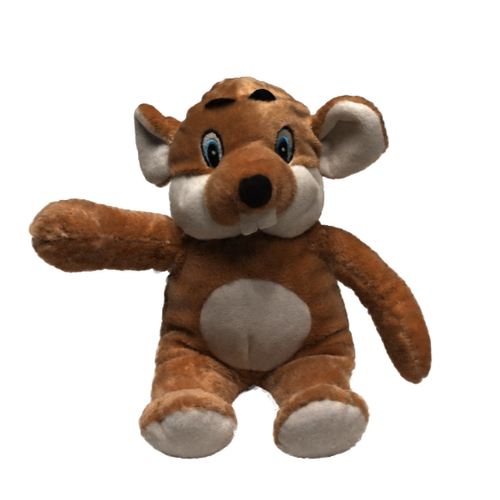} &
        \includegraphics[width=0.24\linewidth]{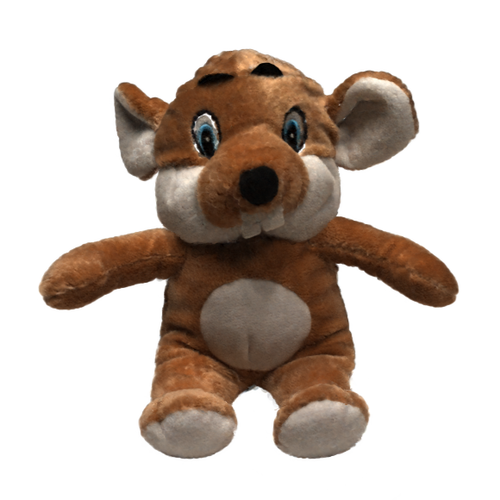} &
        \includegraphics[width=0.24\linewidth]{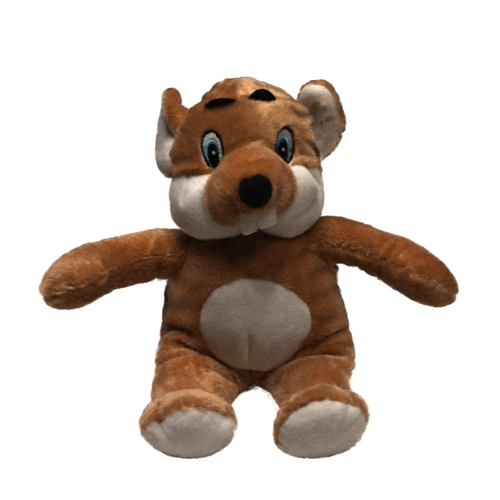} \\
    \end{tabular}
    \caption{DTU dataset}
    \label{fig:real-world-dtu}
\end{subfigure}
\hfill
\begin{subfigure}[t]{0.412\linewidth}
    \centering
    \setlength\tabcolsep{1.5pt}
    \scriptsize
    \begin{tabular}[t]{cc}
        Before Edit & After Edit \\
        \includegraphics[width=0.48\linewidth]{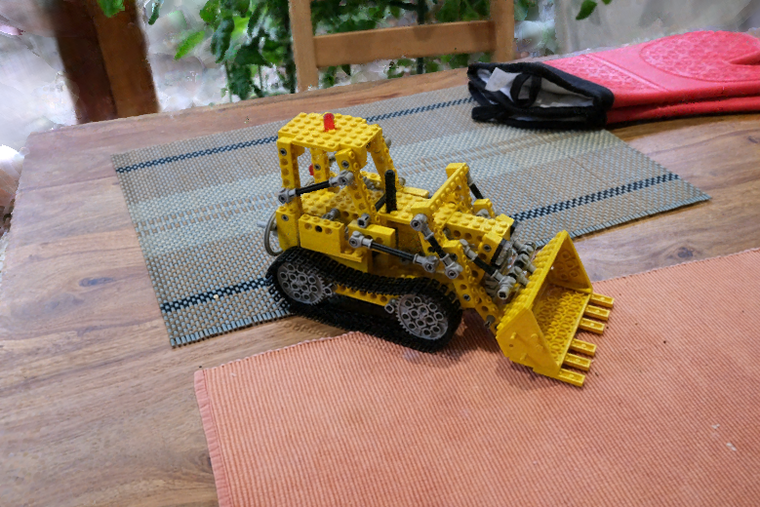} &
        \includegraphics[width=0.48\linewidth]{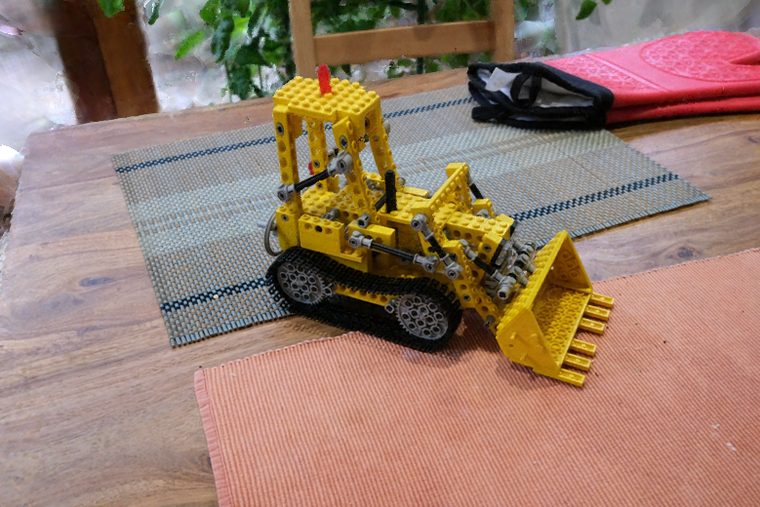} \\
        \includegraphics[width=0.48\linewidth]{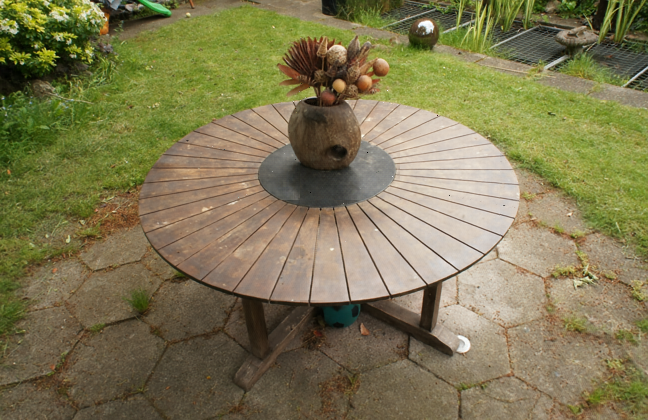} &
        \includegraphics[width=0.48\linewidth]{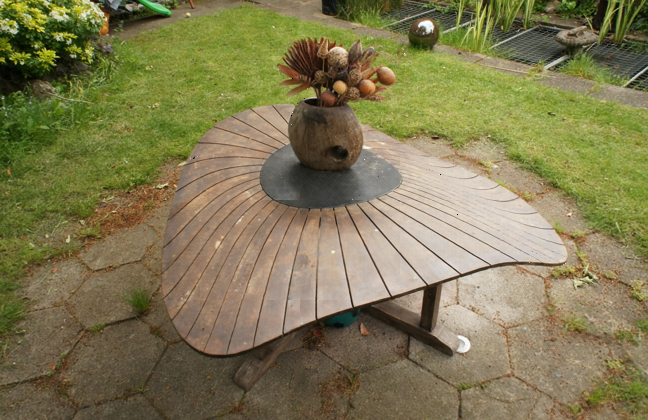} \\
    \end{tabular}
    \caption{Mip-NeRF 360 dataset}
    \label{fig:real-world-mipnerf360}
\end{subfigure}
\caption{
    Editing results on real-world scenes from the DTU and Mip-NeRF 360 datasets.
}
\label{fig:quali-real-world}
\end{figure}

Additionally, we demonstrate the versatility of our method on complex real-world captures. As shown in Fig.~\ref{fig:quali-real-world}, P-CORE extends robustly to the DTU and Mip-NeRF 360 datasets, supporting multiple sequential edits without degrading texture quality or introducing surface tearing.

\begin{table}[htb]
\centering
\setlength{\tabcolsep}{6pt}
\caption{Average quantitative performance on the Neural Editor and Objaverse datasets across all scenes. \textbf{Bold} indicates the best and \underline{underline} indicates the second best.}
\label{tab:quant-mean}
\resizebox{1\linewidth}{!}{
\begin{tabular}{l|cccc|cccc}
\toprule
\multirow{2}{*}{Method} & \multicolumn{4}{c|}{Neural Editor~\cite{Chen2023NeuralEditorEN}} & \multicolumn{4}{c}{Objaverse~\cite{Deitke2022ObjaverseAU}} \\
\cmidrule(lr){2-5} \cmidrule(lr){6-9}
& PSNR $\uparrow$ & SSIM $\uparrow$ & LPIPS $\downarrow$ & \# Primitives & PSNR $\uparrow$ & SSIM $\uparrow$ & LPIPS $\downarrow$ & \# Primitives \\
\midrule
Deforming-NeRF~\cite{xu2022cagedeforming} & 14.71 & 0.722 & 0.197 & - & 15.89 & 0.773 & 0.180 & - \\
Neural Editor~\cite{Chen2023NeuralEditorEN} & \textbf{25.62} & \underline{0.939} & 0.078 & $\sim$1M & \underline{28.02} & \underline{0.956} & \underline{0.048} & $\sim$1M \\
SC-GS~\cite{Huang2023SCGSSG} & 18.55 & 0.832 & 0.123 & $\sim$236K & 25.29 & 0.929 & 0.057 & $\sim$228K \\
Mani-GS~\cite{Gao2024ManiGSGS} & 20.59 & 0.875 & 0.116 & $\sim$1.2M & 26.04 & 0.946 & 0.069 & $\sim$362K \\
PAPR~\cite{zhang2023papr} & 24.22 & 0.930 & \underline{0.066} & 30K & 23.15 & 0.940 & 0.058 & 30K \\
Ours & \underline{25.31} & \textbf{0.942} & \textbf{0.054} & 30K & \textbf{28.30} & \textbf{0.965} & \textbf{0.039} & 30K \\
\bottomrule
\end{tabular}
}
\end{table}

\begin{figure}[htb]
\setlength\tabcolsep{4pt}
\tiny
\centering
\resizebox{1\linewidth}{!}{
\begin{tabularx}{1\linewidth}{l@{\hskip 1em}YYYYYY}
    &   Butterfly
    &   Crab
    &   Dolphin
    &   Giraffe
    &   Lego
    &   Legoman
    \\
\rotatebox[origin=c]{90}{SC-GS}          
    &   \includegraphics[width=\hsize,valign=m]{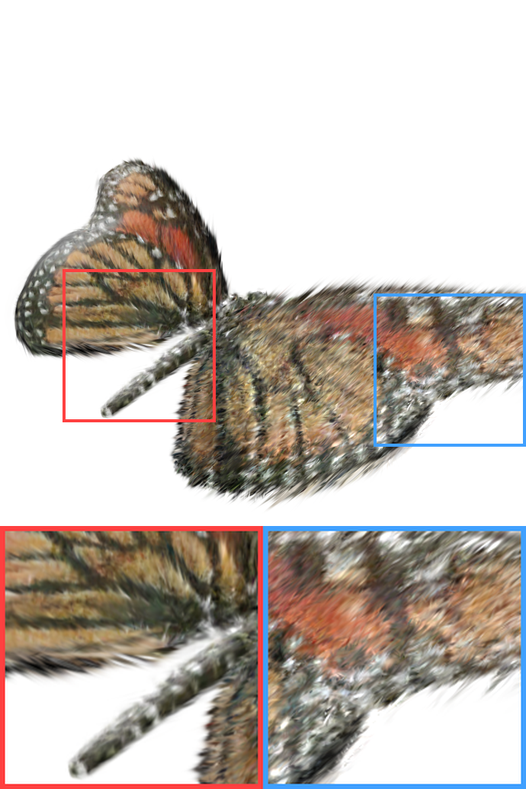}
    &   \includegraphics[width=\hsize,valign=m]{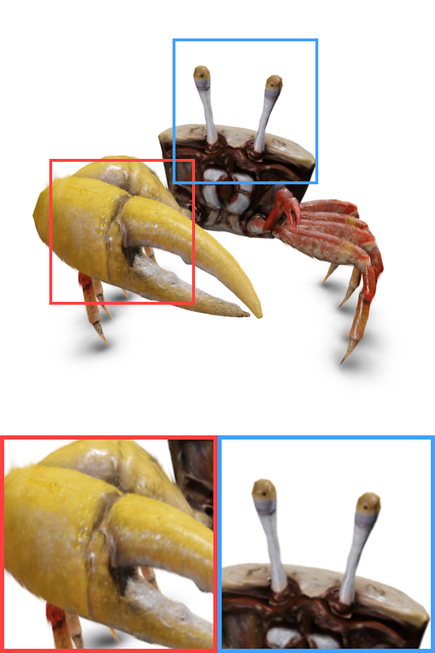}
    &   \includegraphics[width=\hsize,valign=m]{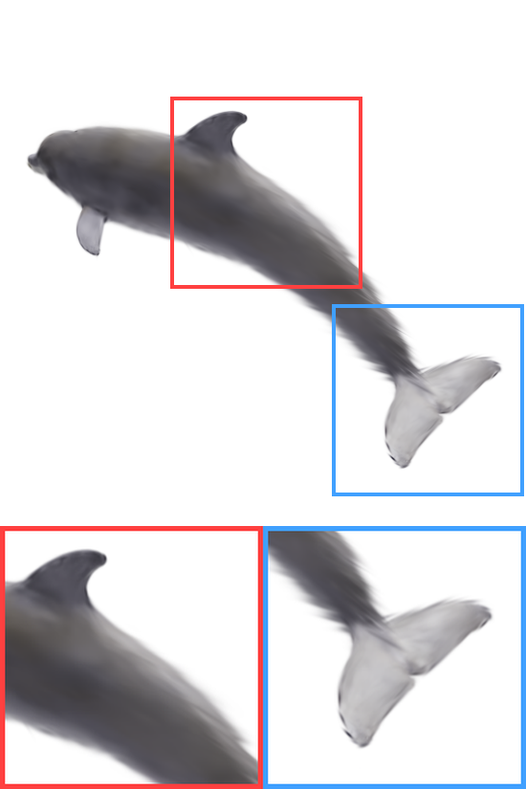}
    &   \includegraphics[width=\hsize,valign=m]{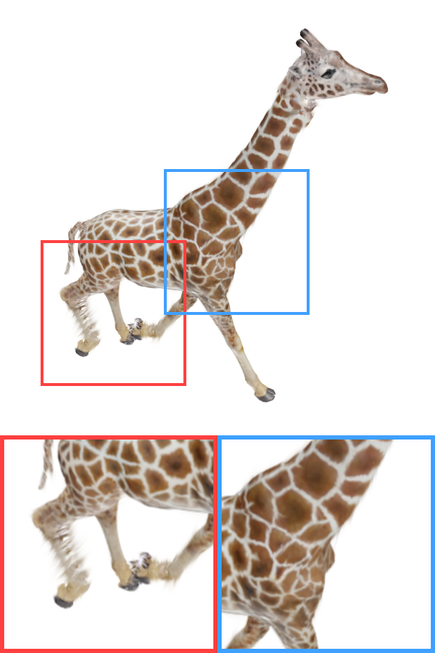}
    &   \includegraphics[width=\hsize,valign=m]{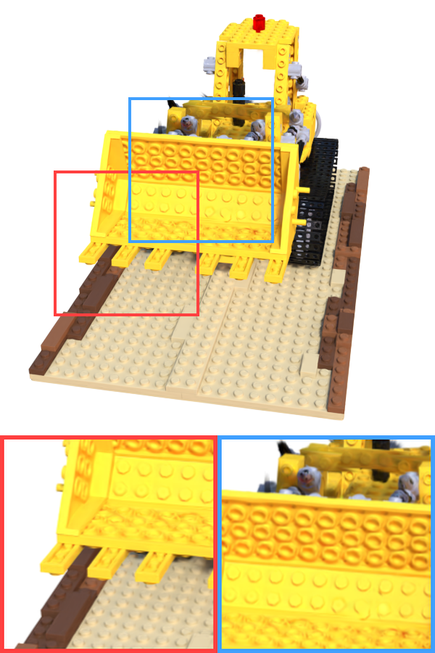}
    &   \includegraphics[width=\hsize,valign=m]{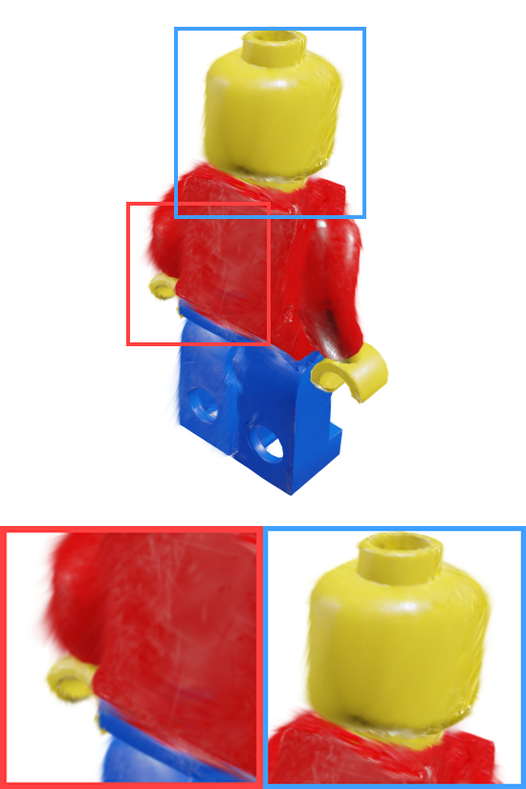}
    \\
\rotatebox[origin=c]{90}{Mani-GS}          
    &   \includegraphics[width=\hsize,valign=m]{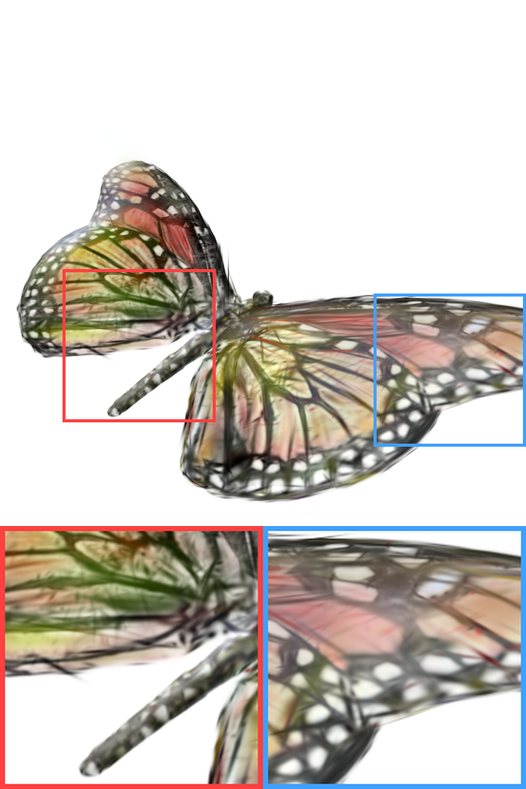}
    &   \includegraphics[width=\hsize,valign=m]{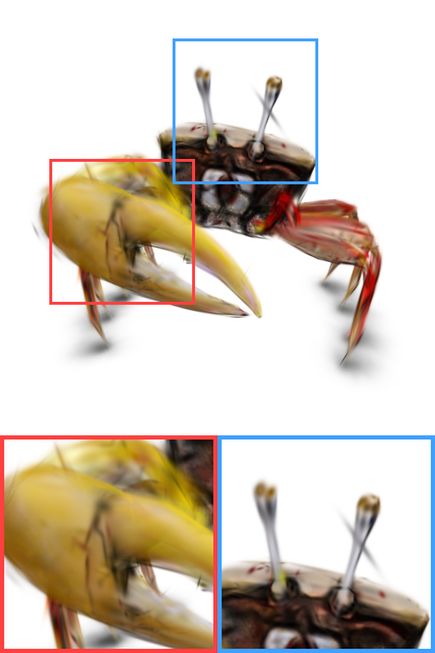}
    &   \includegraphics[width=\hsize,valign=m]{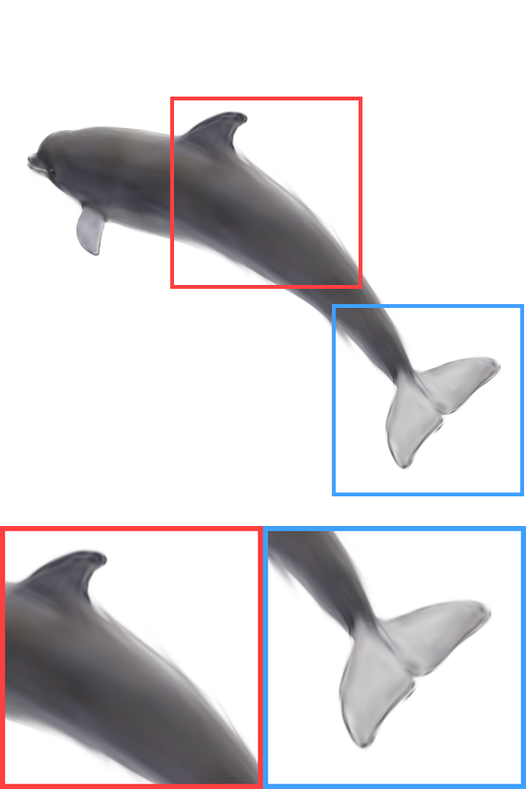}
    &   \includegraphics[width=\hsize,valign=m]{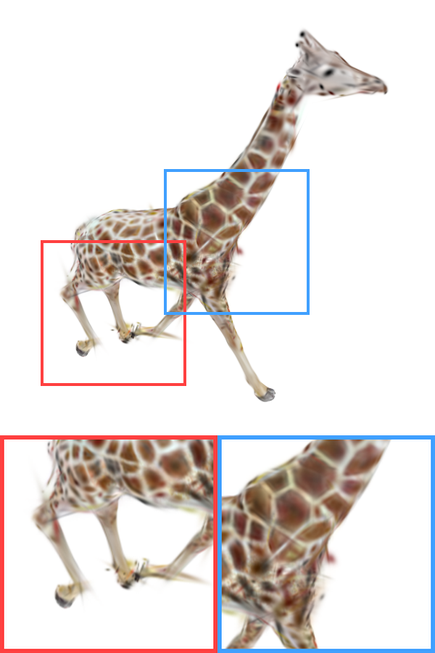}
    &   \includegraphics[width=\hsize,valign=m]{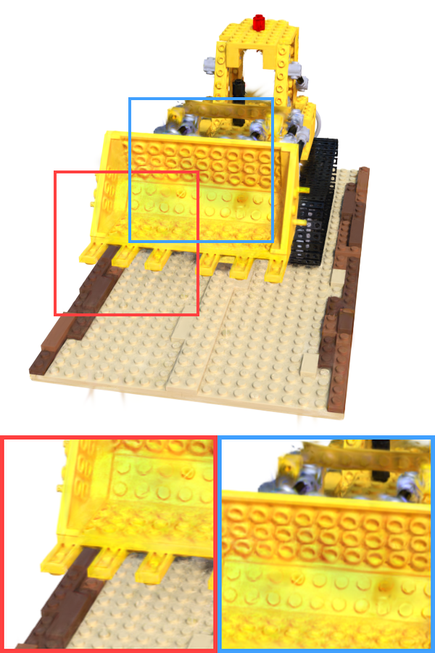}
    &   \includegraphics[width=\hsize,valign=m]{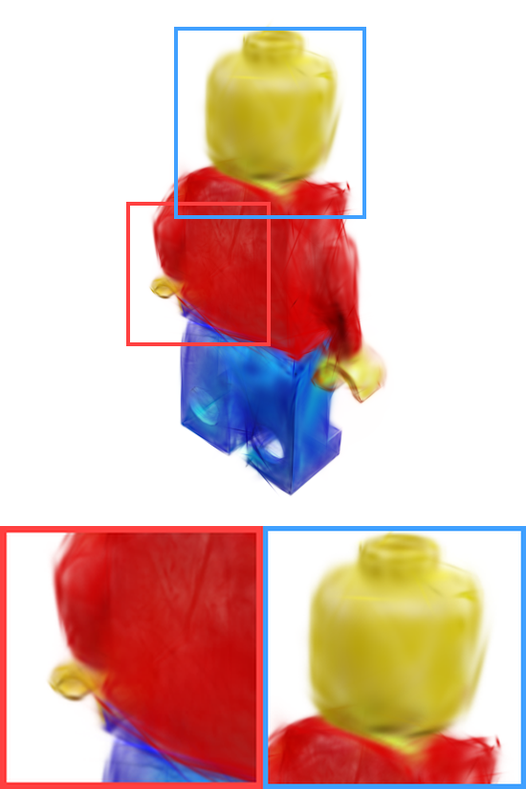}
    \\
\rotatebox[origin=c]{90}{PAPR}          
    &   \includegraphics[width=\hsize,valign=m]{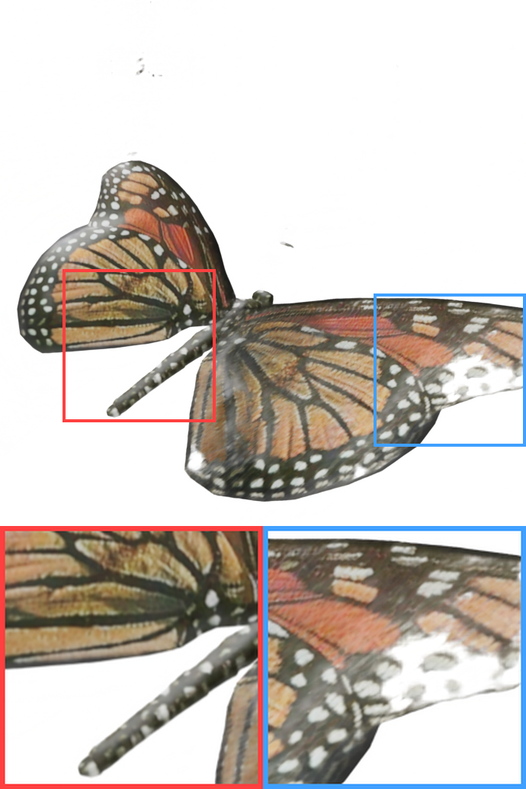}
    &   \includegraphics[width=\hsize,valign=m]{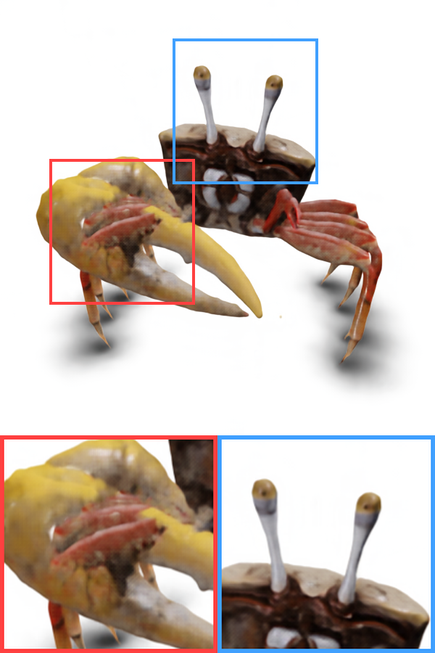}
    &   \includegraphics[width=\hsize,valign=m]{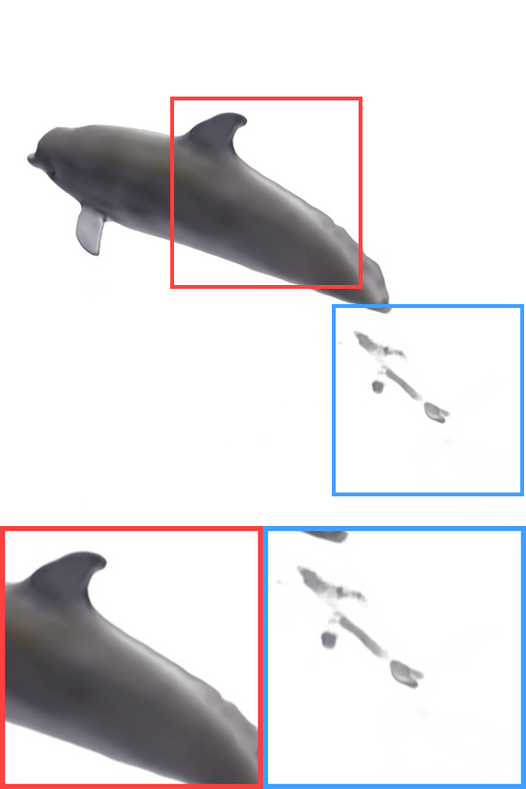}
    &   \includegraphics[width=\hsize,valign=m]{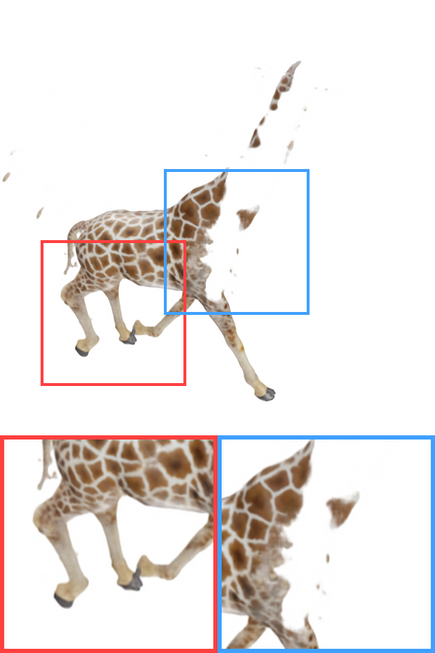}
    &   \includegraphics[width=\hsize,valign=m]{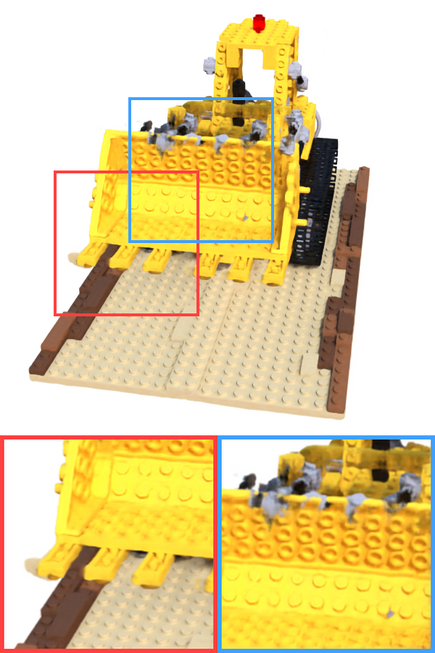}
    &   \includegraphics[width=\hsize,valign=m]{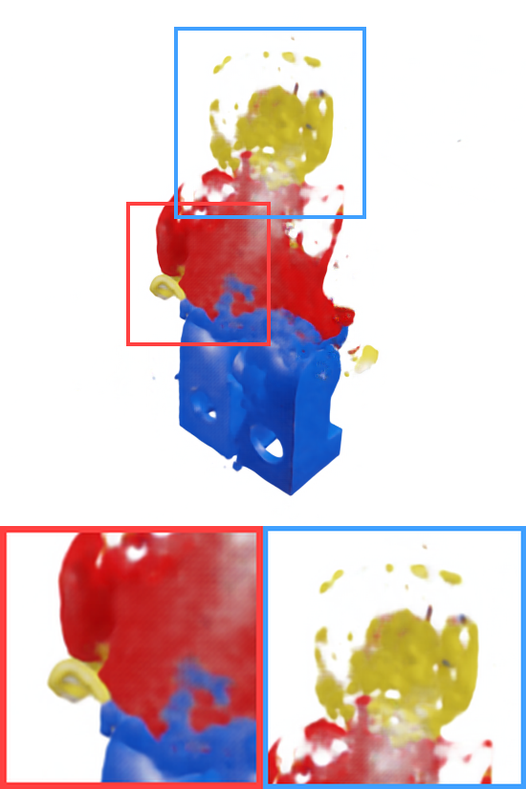}
    \\
\rotatebox[origin=c]{90}{Ours}          
    &   \includegraphics[width=\hsize,valign=m]{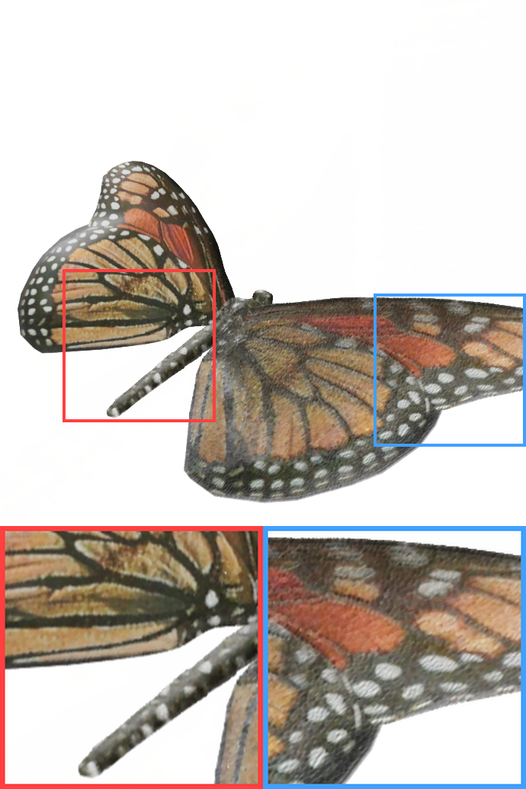}
    &   \includegraphics[width=\hsize,valign=m]{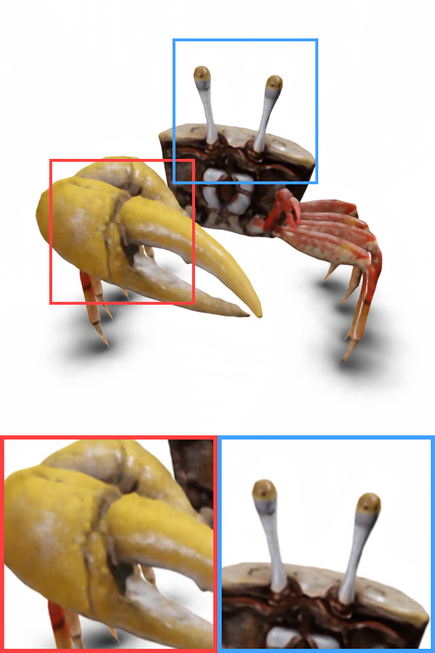}
    &   \includegraphics[width=\hsize,valign=m]{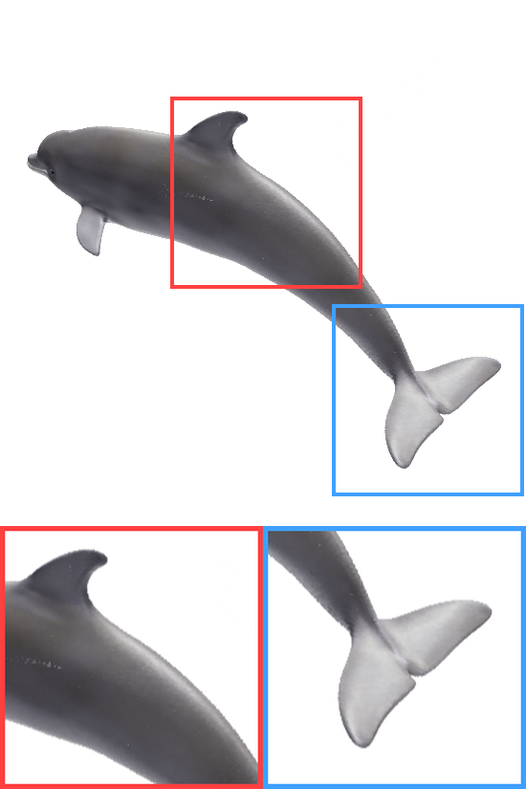}
    &   \includegraphics[width=\hsize,valign=m]{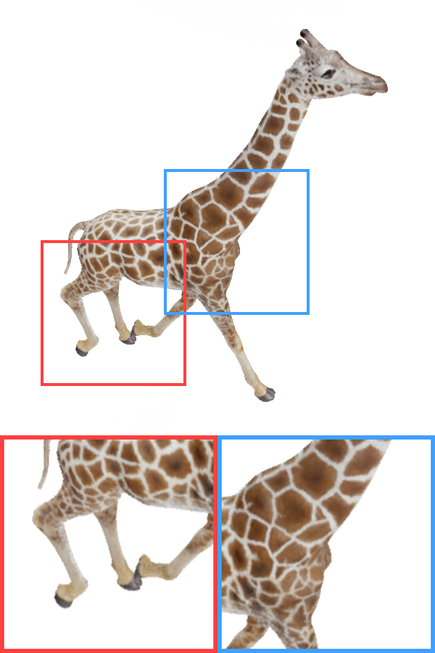}
    &   \includegraphics[width=\hsize,valign=m]{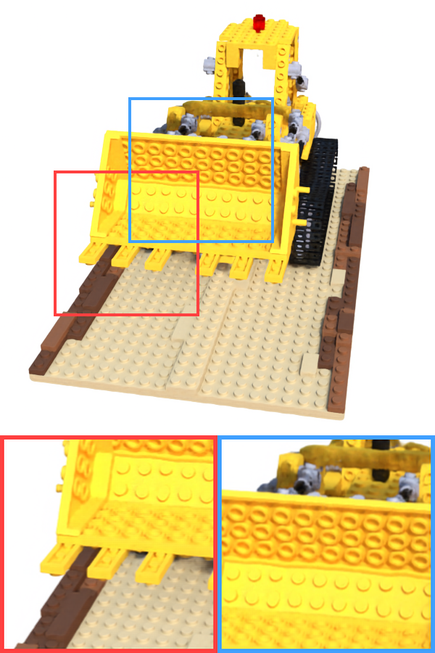}
    &   \includegraphics[width=\hsize,valign=m]{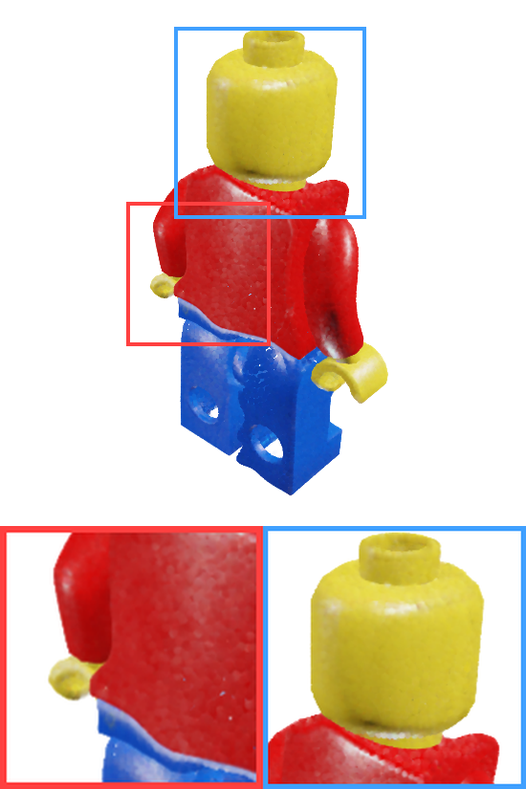}
    \\
\rotatebox[origin=c]{90}{GT}          
    &   \includegraphics[width=\hsize,valign=m]{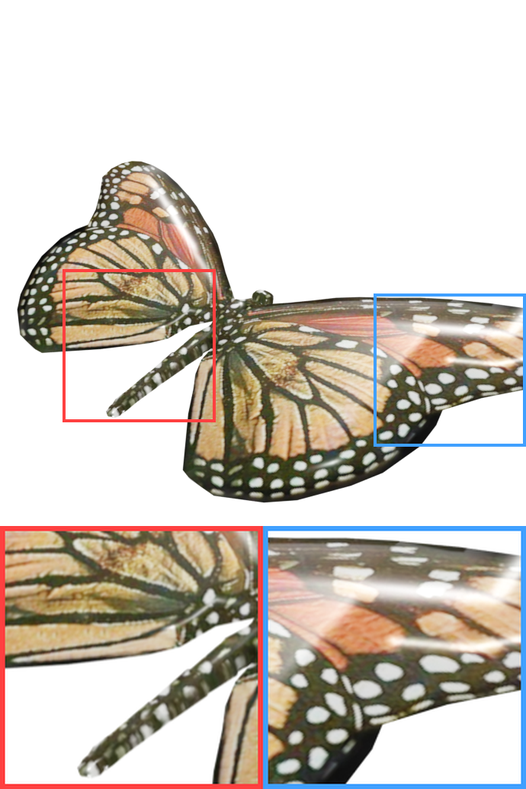}
    &   \includegraphics[width=\hsize,valign=m]{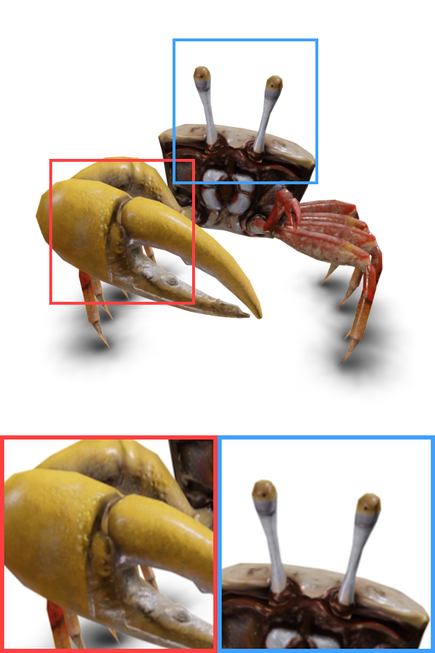}
    &   \includegraphics[width=\hsize,valign=m]{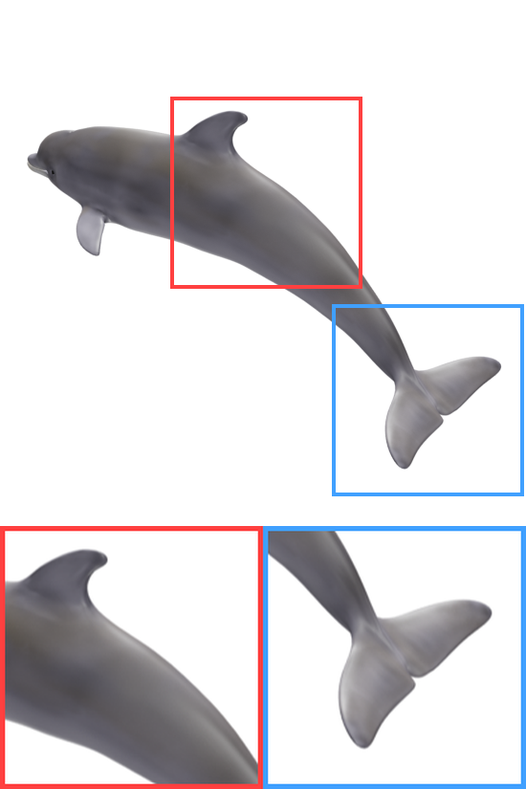}
    &   \includegraphics[width=\hsize,valign=m]{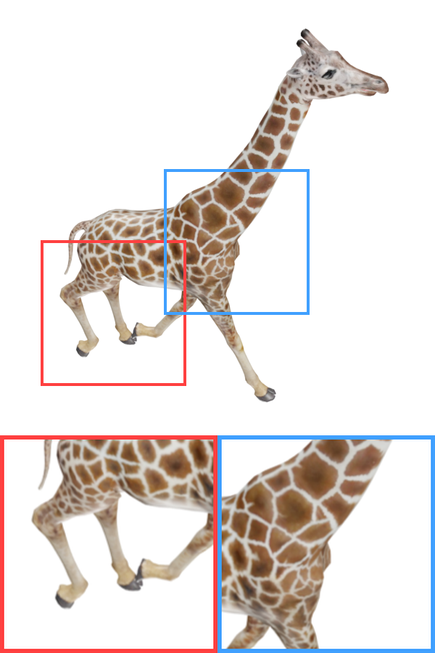}
    &   \includegraphics[width=\hsize,valign=m]{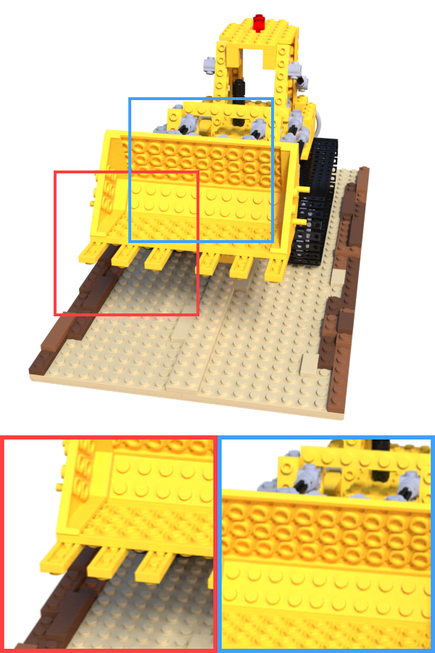}
    &   \includegraphics[width=\hsize,valign=m]{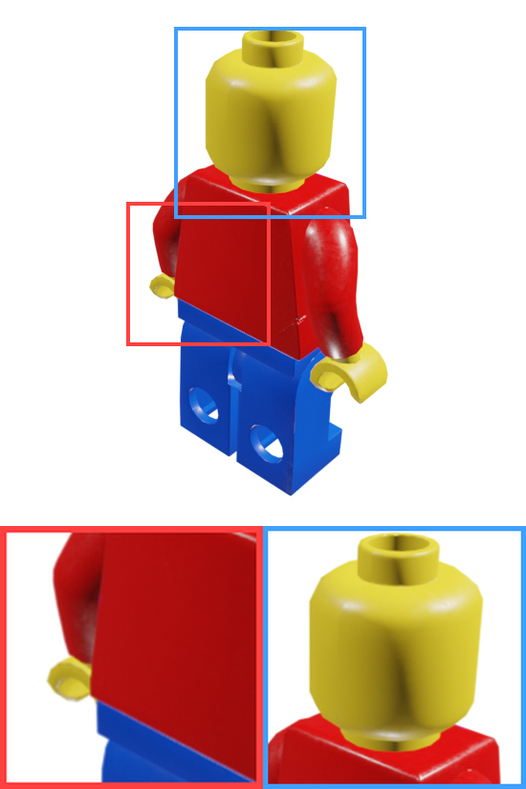}
\end{tabularx}
}
\caption{
    Qualitative comparison on scenes from Objaverse.
}
\label{fig:quali-pim}
\end{figure}

\subsection{Quantitative Results}

To quantitatively evaluate our zero-shot editing performance, we compare our method against all baselines on scenes from the Neural Editor~\cite{Chen2023NeuralEditorEN} and Objaverse~\cite{Deitke2022ObjaverseAU} datasets where ground truth deformed states are available. By applying identical deformation fields to the points or proxies, we ensure a fair comparison across all methods.

Table~\ref{tab:quant-mean} summarizes the average quantitative performance. P-CORE achieves state-of-the-art rendering quality across all metrics (PSNR, SSIM, and LPIPS) on the Objaverse dataset, and either matches or outperforms the baselines on the Neural Editor dataset, with a much smaller number of primitives (30K vs. 1M). Although Neural Editor attains strong metrics, it relies on $\sim$1M points and is far more expensive, requiring $\sim$7 days of training per scene versus $\sim$1 minute of fine-tuning for our method; we provide side-by-side qualitative comparisons with Neural Editor in the supplementary material.

\subsection{User Study}

To directly assess the geometric artifacts that pixel-level metrics (PSNR/SSIM/LPIPS) cannot fully capture, we conduct a two-alternative forced-choice (2AFC) user study. We collected $525$ responses from $25$ participants, who compared pairs of deformed renderings across all scenes against the three strongest baselines (PAPR~\cite{zhang2023papr}, Mani-GS~\cite{Gao2024ManiGSGS}, and SC-GS~\cite{Huang2023SCGSSG}) and were asked to select the result with \emph{fewer geometric artifacts such as holes, tearing, and broken topology}. As shown in Table~\ref{tab:userstudy}, our method was preferred in $94.7\%$ of trials overall and in $>89\%$ of every baseline--dataset pair, confirming that P-CORE is consistently judged to produce fewer geometric artifacts than all baselines.

\begin{table}[htb]
\centering
\setlength{\tabcolsep}{8pt}
\caption{User-study win rate of our method (\%, with raw counts). On each trial, participants selected the deformed rendering with fewer geometric artifacts (holes, tearing, broken topology).}
\label{tab:userstudy}
\begin{tabular}{lccc}
\toprule
Pairing & Neural Editor & Objaverse & All \\
\midrule
Ours vs.\ PAPR      & $90.2\%$ ($83/92$)   & $95.3\%$ ($82/86$)  & $92.7\%$ ($165/178$) \\
Ours vs.\ Mani-GS  & $98.0\%$ ($100/102$) & $94.9\%$ ($75/79$)  & $96.7\%$ ($175/181$) \\
Ours vs.\ SC-GS   & $98.9\%$ ($88/89$)   & $89.6\%$ ($69/77$)  & $94.6\%$ ($157/166$) \\
\midrule
\textbf{Overall}                       & $\mathbf{95.8\%}$ ($271/283$) & $\mathbf{93.4\%}$ ($226/242$) & $\mathbf{94.7\%}$ ($497/525$) \\
\bottomrule
\end{tabular}
\end{table}

\subsection{Ablation Study}

\begin{figure}[htb]
\centering
\setlength\tabcolsep{1.5pt}
\scriptsize
\begin{tabularx}{1\linewidth}{lYYYYYY}
    & Canonical & PAPR & w/o R\&P & w/o Pt-Only & w/o $\mathcal{L}_{\text{tex}}$ & Full \\
\rotatebox[origin=c]{90}{Rendering} &
    \includegraphics[width=\hsize,valign=m]{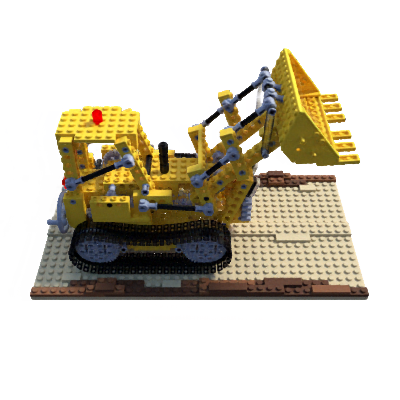} &
    \includegraphics[width=\hsize,valign=m]{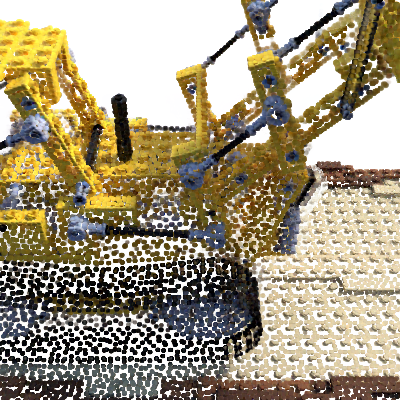} &
    \includegraphics[width=\hsize,valign=m]{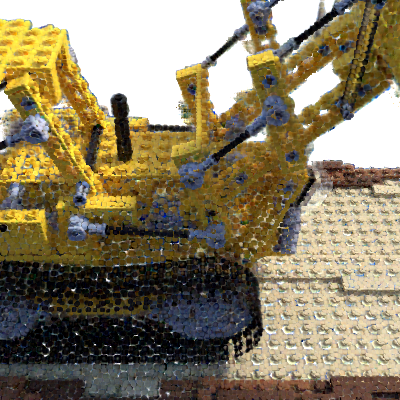} &
    \includegraphics[width=\hsize,valign=m]{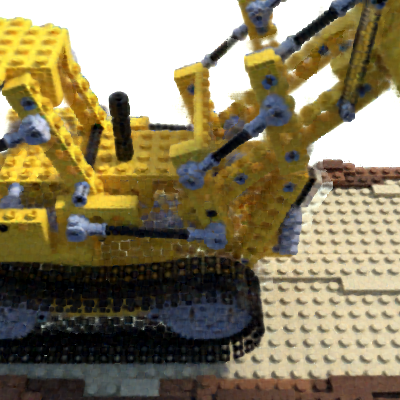} &
    \includegraphics[width=\hsize,valign=m]{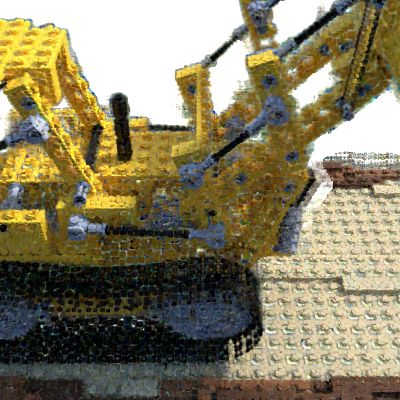} &
    \includegraphics[width=\hsize,valign=m]{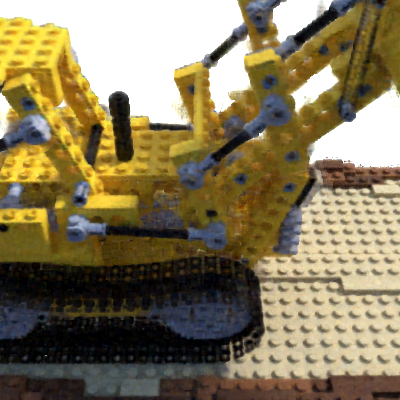} \\
\rotatebox[origin=c]{90}{Depth Map} &
    \includegraphics[width=\hsize,valign=m]{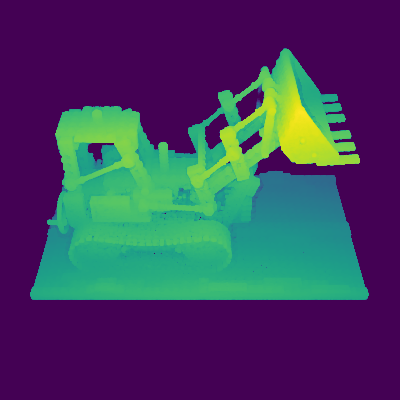} &
    \includegraphics[width=\hsize,valign=m]{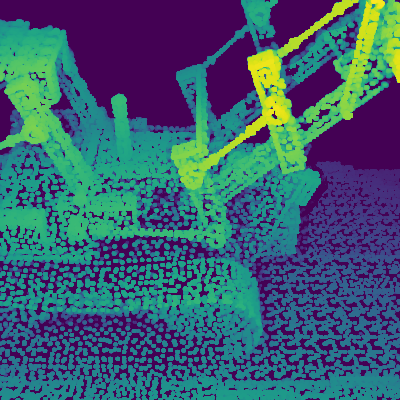} &
    \includegraphics[width=\hsize,valign=m]{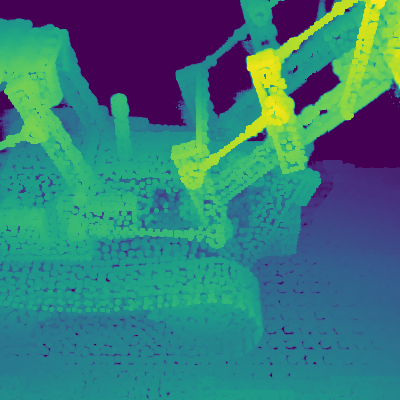} &
    \includegraphics[width=\hsize,valign=m]{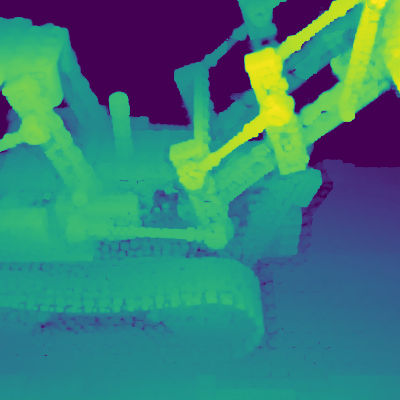} &
    \includegraphics[width=\hsize,valign=m]{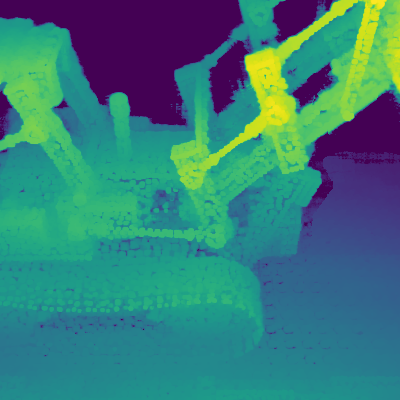} &
    \includegraphics[width=\hsize,valign=m]{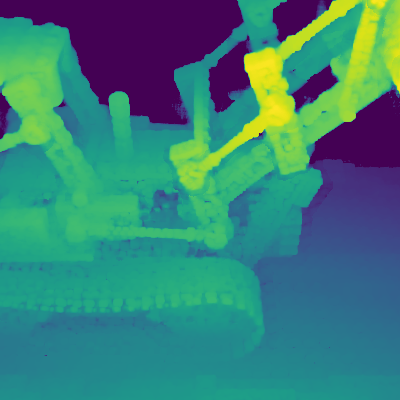} \\
\end{tabularx}
\caption{Ablation study on the Lego scene under $2\times$ scaling. From left to right: canonical (undeformed), vanilla PAPR, without rays-and-points (R\&P) mode, without points-only mode, without texture loss $\mathcal{L}_{\text{tex}}$, and our full pipeline.}
\label{fig:ablation}
\end{figure}

To evaluate the contribution of each component in our self-supervised fine-tuning framework, we conduct an ablation study on the Lego scene under a $2\times$ uniform scaling deformation. Fig.~\ref{fig:ablation} provides a qualitative comparison of the rendered novel views and depth maps across different configurations.

\textbf{Vanilla PAPR} (without fine-tuning) exhibits obvious holes in both the novel views and depth maps, confirming the base representation's inability to maintain surface continuity under large deformations. \textbf{Without the rays-and-points (R\&P) mode}, many holes still remain in both the depth and rendering, as the ray directions are not diverse enough for the attention model to properly generalize to the deformed geometry. \textbf{Without the points-only mode}, the depth around the boundaries of the object becomes inaccurate, and there are visible artifacts around the boundaries in the rendering as well, since the fine-tuning overfits to the close viewpoints sampled in the rays-and-points mode. \textbf{Without the texture consistency loss} ($\mathcal{L}_{\text{tex}}$), the depth map has far fewer holes, but the rendering is noticeably blurred. \textbf{Our full method} with all components produces a hole-free depth map and a high-fidelity rendering with preserved details, demonstrating that all proposed components are essential for robust deformation.

\subsection{Application: Skinning Weights Optimization}

A downstream application of our method is to optimize the skinning weights of the parameterized scenes without a ground truth motion sequence.

Specifically, we extract $M (=128)$ sparse control anchors $\{\mathbf{a}_m\}_{m=1}^{M}$ from the canonical dense point cloud using Farthest Point Sampling (FPS). We parameterize the dense point cloud using a Neural Blend Skinning (NBS) approach, where an anchor skinning MLP $f_\phi$ with positional encoding takes a normalized 3D point coordinate and outputs logits over all $M$ anchors: $\boldsymbol{\ell}(\mathbf{p}) = f_\phi(\gamma(\mathbf{p})) \in \mathbb{R}^{M}$. For each point $\mathbf{p}$ to be deformed, we find its $K_{\text{nn}} (=8)$ nearest anchors by Euclidean distance, gather the corresponding logits, and apply softmax to obtain sparse skinning weights: $w_k = \text{softmax}(\ell_{m_k}(\mathbf{p}))$.

To derive per-anchor rigid transformations from an analytical deformation $\mathcal{D}$, we estimate the Jacobian $\mathbf{J}_m = \nabla \mathcal{D}(\mathbf{a}_m)$ via finite differences, extract the closest rotation matrix $\mathbf{R}_m$ via SVD projection, and recover the translation $\mathbf{t}_m = \mathcal{D}(\mathbf{a}_m) - \mathbf{R}_m \mathbf{a}_m$. The deformed position is then predicted via Linear Blend Skinning:
\begin{align}
    \mathbf{p}' = \sum_{k=1}^{K_{\text{nn}}} w_k \cdot (\mathbf{R}_{m_k} \mathbf{p} + \mathbf{t}_{m_k})
\end{align}

This NBS parameterization replaces the direct analytical deformation for the point cloud and surface points within our self-supervised framework, while camera rays are still deformed analytically. As the pipeline is end-to-end differentiable, during the self-supervised fine-tuning, we jointly optimize these skinning MLP weights alongside the neural renderer. At inference time, users perform edits by simply repositioning the sparse anchor points.

Table~\ref{tab:nbs-ablation} compares our optimized NBS weights against distance-based Gaussian-kernel RBF weights~\cite{Huang2023SCGSSG}, where $\hat{w}_{jk}\!=\!\exp(-d_{jk}^2/2\sigma^2)$ with $\sigma\!=\!1$, normalized to sum to one. For the geometry evaluation, we compare the deformed full point cloud derived from the deformed anchor points with the ground truth deformed mesh. For a fair comparison we use the same set of anchor points for both methods. As shown, our NBS consistently outperforms the RBF baseline across all rendering and geometry metrics, confirming that self-supervised deformation augmentation yields spatially discriminative skinning weights that generalize and better preserve geometry detail during editing. Fig.~\ref{fig:nbs-quali} shows qualitative editing results on two Objaverse scenes.

\begin{table}[htb]
\centering
\caption{Ablation on skinning weight estimation, averaged over 6 Objaverse scenes. Both methods use the same FPS anchors; the dense point cloud is deformed via anchor-driven RBF interpolation and compared against ground-truth end-state views and meshes (normalized to $[-1,1]^3$).}
\label{tab:nbs-ablation}
\setlength{\tabcolsep}{4pt}
\resizebox{\linewidth}{!}{
\begin{tabular}{l|ccc|cccc}
\toprule
\multirow{2}{*}{Skinning Weights} & \multicolumn{3}{c|}{Rendering Quality} & \multicolumn{4}{c}{Geometry Quality} \\
\cmidrule(lr){2-4} \cmidrule(lr){5-8}
& PSNR $\uparrow$ & SSIM $\uparrow$ & LPIPS $\downarrow$ & CD $\downarrow$ & EMD $\downarrow$ & F$_{0.01}$ $\uparrow$ & F$_{0.02}$ $\uparrow$ \\
\midrule
RBF~\cite{Huang2023SCGSSG} & 25.32 & 0.945 & 0.054 & \textbf{0.077} & 0.307 & 0.409 & 0.626 \\
NBS (Ours) & \textbf{26.46} & \textbf{0.952} & \textbf{0.050} & \textbf{0.077} & \textbf{0.303} & \textbf{0.438} & \textbf{0.640} \\
\bottomrule
\end{tabular}
}
\end{table}

\begin{figure}[htb]
\centering
\setlength\tabcolsep{0pt}
\scriptsize
\begin{tabular}{cccccc}
    GT & \makecell{w/o Aug.\\(RBF)} & \makecell{w/ Aug.\\(Ours NBS)} & GT & \makecell{w/o Aug.\\(RBF)} & \makecell{w/ Aug.\\(Ours NBS)} \\
    \includegraphics[width=0.155\linewidth]{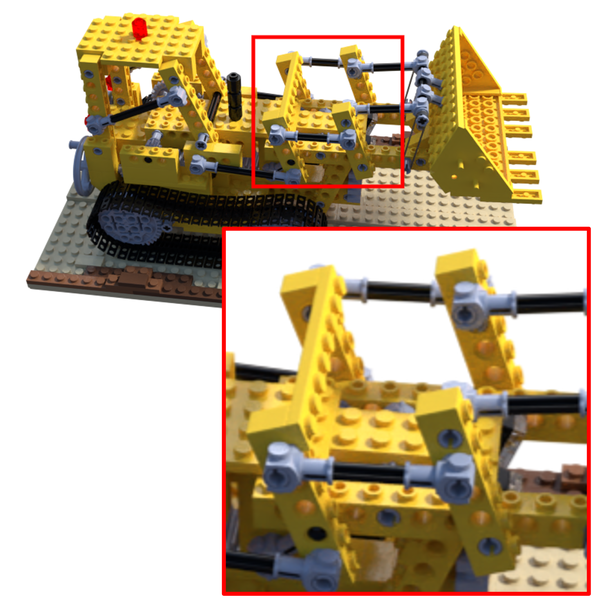} &
    \includegraphics[width=0.155\linewidth]{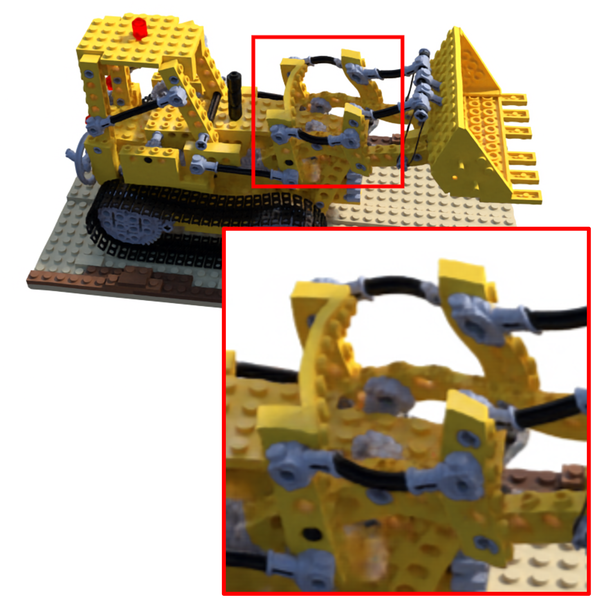} &
    \includegraphics[width=0.155\linewidth]{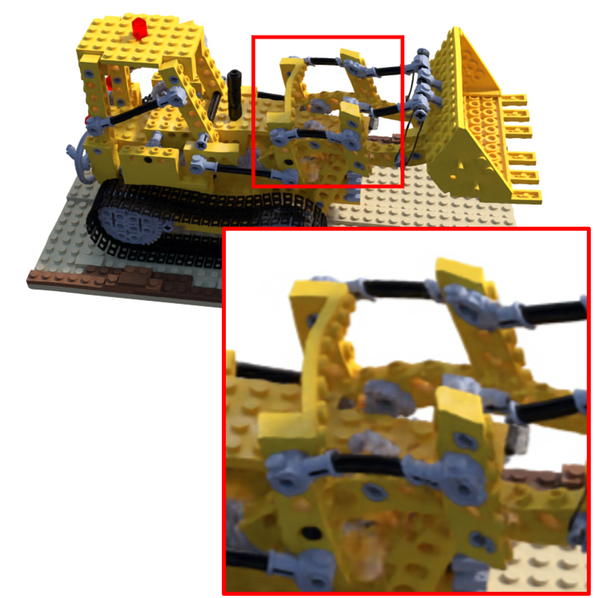} &
    \includegraphics[width=0.155\linewidth]{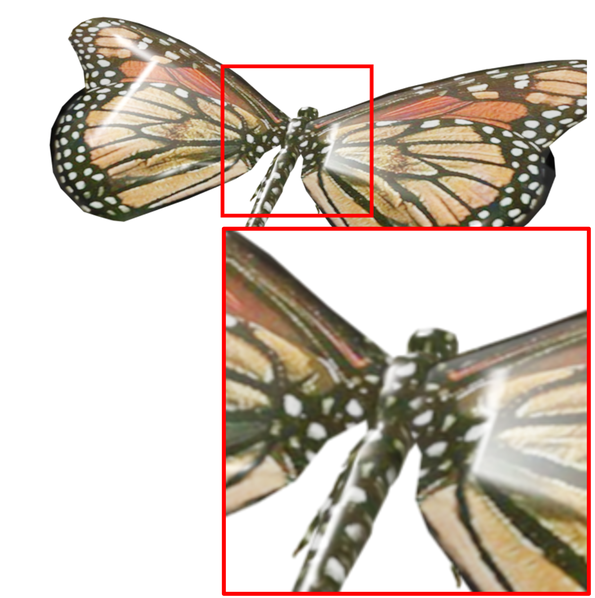} &
    \includegraphics[width=0.155\linewidth]{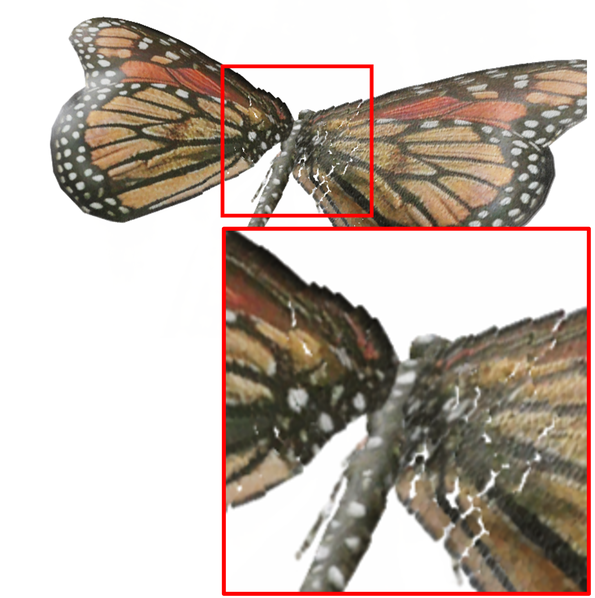} &
    \includegraphics[width=0.155\linewidth]{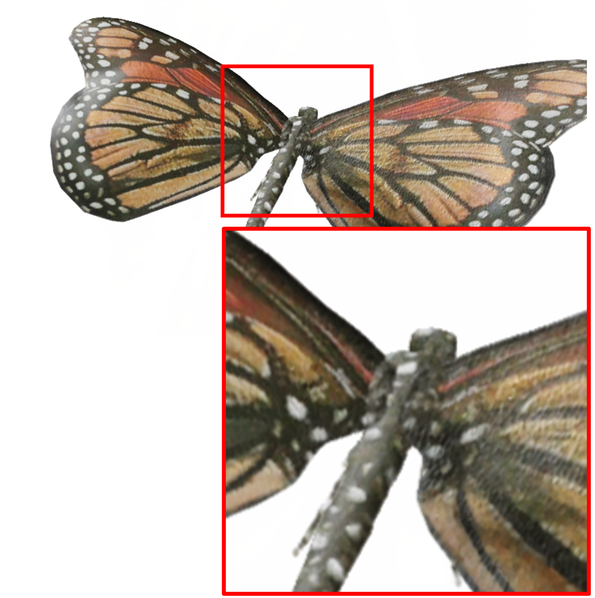} \\
\end{tabular}
\caption{Qualitative comparison of proxy-free sparse control editing on the Lego and Butterfly scenes. Without self-supervised augmentation, the RBF-deformed renderings exhibit unfaithful deformation and surface discontinuities. With our self-supervised fine-tuning, the edited geometry is better aligned with the ground-truth target.}
\label{fig:nbs-quali}
\end{figure}

\section{Conclusion and Future Work}
In this paper, we presented a self-supervised surface-consistency framework that addresses a fundamental tension in point-based neural rendering: the absence of fixed connectivity provides flexibility for reconstruction, but can also lead to holes and surface discontinuities under large deformations. Our approach generates random deformations and enforces that surface prediction commutes with deformation, eliminating the need for ground-truth dynamic data or explicit geometry proxies. We integrated this framework into PAPR, an attention-based neural point renderer whose learned interpolation kernel can adapt to large deformations without adding or removing points. Experiments on synthetic benchmarks and real-world scenes demonstrate the effectiveness of our approach, and we further show its applicability to downstream tasks such as Neural Blend Skinning (NBS). A limitation of the current method is that excessive fine-tuning can lead to over-smoothed or blurry renderings, suggesting a trade-off between deformation consistency and appearance fidelity. Future work could explore improved regularization or adaptive stopping criteria, as well as integration with physically based simulation, more sophisticated multi-level editing controls, and extensions to other point-based architectures.

\section{Acknowledgments}

This research was enabled in part by support provided by NSERC, the BC DRI Group, the Digital Research Alliance of Canada and the Canada CIFAR AI Chairs Program.

\newpage
\bibliographystyle{splncs04}
\bibliography{main}

\newpage
\appendix

\section{Qualitative Comparison with Neural Editor}

\begin{figure}[htb]
\centering
\tiny
\begin{tabularx}{1\linewidth}{l@{\hskip 1em}YYYYYYYY}
    &   Chair
    &   Drums
    &   Ficus
    &   Hotdog
    &   Lego
    &   Materials
    &   Mic
    &   Ship
    \\
\rotatebox[origin=c]{90}{Canonical Pose}
    &   \includegraphics[width=\hsize,valign=m]{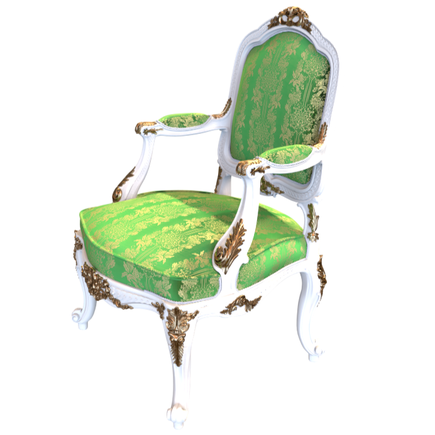}
    &   \includegraphics[width=\hsize,valign=m]{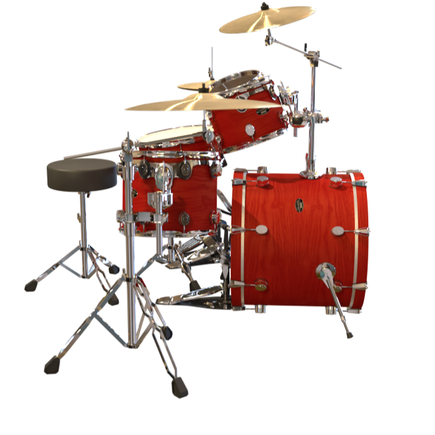}
    &   \includegraphics[width=\hsize,valign=m]{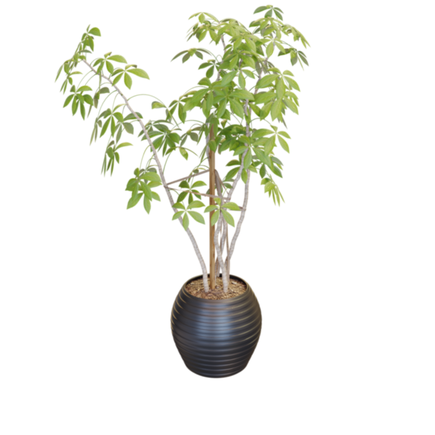}
    &   \includegraphics[width=\hsize,valign=m]{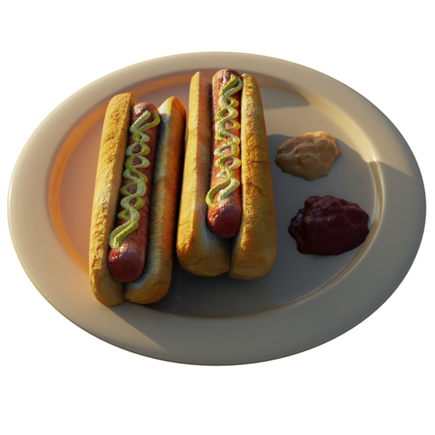}
    &   \includegraphics[width=\hsize,valign=m]{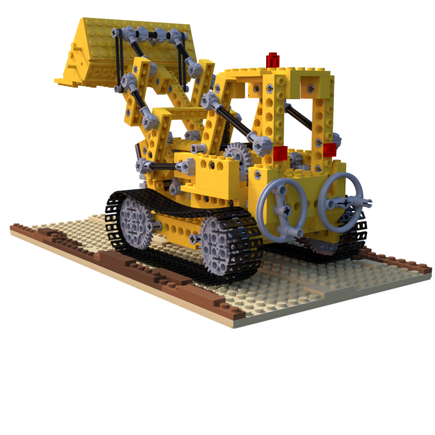}
    &   \includegraphics[width=\hsize,valign=m]{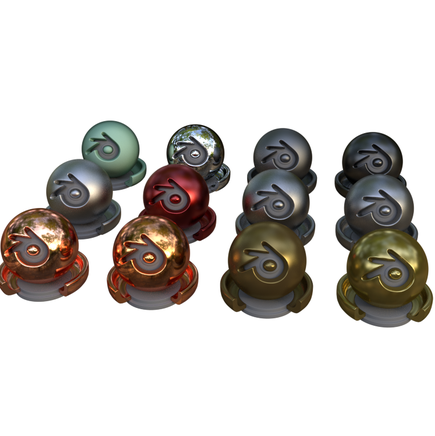}
    &   \includegraphics[width=\hsize,valign=m]{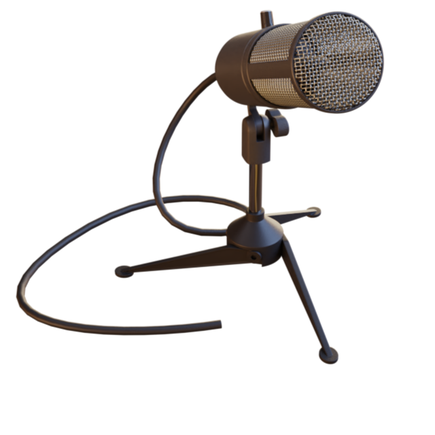}
    &   \includegraphics[width=\hsize,valign=m]{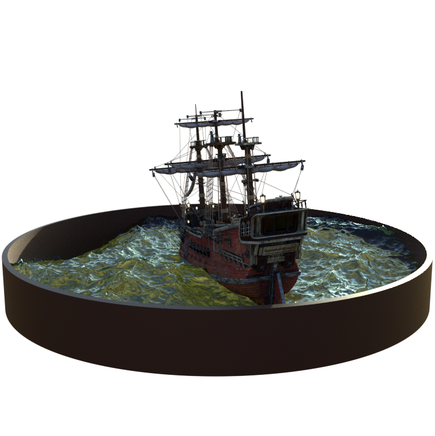}
    \\
\rotatebox[origin=c]{90}{Neural Editor}
    &   \includegraphics[width=\hsize,valign=m]{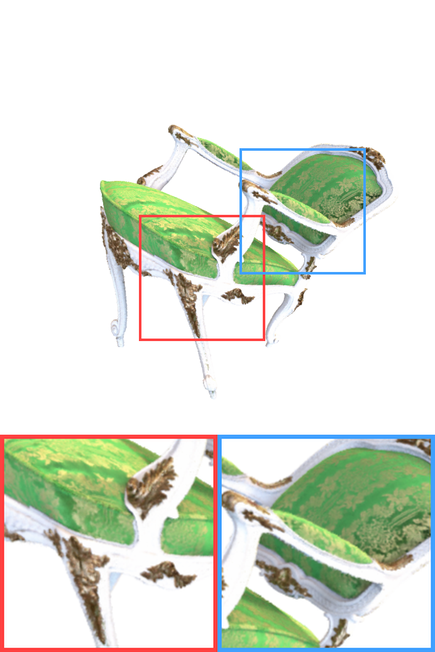}
    &   \includegraphics[width=\hsize,valign=m]{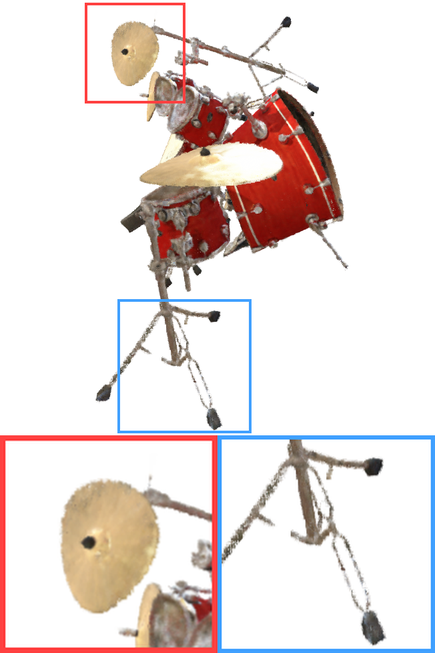}
    &   \includegraphics[width=\hsize,valign=m]{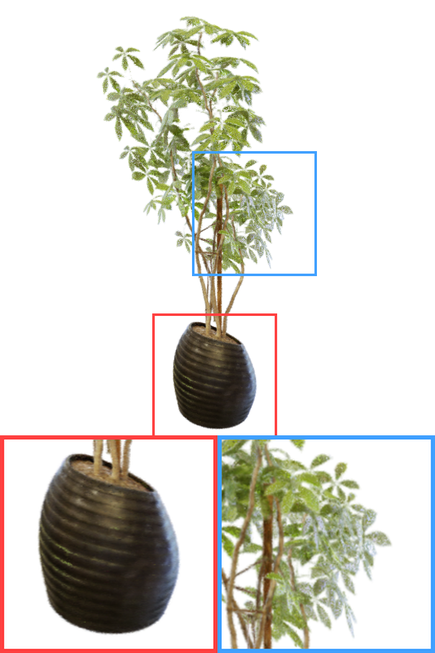}
    &   \includegraphics[width=\hsize,valign=m]{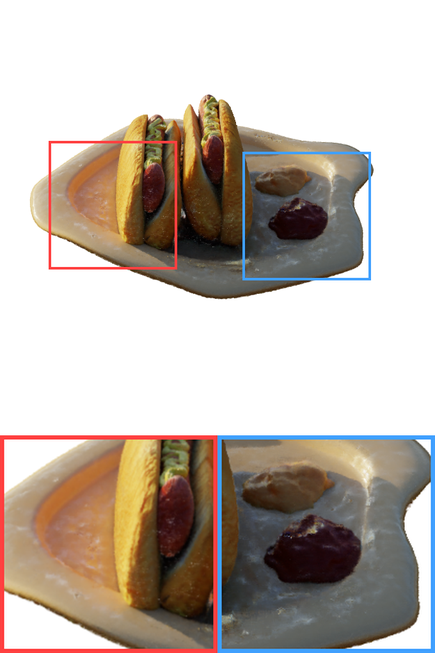}
    &   \includegraphics[width=\hsize,valign=m]{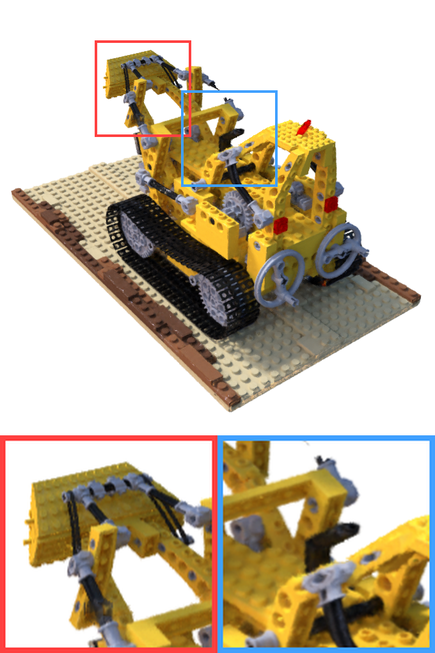}
    &   \includegraphics[width=\hsize,valign=m]{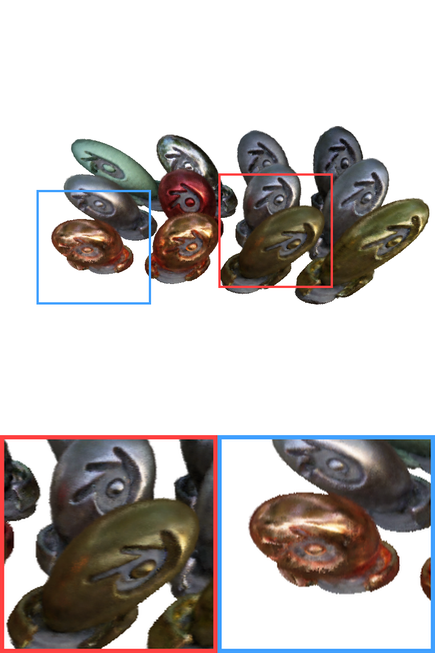}
    &   \includegraphics[width=\hsize,valign=m]{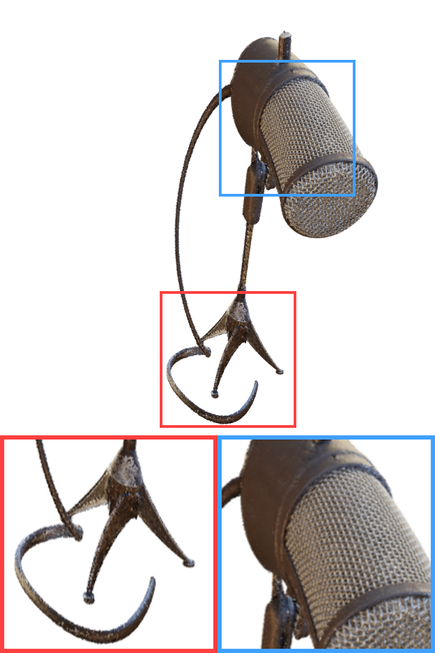}
    &   \includegraphics[width=\hsize,valign=m]{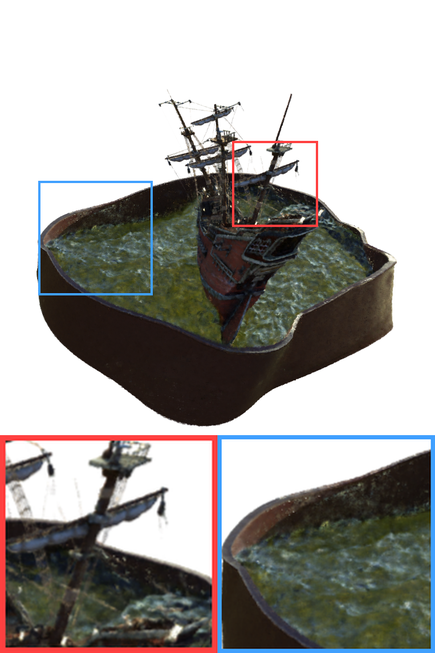}
    \\
\rotatebox[origin=c]{90}{Ours}
    &   \includegraphics[width=\hsize,valign=m]{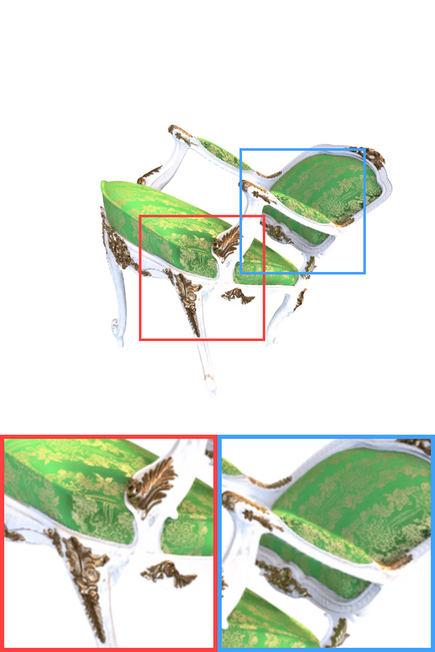}
    &   \includegraphics[width=\hsize,valign=m]{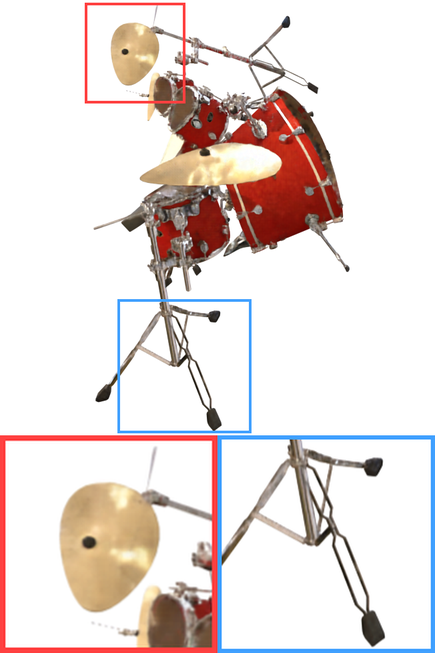}
    &   \includegraphics[width=\hsize,valign=m]{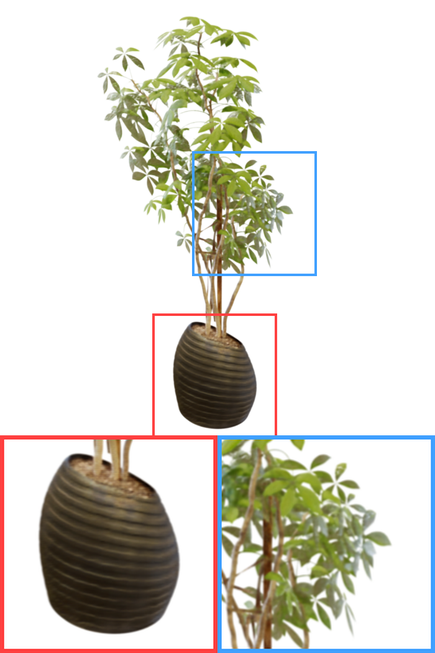}
    &   \includegraphics[width=\hsize,valign=m]{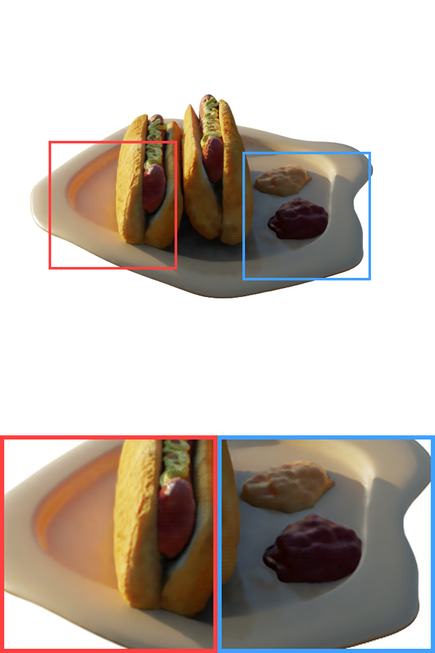}
    &   \includegraphics[width=\hsize,valign=m]{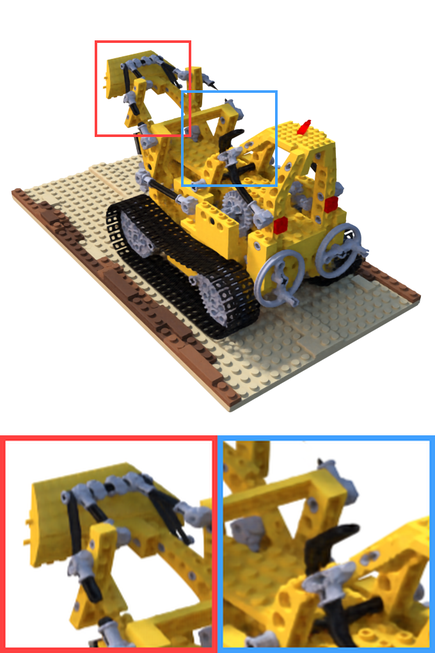}
    &   \includegraphics[width=\hsize,valign=m]{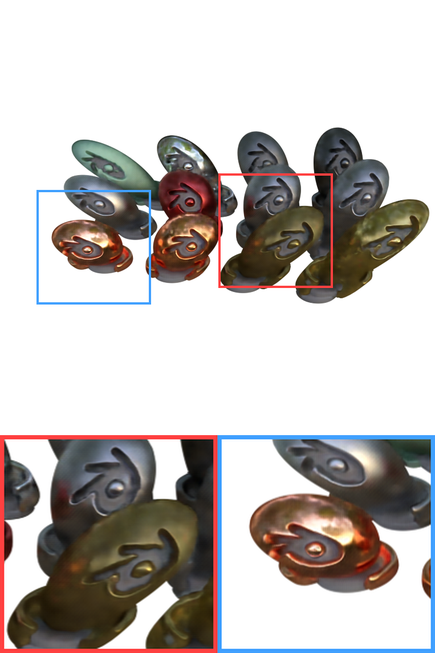}
    &   \includegraphics[width=\hsize,valign=m]{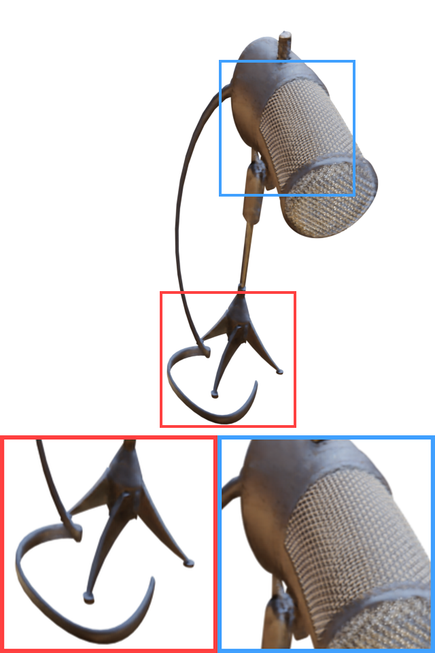}
    &   \includegraphics[width=\hsize,valign=m]{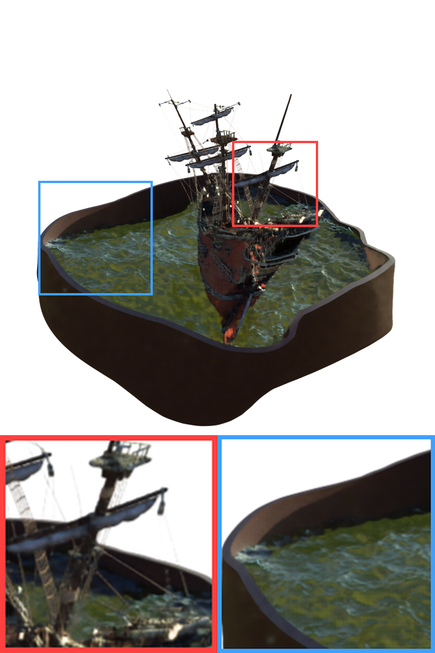}
    \\
\rotatebox[origin=c]{90}{GT}
    &   \includegraphics[width=\hsize,valign=m]{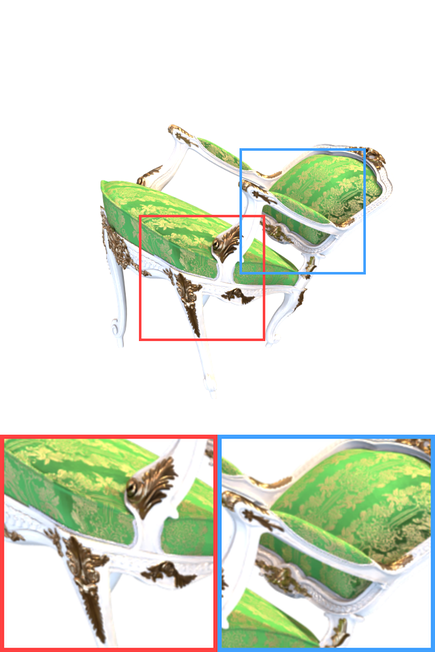}
    &   \includegraphics[width=\hsize,valign=m]{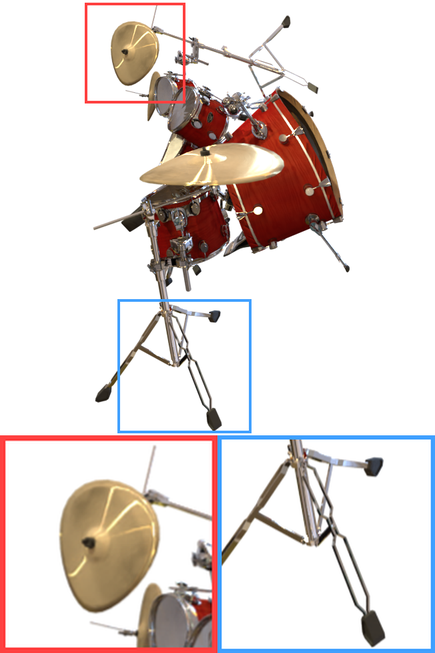}
    &   \includegraphics[width=\hsize,valign=m]{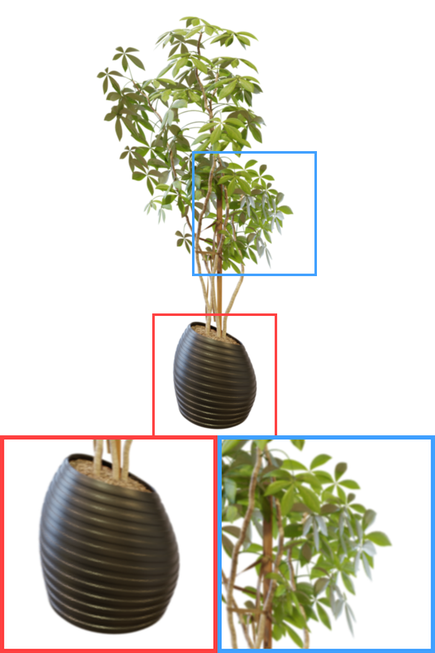}
    &   \includegraphics[width=\hsize,valign=m]{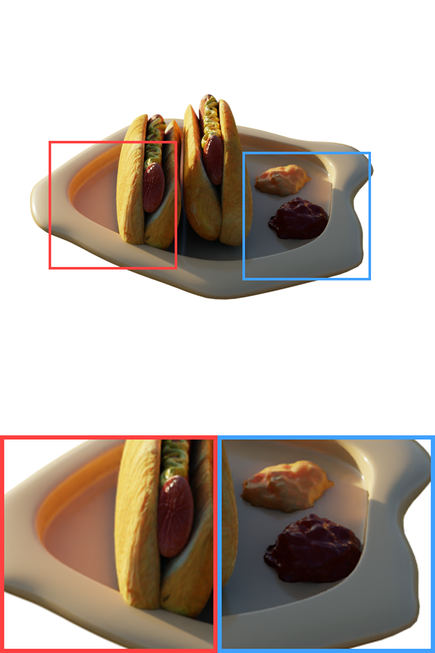}
    &   \includegraphics[width=\hsize,valign=m]{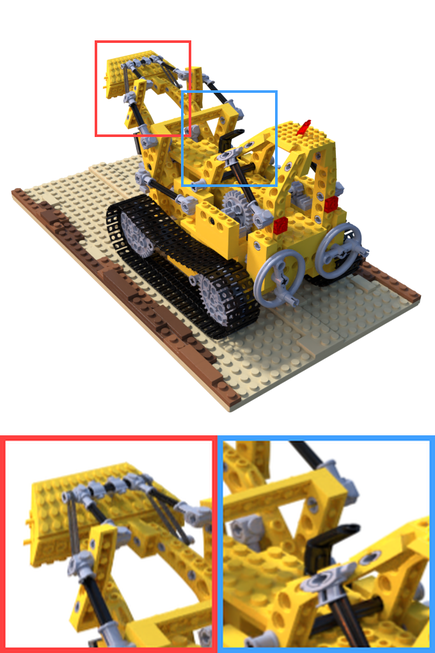}
    &   \includegraphics[width=\hsize,valign=m]{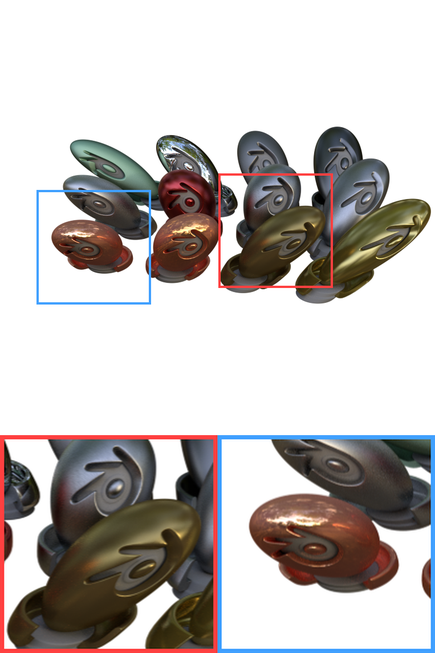}
    &   \includegraphics[width=\hsize,valign=m]{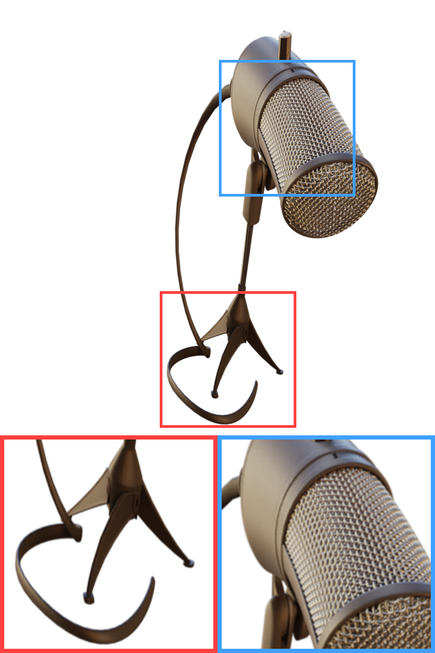}
    &   \includegraphics[width=\hsize,valign=m]{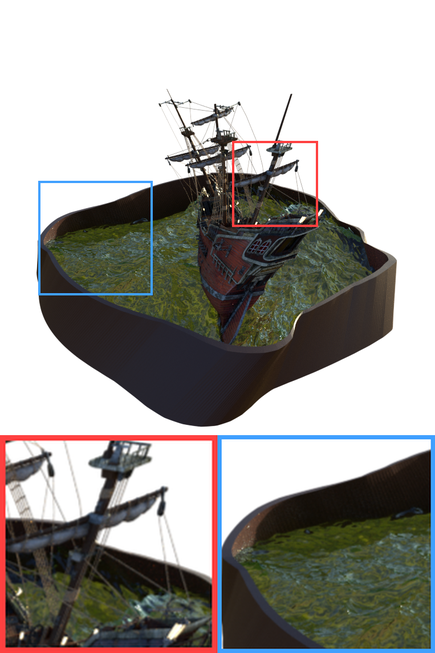}
\end{tabularx}
\caption{
    Qualitative comparison between our method and Neural Editor on the Neural Editor dataset~\cite{Chen2023NeuralEditorEN}.
}
\label{fig:compare-ne-neural-editor}
\end{figure}

\begin{figure}[htb]
\centering
\tiny
\setlength\tabcolsep{5.5pt}
\begin{tabularx}{1\linewidth}{l@{\hskip 1em}YYYYYY}
    &   Butterfly
    &   Crab
    &   Dolphin
    &   Giraffe
    &   Lego
    &   Legoman
    \\
\rotatebox[origin=c]{90}{Canonical Pose}
    &   \includegraphics[width=\hsize,valign=m]{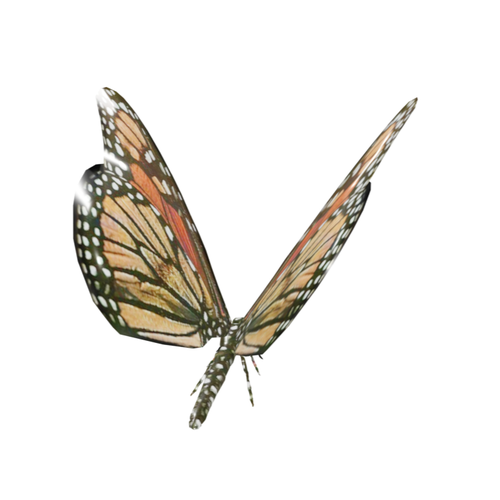}
    &   \includegraphics[width=\hsize,valign=m]{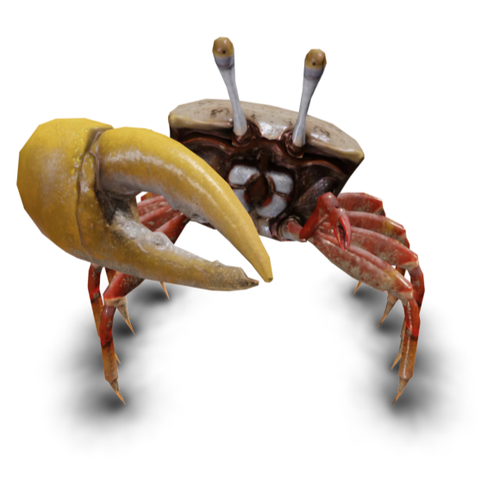}
    &   \includegraphics[width=\hsize,valign=m]{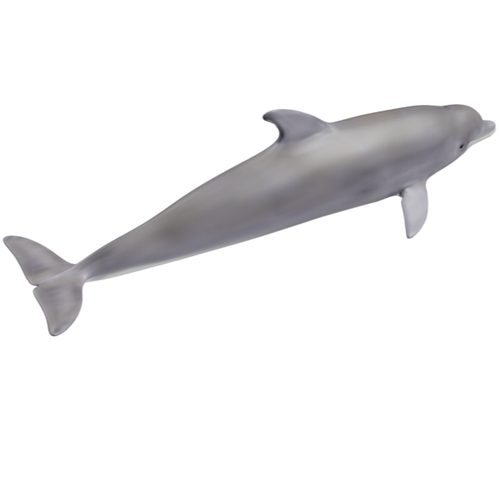}
    &   \includegraphics[width=\hsize,valign=m]{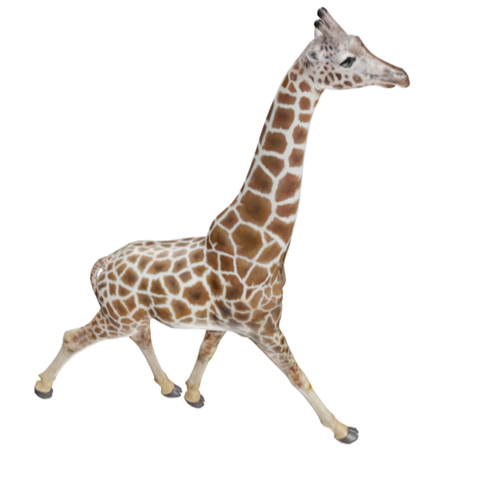}
    &   \includegraphics[width=\hsize,valign=m]{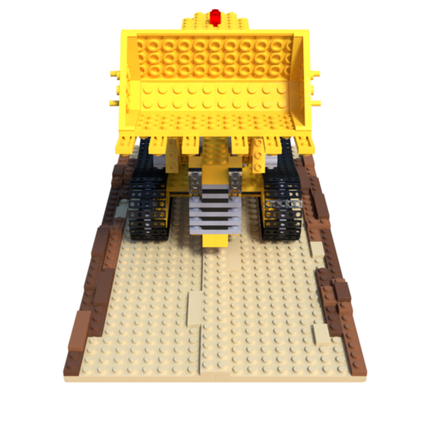}
    &   \includegraphics[width=\hsize,valign=m]{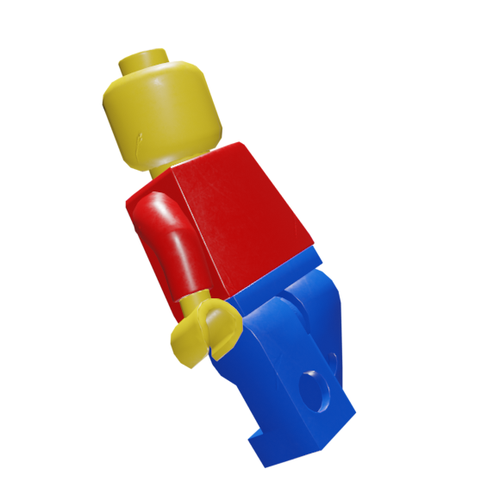}
    \\
\rotatebox[origin=c]{90}{Neural Editor}
    &   \includegraphics[width=\hsize,valign=m]{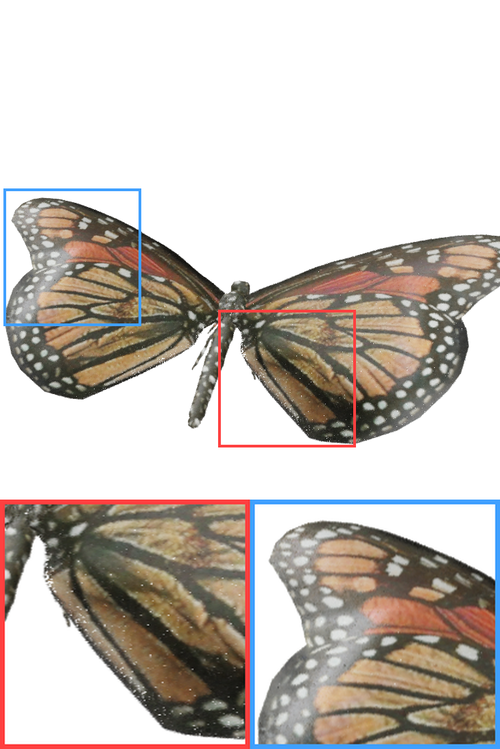}
    &   \includegraphics[width=\hsize,valign=m]{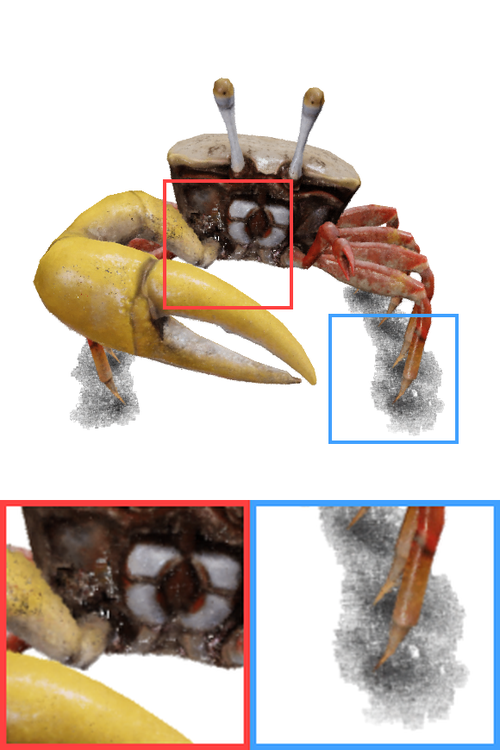}
    &   \includegraphics[width=\hsize,valign=m]{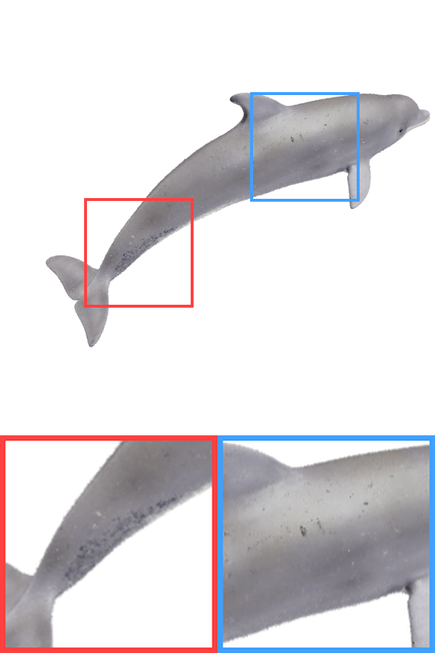}
    &   \includegraphics[width=\hsize,valign=m]{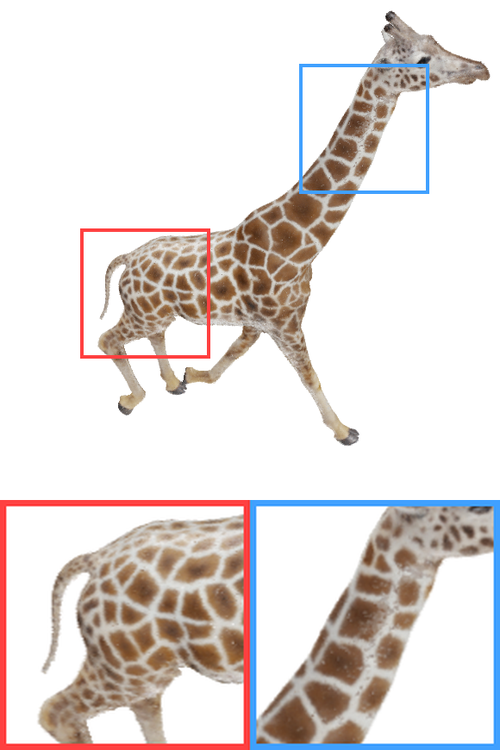}
    &   \includegraphics[width=\hsize,valign=m]{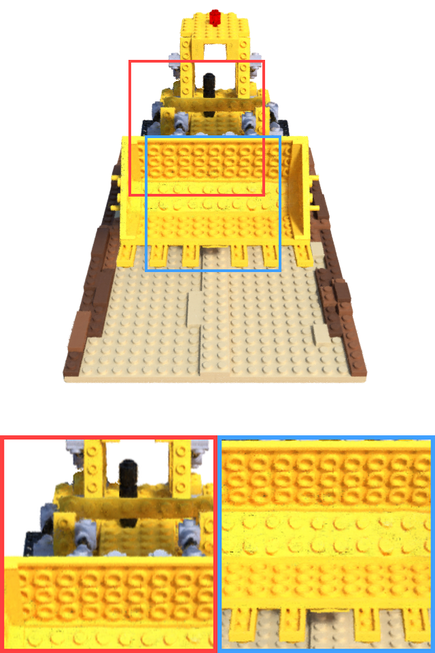}
    &   \includegraphics[width=\hsize,valign=m]{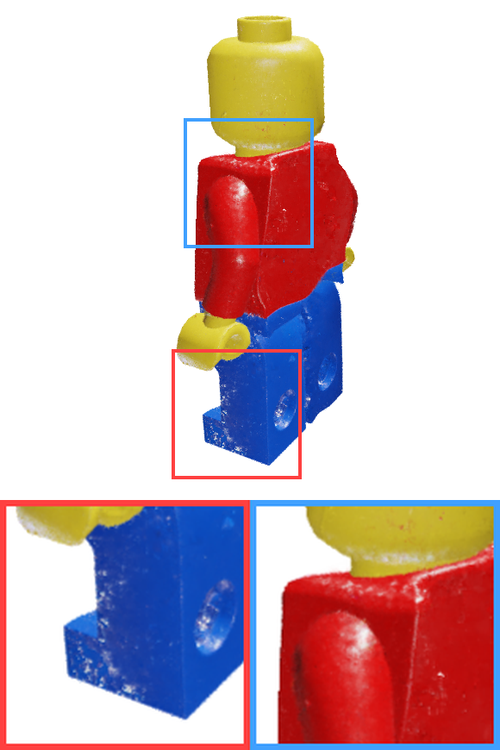}
    \\
\rotatebox[origin=c]{90}{Ours}
    &   \includegraphics[width=\hsize,valign=m]{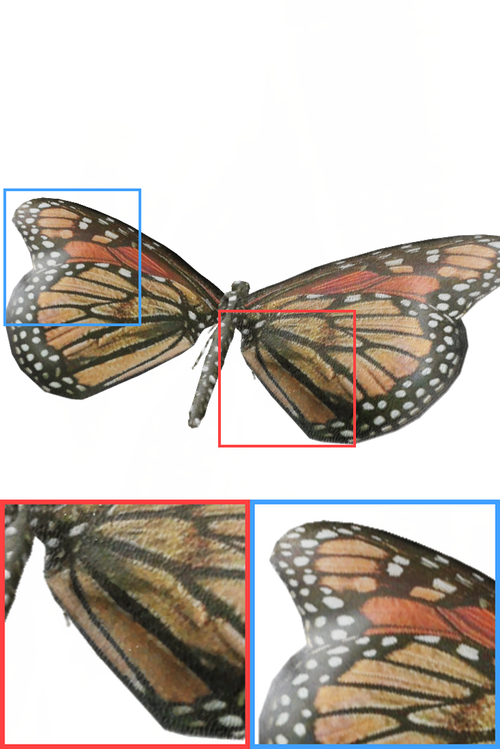}
    &   \includegraphics[width=\hsize,valign=m]{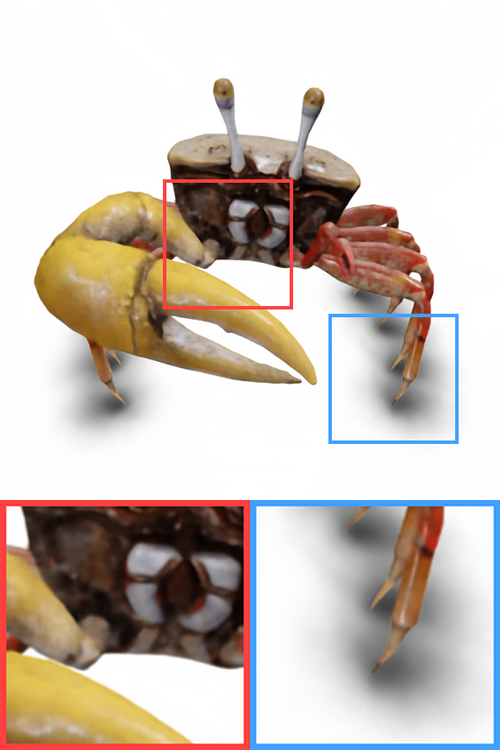}
    &   \includegraphics[width=\hsize,valign=m]{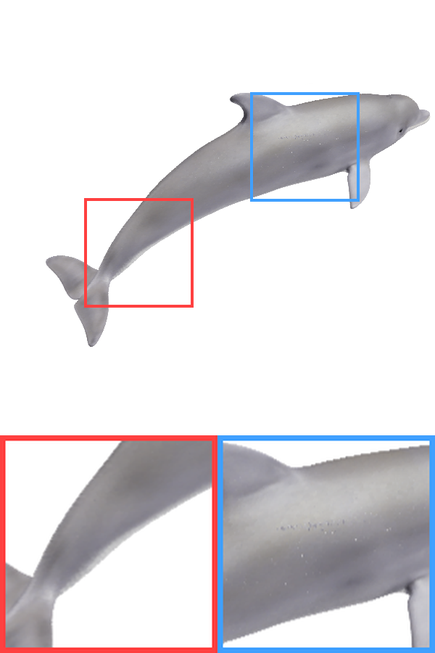}
    &   \includegraphics[width=\hsize,valign=m]{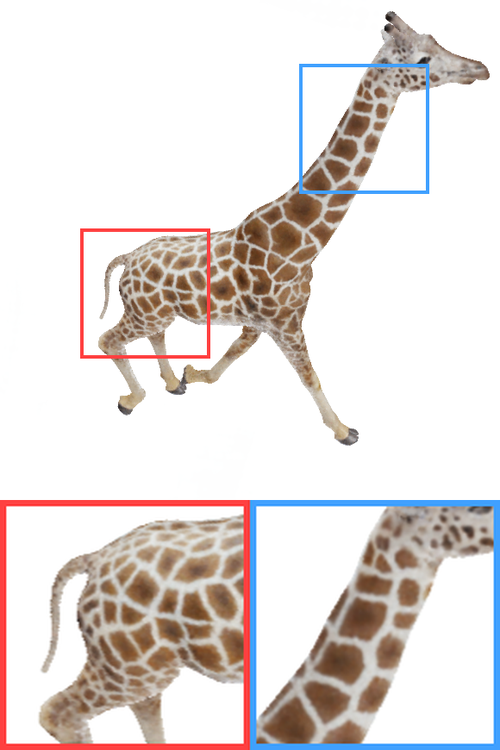}
    &   \includegraphics[width=\hsize,valign=m]{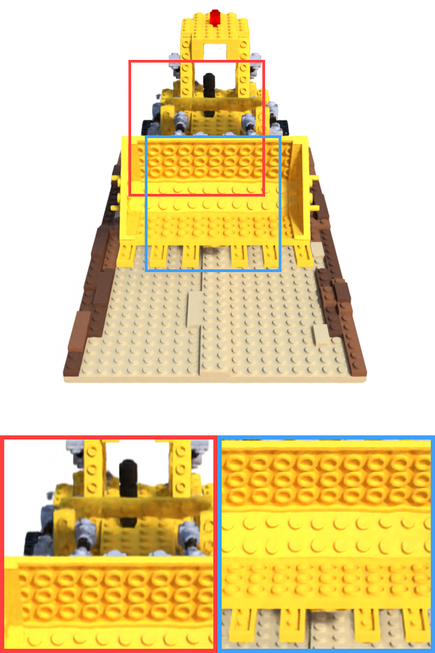}
    &   \includegraphics[width=\hsize,valign=m]{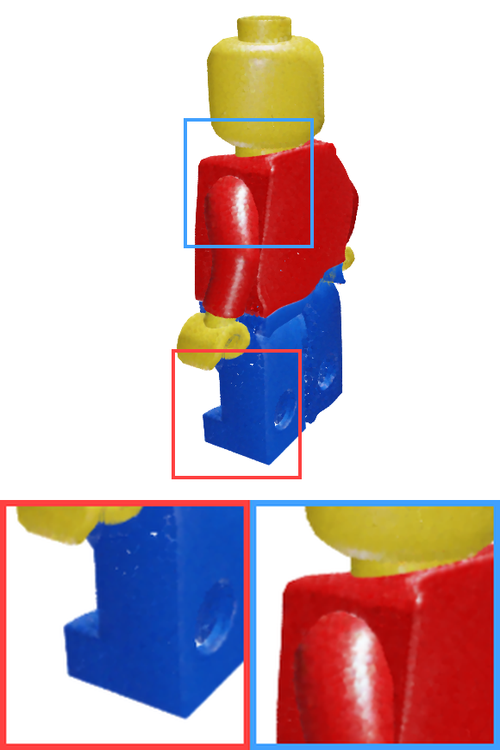}
    \\
\rotatebox[origin=c]{90}{GT}
    &   \includegraphics[width=\hsize,valign=m]{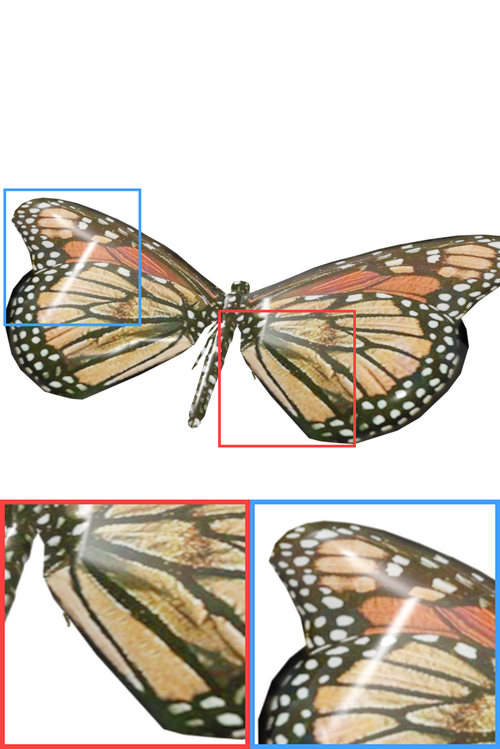}
    &   \includegraphics[width=\hsize,valign=m]{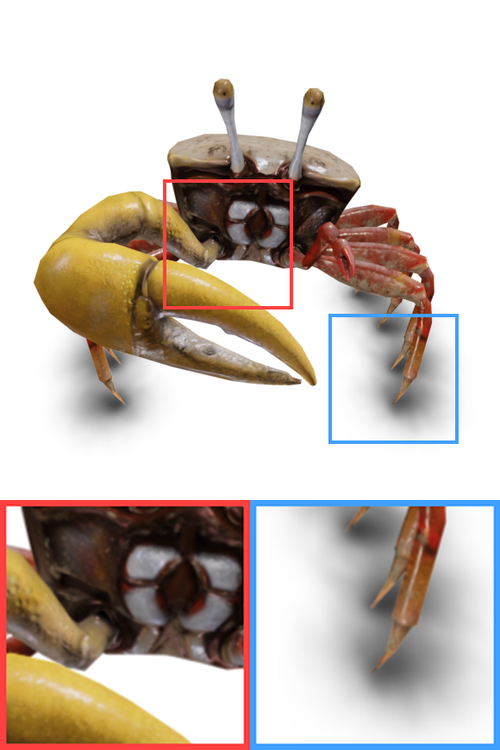}
    &   \includegraphics[width=\hsize,valign=m]{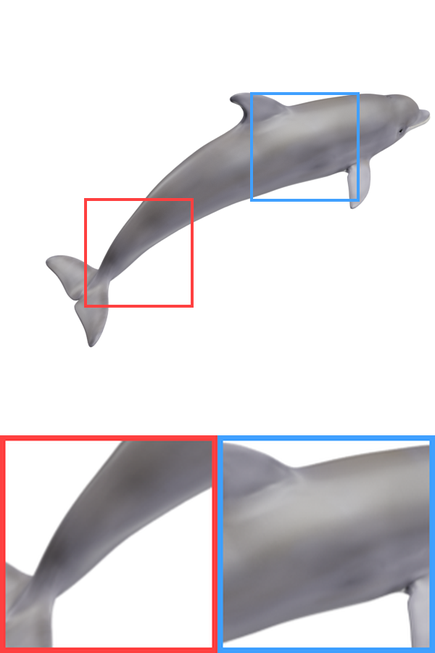}
    &   \includegraphics[width=\hsize,valign=m]{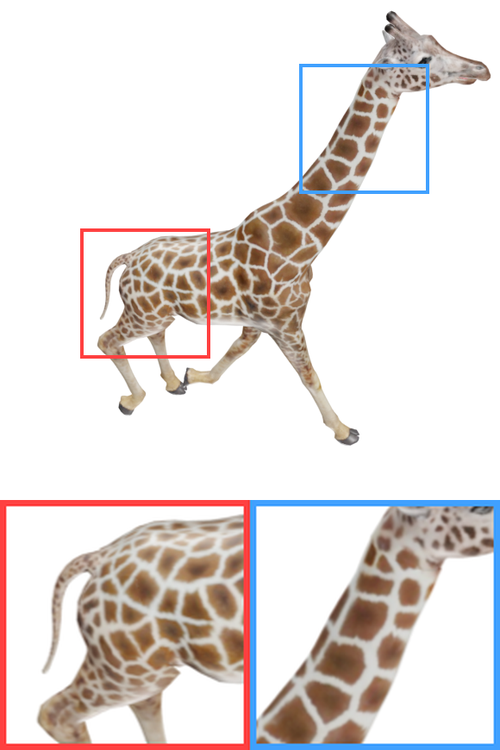}
    &   \includegraphics[width=\hsize,valign=m]{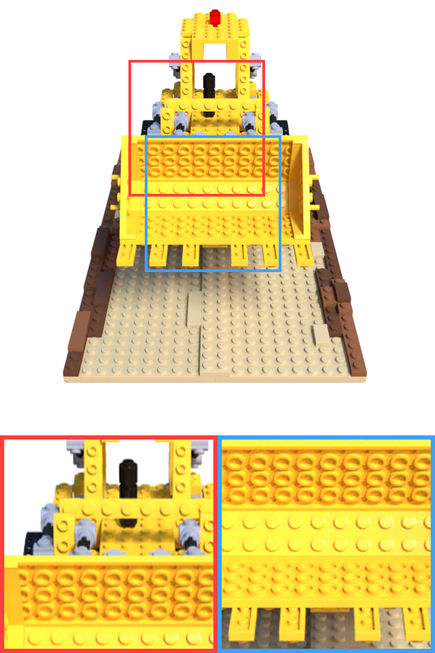}
    &   \includegraphics[width=\hsize,valign=m]{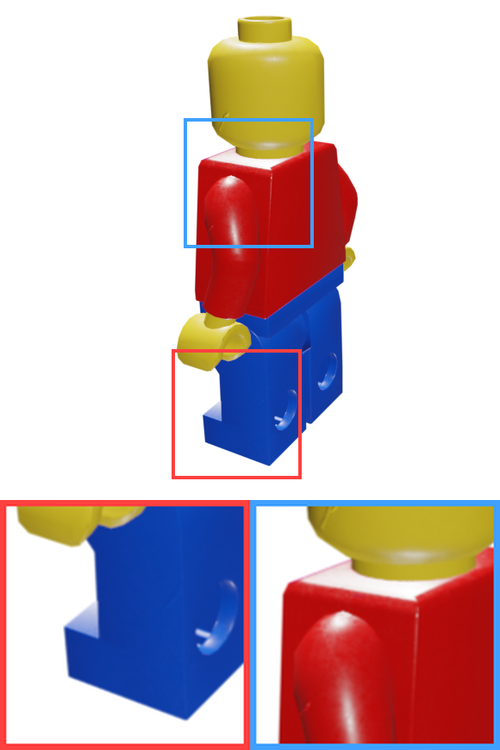}
\end{tabularx}
\caption{
    Qualitative comparison between our method and Neural Editor on scenes from Objaverse~\cite{Deitke2022ObjaverseAU, Peng2024PAPRIM}.
}
\label{fig:compare-ne-objaverse}
\end{figure}

Figures~\ref{fig:compare-ne-neural-editor} and~\ref{fig:compare-ne-objaverse} present side-by-side qualitative comparisons between our method and Neural Editor~\cite{Chen2023NeuralEditorEN} on the Neural Editor and Objaverse datasets, respectively. Both methods produce visually faithful deformations; however, Neural Editor occasionally exhibits holes or color bleeding in fine-detail regions, whereas our method preserves surface continuity. Beyond rendering quality, Neural Editor is also far more expensive: it requires $\sim$7 days of training per scene and a further $\sim$23 minutes of preprocessing per frame at inference on an NVIDIA RTX 3090---including KD-tree construction, mesh normal recomputation, and infinitesimal surface transformation---whereas our method needs only $\sim$1 minute of fine-tuning and no preprocessing of any kind.

\section{Comparison with PAPR-in-Motion}

PAPR-in-Motion~\cite{Peng2024PAPRIM} uses a supervised two-stage pipeline: a geometry fine-tuning stage that moves the start-state point cloud to match the end-state geometry, followed by an appearance fine-tuning stage that captures deformation-induced appearance changes. Table~\ref{tab:papr-motion} compares our zero-shot result against the renderings from both stages over the six Objaverse scenes. Without any target-state rendering or geometry supervision, our method substantially outperforms the geometry-only stage. The remaining gap to the appearance-finetuned result is expected, since that variant is directly supervised by target-state ground-truth images, including shadows, highlights, and reflections that are difficult to infer zero-shot.

\begin{table}[htb]
\centering
\footnotesize
\setlength{\tabcolsep}{4pt}
\caption{Ours vs.\ PAPR-in-Motion over the six Objaverse scenes.}
\label{tab:papr-motion}
\resizebox{\linewidth}{!}{%
\begin{tabular}{lccc}
\toprule
Method & PSNR\,$\uparrow$ & SSIM\,$\uparrow$ & LPIPS\,$\downarrow$ \\
\midrule
PAPR-in-Motion (geometry finetune only) & $26.31$ & $0.946$ &  $0.044$ \\
PAPR-in-Motion (geometry + appearance finetune) & $33.46$ & $0.986$ &  $0.021$ \\
Ours (zero-shot, no end-state supervision)            & $28.84$ & $0.968$ & $0.030$ \\
\bottomrule
\end{tabular}}
\end{table}

\section{Range of Deformation}

During fine-tuning, we use analytic deformations following Blender's Simple Deform operations, including stretch, twist, bend, and uniform scaling. Stretch allows up to a $3{\times}$ factor, twist and bend allow up to $60^{\circ}$, and uniform scaling allows up to $2{\times}$ dilation or $0.5{\times}$ shrinkage. The deformation type, axis/direction, and magnitude are uniformly sampled at each iteration, without composing multiple deformations.

\section{Additional Realistic Deformation Results}

\begin{figure}[htb]
\centering
\setlength{\tabcolsep}{0pt}
\renewcommand{\arraystretch}{0}
\scriptsize
\begin{tabular}{ccc}
Train ($t_0$) & PAPR ($t_1$) \hspace{18pt} Ours ($t_1$) & PAPR ($t_2$) \hspace{18pt} Ours ($t_2$) \\
\includegraphics[width=0.1316\linewidth]{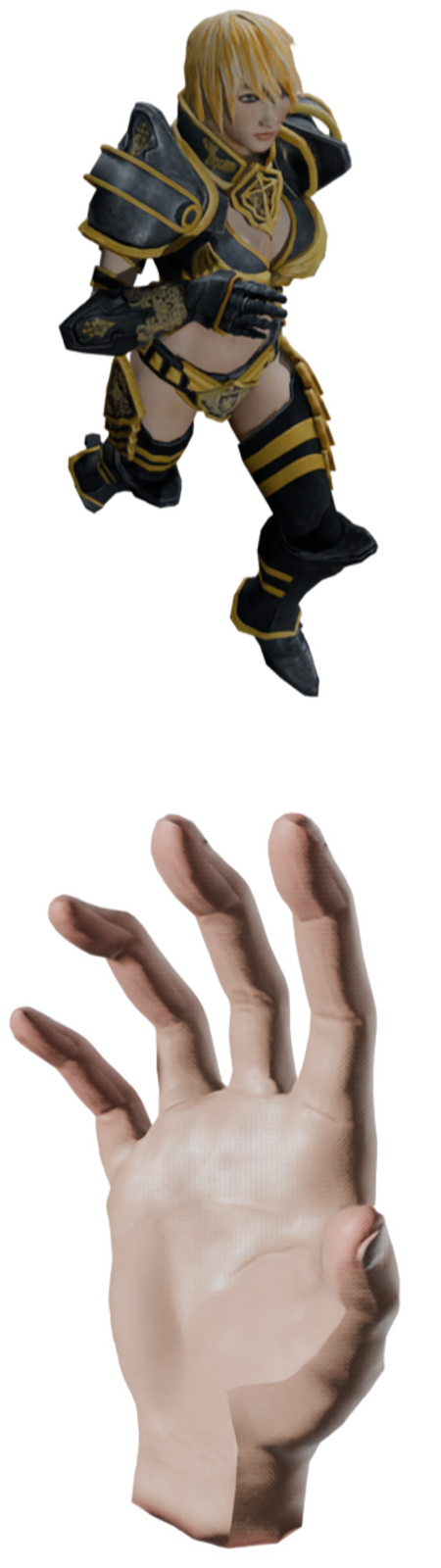} &
\includegraphics[width=0.4386\linewidth]{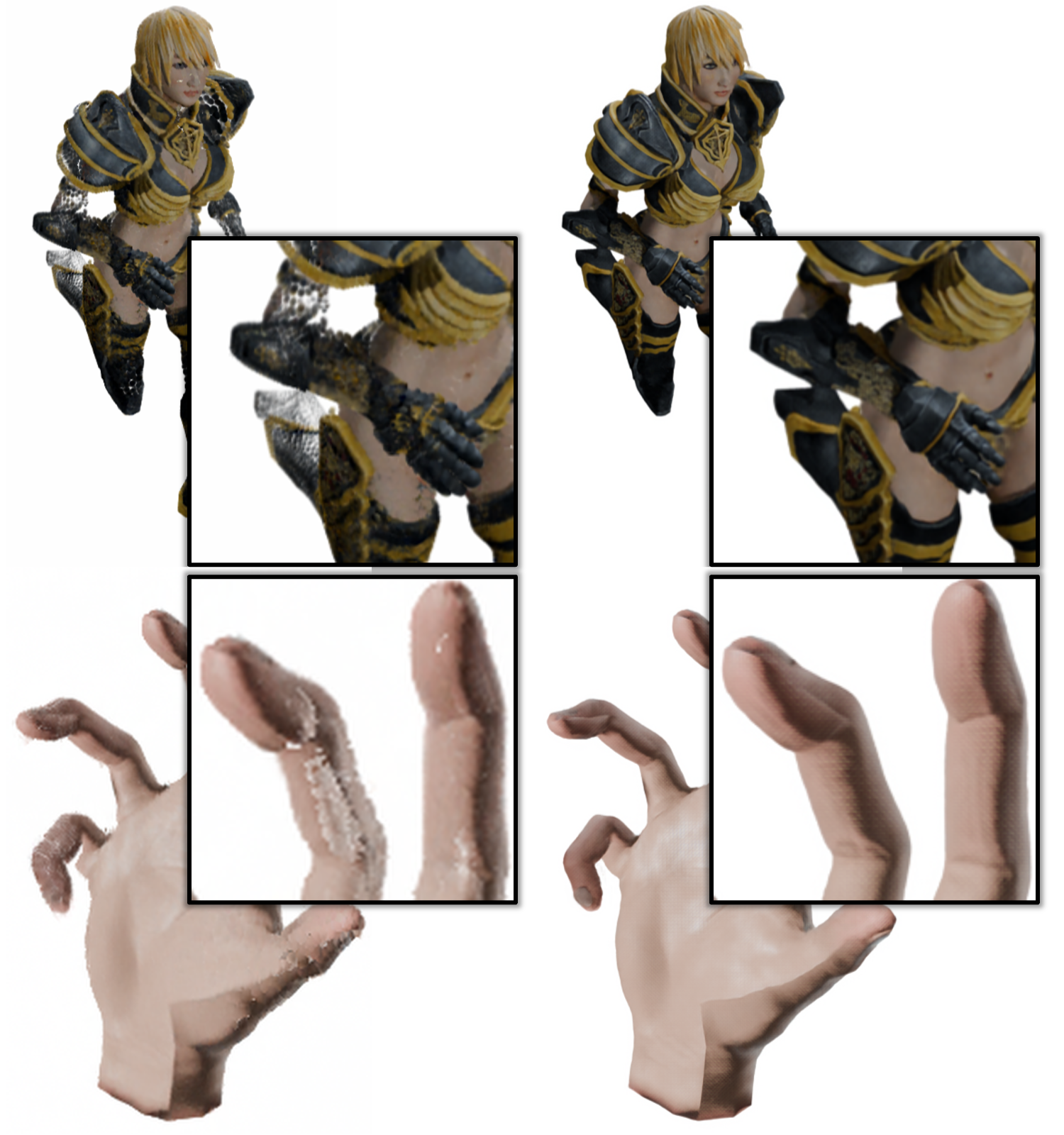} &
\includegraphics[width=0.4298\linewidth]{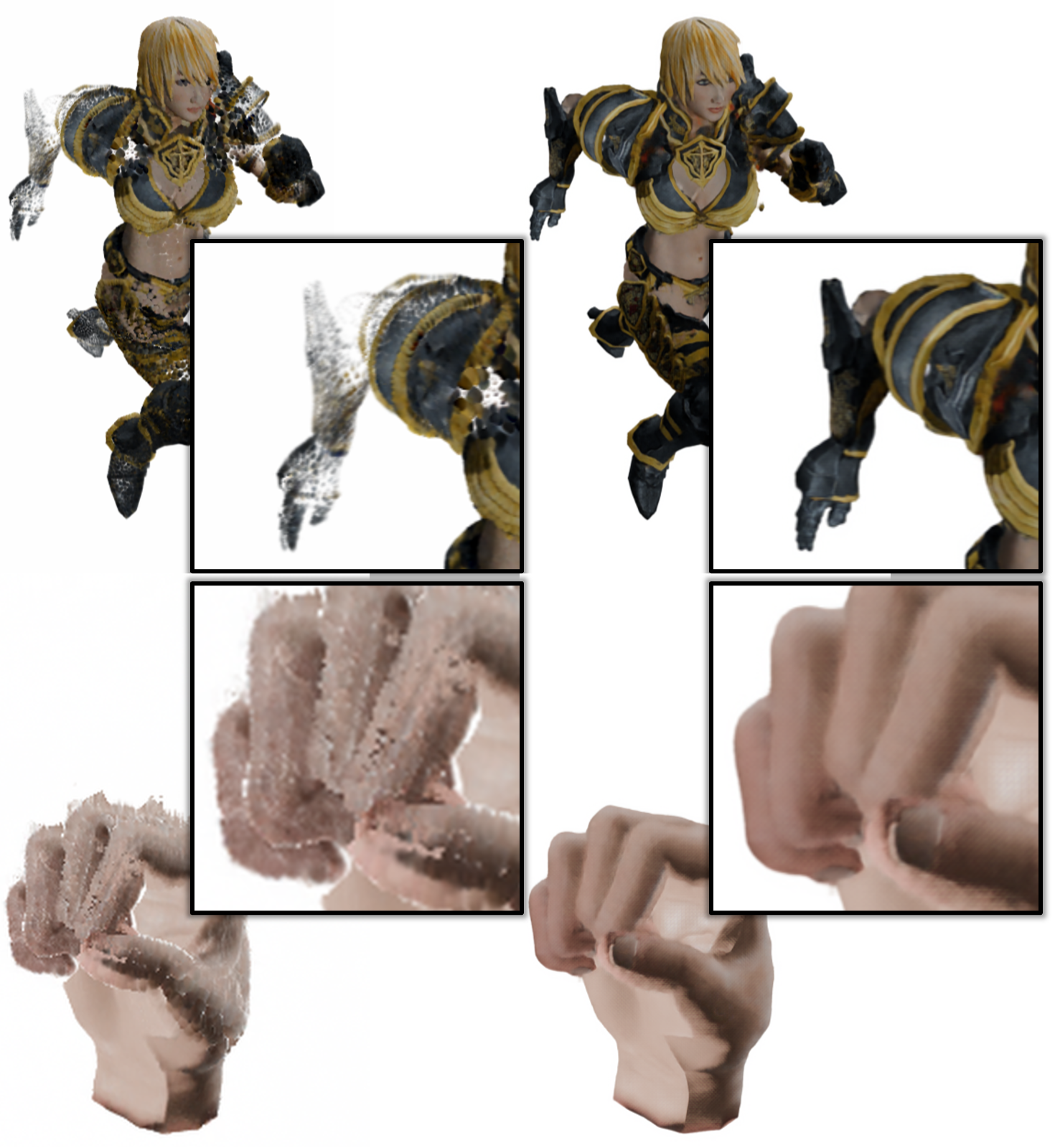} \\
\end{tabular}
\caption{Results for realistic deformation. $t_{1,2}$ are zero-shot edits.}
\label{fig:real-life-deform}
\end{figure}

In addition to the Objaverse scenes with realistic motions evaluated in the main paper (e.g., the flying butterfly and walking giraffe), we evaluate two additional dynamic scenes from Objaverse~\cite{Deitke2022ObjaverseAU} and Adobe Mixamo~\cite{mixamo} with realistic motions: making a fist and running. We train our method on the first time step and evaluate multi-view renderings at three uniformly sampled held-out time steps. As shown in Fig.~\ref{fig:real-life-deform} and Table~\ref{tab:real-life-deform}, our method effectively improves surface continuity for these realistic deformations, despite being fine-tuned only with random analytical deformations.

\begin{table}[htb]
\centering
\footnotesize
\setlength{\tabcolsep}{6pt}
\caption{Quantitative results for the realistic deformations.}
\label{tab:real-life-deform}
\begin{tabular}{lcccccc}
\toprule
 & \multicolumn{3}{c}{Objaverse hand} & \multicolumn{3}{c}{Mixamo running} \\
\cmidrule(lr){2-4} \cmidrule(lr){5-7}
Method & PSNR\,$\uparrow$ & SSIM\,$\uparrow$ & LPIPS\,$\downarrow$ & PSNR\,$\uparrow$ & SSIM\,$\uparrow$ & LPIPS\,$\downarrow$ \\
\midrule
PAPR          & $28.55$ & $0.964$ & $0.040$ & $24.40$ & $0.956$ & $0.046$ \\
\textbf{Ours} & $\mathbf{29.66}$ & $\mathbf{0.972}$ & $\mathbf{0.028}$ & $\mathbf{29.95}$ & $\mathbf{0.979}$ & $\mathbf{0.021}$ \\
\bottomrule
\end{tabular}
\end{table}

\section{Comparison with GaussianEditor}

\begin{figure}[htb]
\centering
\includegraphics[width=\linewidth]{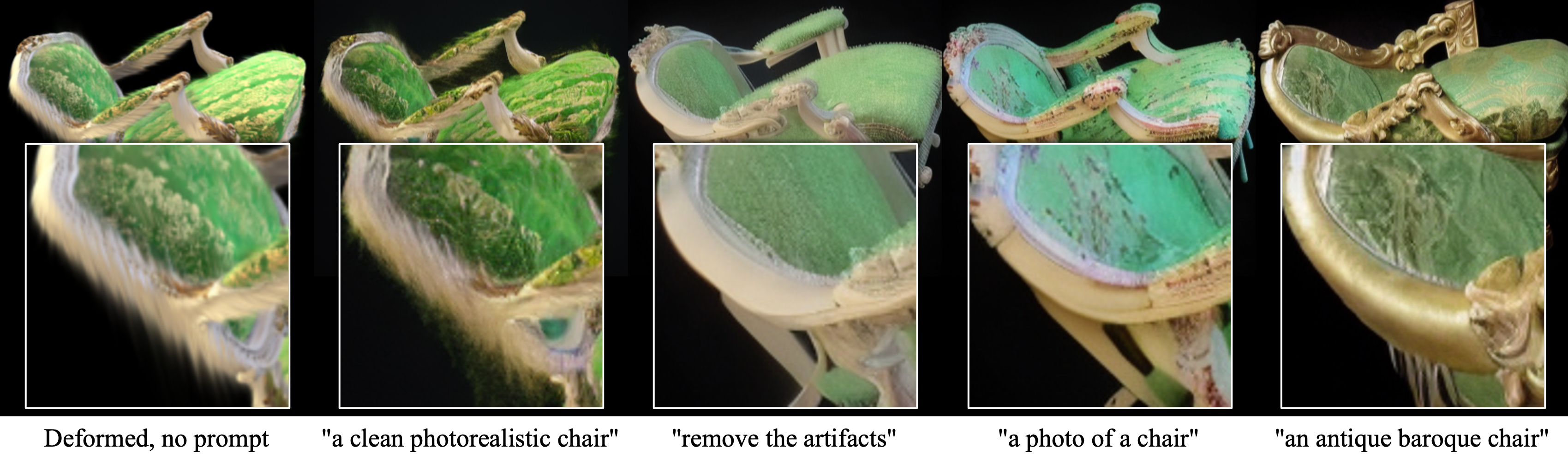}
\caption{GaussianEditor refinement (please zoom in for prompts).}
\label{fig:gseditor}
\end{figure}

GaussianEditor~\cite{chen2024gaussianeditor} mainly focuses on text-driven appearance editing and object addition/removal, rather than free-form deformation. Nevertheless, we evaluate whether it can refine deformed Gaussians and reduce the resulting artifacts using different prompts, as shown in Fig.~\ref{fig:gseditor}. As shown, it fails to reduce the artifacts while preserving identity or sharp details.

\section{Reconstruction Quality on Real-World Scenes}

We evaluate the novel-view synthesis quality of our method on the real-world DTU~\cite{jensen2014dtu} and Mip-NeRF 360~\cite{barron2022mipnerf360} datasets \emph{before} any editing. Table~\ref{tab:real-world-nvs} reports the mean PSNR, SSIM, and LPIPS against 3DGS~\cite{kerbl20233dgauss}, 2DGS~\cite{Huang20242DGS} and Mani-GS~\cite{Gao2024ManiGSGS}. As full resolution training on the Mip-NeRF 360 dataset exceeds the memory of our GPUs, we evaluate our method at \emph{half} the resolution of the Mip-NeRF 360 dataset. The numbers for the Gaussian splatting-based baselines are from their original papers and NeST-Splatting~\cite{Zhang2025NeuralST}.

\begin{table}[htb]
\centering
\setlength{\tabcolsep}{5pt}
\caption{Mean novel-view synthesis quality on real-world DTU~\cite{jensen2014dtu} and Mip-NeRF 360~\cite{barron2022mipnerf360} scenes (no editing). $\ddagger$~For Mip-NeRF 360 dataset P-CORE is trained and rendered at the half-resolution of the dataset, as full resolution evaluation exceeds our GPU memory.}
\label{tab:real-world-nvs}
\resizebox{\linewidth}{!}{%
\begin{tabular}{l|ccc|ccc}
\toprule
\multirow{2}{*}{Method} & \multicolumn{3}{c|}{DTU~\cite{jensen2014dtu}} & \multicolumn{3}{c}{Mip-NeRF 360~\cite{barron2022mipnerf360}} \\
\cmidrule(lr){2-4} \cmidrule(lr){5-7}
& PSNR $\uparrow$ & SSIM $\uparrow$ & LPIPS $\downarrow$ & PSNR $\uparrow$ & SSIM $\uparrow$ & LPIPS $\downarrow$ \\
\midrule
3DGS~\cite{kerbl20233dgauss}   & 33.77 & 0.965 & 0.044 & 26.54 & 0.776 & 0.266 \\
2DGS~\cite{Huang20242DGS}   & 33.89 & 0.966 & 0.048 & 27.03 & 0.805 & 0.223 \\
Mani-GS~\cite{Gao2024ManiGSGS} & 31.50 & 0.943 & 0.088 & -- & -- & -- \\
P-CORE (Ours)$^\ddagger$  & 33.48 & 0.973 & 0.023 & 28.04 & 0.808 & 0.183 \\
\bottomrule
\end{tabular}}
\end{table}

\end{document}